\documentclass[table]{article}

\usepackage[preprint,nonatbib]{neurips_2023}

\usepackage{times}
\usepackage{latexsym}

\usepackage{microtype}

\usepackage{inconsolata}
\usepackage[utf8]{inputenc} 
\usepackage[T1]{fontenc}    
\usepackage{hyperref}       
\usepackage{url}            
\usepackage{booktabs}       
\usepackage{amsfonts}       
\usepackage{nicefrac}       
\usepackage{microtype}      
\usepackage{tcolorbox}
\usepackage{tabularx}
\usepackage{longtable}

\usepackage{xcolor}
\definecolor{green}{HTML}{66FF66}
\definecolor{myGreen}{HTML}{009900}
\newcolumntype{b}{X}
\newcolumntype{s}{>{\hsize=.3\hsize}X}
\newcolumntype{z}{>{\hsize=.8\hsize}X}

\usepackage{graphicx}
\usepackage{pdfpages}
\newcommand{\dataset}[0]{\textsc{IndicVoices}}
\newcommand{\model}[0]{IndicASR}

\usepackage{subfig}
\usepackage{caption} 
\newcommand{\cmark}{\ding{51}}%
\newcommand{\xmark}{\ding{55}}%
\usepackage{multirow}
\usepackage{pifont}
\usepackage[linesnumbered,ruled,vlined]{algorithm2e}
\usepackage{algpseudocode}
\usepackage{amsmath}

\title{\textsc{IndicVoices}: Towards building an Inclusive Multilingual Speech Dataset for Indian Languages}

\newcommand{\balpha}{{\boldsymbol{\alpha}}}
\newcommand{\bbeta}{{\boldsymbol{\beta}}}
\newcommand{\bgamma}{{\boldsymbol{\gamma}}}

\newcolumntype{Y}{>{\raggedleft\arraybackslash}X}

\tcbset{tab1/.style={fonttitle=\bfseries\large,fontupper=\normalsize\sffamily,
colback=yellow!10!white,colframe=red!75!black,colbacktitle=Salmon!40!white,
coltitle=black,center title,freelance,frame code={
\foreach \n in {north east,north west,south east,south west}
{\path [fill=red!75!black] (interior.\n) circle (3mm); };},}}

\tcbset{tab2/.style={enhanced,fonttitle=\bfseries,fontupper=\normalsize\sffamily,
colback=yellow!10!white,colframe=red!50!black,colbacktitle=Salmon!40!white,
coltitle=black,center title}}

\author{
  {\bf
    {Tahir Javed\thanks{Corresponding author: tahir@cse.iitm.ac.in}}
    \hspace{0.1em}
    $^{\balpha}$$^{\bbeta}$
    \enskip
    Janki Atul Nawale
    \hspace{-0.5em}
    $^{\balpha}$
   \vspace{5pt}
  } \\
  {\bf
    Eldho Ittan George
    \hspace{-0.5em}
    $^{\balpha}$
    \enskip
    Sakshi Joshi
    \hspace{-0.5em}
    $^{\balpha}$$^{\bbeta}$
    \enskip
     Kaushal Santosh Bhogale
    \hspace{-0.5em}
    $^{\balpha}$$^{\bbeta}$
    \enskip
    Deovrat Mehendale
    \hspace{-0.5em}
    $^{\balpha}$
    \vspace{5pt}
  } \\
  {\bf
    Ishvinder Virender Sethi
    \hspace{-0.5em}
    $^{\balpha}$
    \enskip
    Aparna Ananthanarayanan
    \hspace{-0.5em}
    $^{\balpha}$
    \enskip
    Hafsah Faquih
    \hspace{-0.5em}
    $^{\balpha}$
    \enskip
    Pratiti Palit
    \hspace{-0.5em}
    $^{\balpha}$
    \vspace{5pt}
  } \\
  {\bf
    Sneha Ravishankar
    \hspace{-0.5em}
    $^{\balpha}$
    \enskip
    Saranya Sukumaran
    \hspace{-0.5em}
    $^{\balpha}$
    \enskip
    Tripura Panchagnula
    \hspace{-0.5em}
    $^{\balpha}$
    \enskip
    Sunjay Murali
    \hspace{-0.5em}
    $^{\balpha}$
    \vspace{5pt}
  } \\
  {\bf
    Kunal Sharad Gandhi
    \hspace{-0.5em}
    $^{\balpha}$
    \enskip
    Ambujavalli R
    \hspace{-0.5em}
    $^{\balpha}$
    \enskip
    Manickam K M
    \hspace{-0.5em}
    $^{\balpha}$
    \enskip
    C Venkata Vaijayanthi
    \hspace{-0.5em}
    $^{\balpha}$
    \enskip
    \vspace{5pt}
  } \\
  {\bf
    Krishnan Srinivasa Raghavan Karunganni
    \hspace{-0.5em}
    $^{\balpha}$
    \vspace{5pt}
  } \\
  {\bf
    \vspace{1.5pt}
    Pratyush Kumar
    $^{\bbeta}$$^{\bgamma}$
    \hspace{-0.5em}
    $ $
    \enskip
    Mitesh M Khapra
    \hspace{-0.5em}
    $^{\balpha}$$^{\bbeta}$
    \vspace{5pt}
  } \\
  {
    $^{\balpha}$AI4Bharat
   } \\
   {
    $^{\bbeta}$Indian Institute of Technology Madras \quad
  } \\
  {
    $^{\bgamma}$Sarvam AI
  }
}

\begin{document}

\maketitle

\begin{abstract}
We present \dataset, a dataset of natural and spontaneous speech containing a total of 7348 hours of read (9\%), extempore (74\%) and conversational (17\%) audio from 16237 speakers covering 145 Indian districts and 22 languages. Of these 7348 hours, 1639 hours have already been transcribed, with a median of 73 hours per language. Through this paper, we share our journey of capturing the cultural, linguistic and demographic diversity of India to create a one-of-its-kind inclusive and representative dataset. More specifically, we share an open-source blueprint for data collection at scale comprising of standardised protocols, centralised tools, a repository of engaging questions, prompts and conversation scenarios spanning multiple domains and topics of interest, quality control mechanisms, comprehensive transcription guidelines and transcription tools. We hope that this open source blueprint will serve as a comprehensive starter kit for data collection efforts in other multilingual regions of the world. Using \dataset, we build \model, the first ASR model to support all the 22 languages listed in the 8th schedule of the Constitution of India. All the data, tools, guidelines, models and other materials developed as a part of this work will be made publicly available. 

\begin{center}
    \renewcommand{\arraystretch}{1.2}
    \begin{tabular}{rcl}
         \textbf{Explore IndicVoices} & \url{https://ai4bharat.iitm.ac.in/indicvoices}\\
    \end{tabular}
    \end{center}

\end{abstract}

\section{Introduction}
Recent advancements in Automatic Speech Recognition (ASR) have achieved remarkable success in English \cite{whisper,wav2vec2,data2vec,wavlm, DBLP:conf/interspeech/GulatiQCPZYHWZW20}, with single digit WERs on multiple benchmarks \cite{LibreSpeech, artie, commonvoice-english, tedlium, voxpopuli-en, Svarah}. However, despite several massively multilingual efforts such as OpenAI's Whisper \cite{whisper}, Google's USM \cite{google-usm} and Meta's MMS \cite{DBLP:journals/corr/abs-2305-13516}, the progress on mid and low resource languages is not at par with English. The main reason for this is lack of labeled data in these languages. 
To mitigate this problem and improve the performance on low resource languages, three popular solutions are commonly used: (i) self supervised learning using unlabeled data \cite{DBLP:conf/aaai/JavedDRBRKKK22}, (ii) multilingual training to enable crosslingual transfer from high resource languages \cite{DBLP:journals/corr/abs-2305-13516} and (iii) mining \textit{noisy} training data by auto-aligning audio and transcripts curated from the web \cite{DBLP:journals/corr/abs-2308-11466, DBLP:conf/icassp/BhogaleRJDKKK23}. While these solutions help to an extent, they do not address the elephant in the room, which is lack of \textit{sufficient}, \textit{diverse} and \textit{high quality} training data in these languages. In this study, we confront the elephant and set out with an ambitious goal to collect spontaneous speech data for Indian languages while respecting the linguistic, cultural and demographic diversity of India.

\begin{figure}
    \centering

    \includegraphics[width=0.7\linewidth]{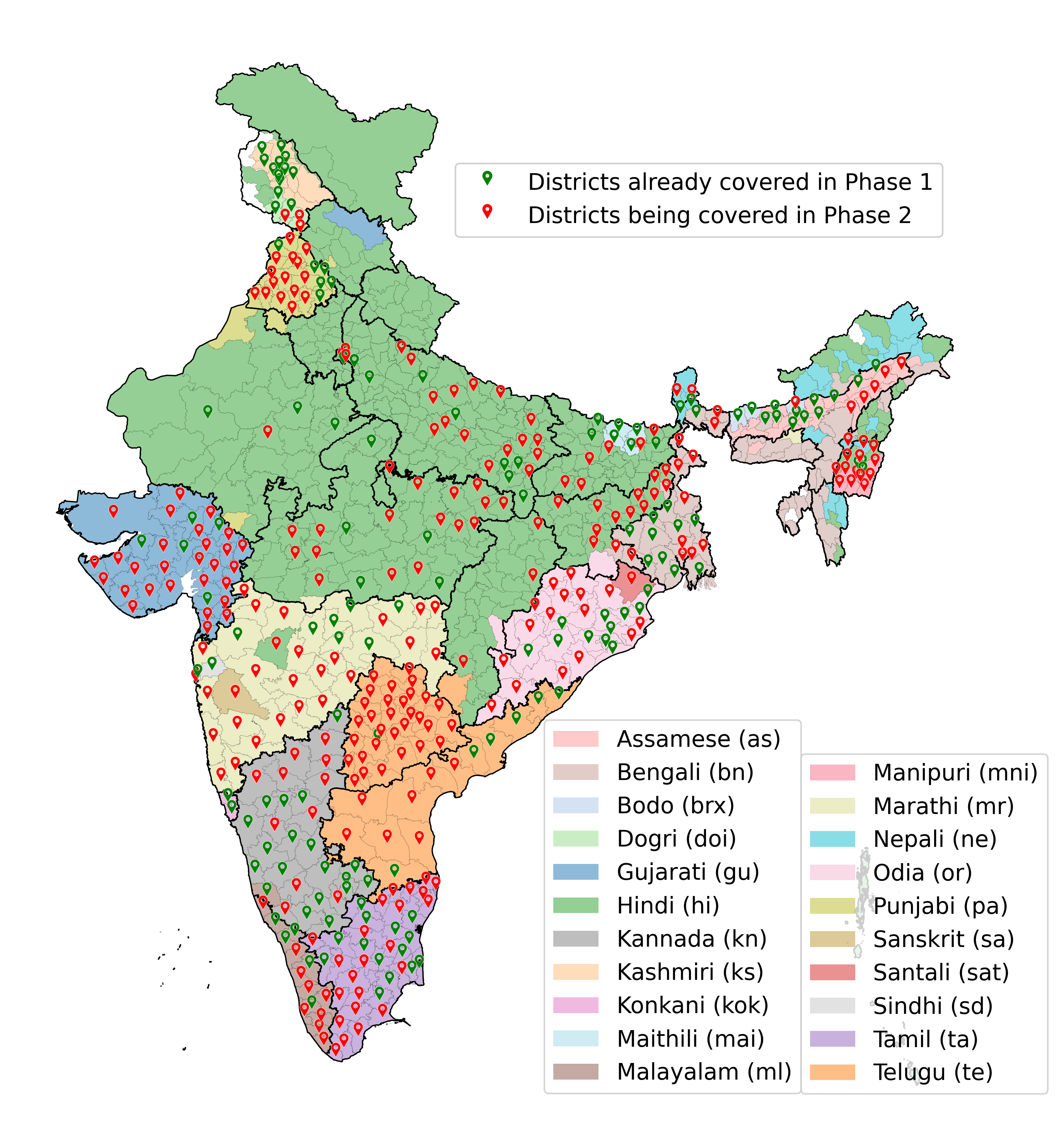}
    \caption{Primary regions of India where each of the 22 languages is spoken.} 
    \label{fig:language_atlas_of_India}
\end{figure}

To understand the scale of the problem, it is important to note that there are 22 languages listed in the 8th schedule of Indian constitution, belonging to 4 different language families. These languages have a collective speaker base of 1.2B, spread across 742 districts in India. Figure \ref{fig:language_atlas_of_India} shows the regions of India in which each of these languages is primarily spoken. The geographical spread of the 22 languages across India contributes to a rich diversity in culture, traditions, customs, beliefs, lifestyles, preferences and interests, all underpinned by a shared Indian ethos. The onus then is to collect data which is inclusive and representative of this diversity of India.

\begin{table}[!t]
\centering
\scriptsize
\begin{tabular}
{@{\hspace{0.5em}}l|rr@{\hspace{0.6em}}rr@{\hspace{0.8em}}rc@{\hspace{0.8em}}c@{\hspace{0.8em}}cc@{\hspace{0.5em}}c@{\hspace{0.5em}}}
\arrayrulecolor{green}\toprule\arrayrulecolor{black}
\multicolumn{1}{@{\hspace{0.3em}}l|}{\multirow{2}{*}{\textbf{Dataset}}} & \multicolumn{1}{c|}{\multirow{2}{*}{\textbf{\#L}}} & \multicolumn{2}{@{}c@{}|}{\textbf{\#Hours}} & \multicolumn{1}{@{}c@{}|}{\multirow{2}{*}{\textbf{\#Sp}}} & \multicolumn{1}{@{\hspace{0.7em}}c|}{\multirow{2}{*}{\textbf{\#D}}} & \multicolumn{3}{@{}c@{}|@{}}{\textbf{Type}} & \multicolumn{2}{@{}c@{}}{\textbf{Channel}} \\ \cmidrule{3-4} \cmidrule{7-11} 
\multicolumn{1}{c|}{} & \multicolumn{1}{c|}{} & \multicolumn{1}{c}{\textbf{Un}} & \multicolumn{1}{c|}{\textbf{Lb}} & \multicolumn{1}{c|}{} & \multicolumn{1}{c|}{} & \textbf{R} & \textbf{E} & \multicolumn{1}{@{}c|@{}}{\textbf{C}} & \textbf{WB} & \textbf{NB} \\ 
\arrayrulecolor{green}\midrule\arrayrulecolor{black}

CommonVoice\cite{commonvoice-english} & 8 & - & 373 & - & 4 & \xmark & \xmark & \xmark & \cmark & \xmark \\
FLEURS\cite{fleurs} & 13 & - & 163 & - & - & \cmark & \xmark & \xmark & \cmark & \xmark \\
MSR\cite{srivastava18_sltu} & 3 & - & 150 & 1286 & 1 & \xmark & \xmark & \cmark & \xmark & \cmark \\
OpenSLR\cite{kjartansson-etal-sltu2018} & 3 & - & 618 & 1513 & - & \cmark & \xmark & \xmark & \cmark & \xmark \\
CMS\cite{he-etal-2020-open} & 6 & - & 35 & 243 & 1 & \cmark & \xmark & \xmark & \cmark & \xmark \\
MUCS\cite{diwan2021multilingual} & 3 & - & 351 & 158 & 4 & \cmark & \xmark & \xmark & \cmark & \xmark \\
Kathbath\cite{10.1609/aaai.v37i11.26521} & 12 & - & 1684 & 1218 & 3 & \cmark & \xmark & \xmark & \cmark & \xmark \\
Shrutilipi\cite{10096933} & 12 & - & 6457 & - & - & \cmark & \xmark & \xmark & \cmark & \xmark \\
Graamvaani\footnote{https://sites.google.com/view/gramvaaniasrchallenge/home/} & 1 & 1000 & 108 & - & - & \xmark & \xmark & \cmark & \xmark & \cmark \\
IISc-Mile\cite{mile_1,mile_2} & 2 & - & 500 & 1446 & - & \cmark & \xmark & \xmark & \cmark & \xmark \\
KDC\footnote{https://openslr.org/122/} & 1 & - & 1 & 43 & - & \xmark & \xmark & \cmark & \xmark & \cmark \\
Vāksañcayah\cite{adiga-etal-2021-automatic} & 1 & - & 78 & 27 & 8 & \xmark & \xmark & \xmark & \cmark & \xmark \\
IIITH-ISD\cite{Prahallad2012TheII} & 7 & - & 11 & 35 & 1 & \cmark & \xmark & \xmark & \cmark & \xmark \\
IITB-MSC\cite{marathidata} & 1 & - & 109 & 36 & 1 & \cmark & \xmark & \xmark & \cmark & \xmark \\
SMC-MSC\footnote{https://blog.smc.org.in/malayalam-speech-corpus/} & 1 & - & 2 & 75 & 4 & \xmark & \xmark & \xmark & \xmark & \xmark \\
IITM\footnote{https://sites.google.com/view/indian-language-asrchallenge/home} & 3 & - & 690 & - & - & \cmark & \xmark & \xmark & \cmark & \xmark \\
NPTEL\cite{Bhogale2023VistaarDB} & 8 & - & 857 & - & 1 & \cmark & \xmark & \xmark & \cmark & \xmark \\
IndicTTS\cite{2016ResourcesFI} & 13 & - & 225 & 25 & 4 & \cmark & \xmark & \xmark & \cmark & \xmark \\
Svarah\cite{javed23_interspeech} & 1 & - & 10 & 117 & 37 & \cmark & \cmark & \cmark & \cmark & \xmark \\
SPRING-INX\cite{r2023springinx} & 10 & - & 2005 & 7609 & 16 & \cmark & \cmark & \cmark & \cmark & \cmark \\
SPIRE-SIES\cite{singh2023spiresies} & 1 & 171 & 23 & 1607 & - & \xmark & \cmark & \xmark & \cmark & \xmark \\
\midrule
\textbf{IndicVoices} & \textbf{22} & \textbf{7348} & \textbf{1639} & \textbf{16237} & \textbf{52} & \cmark & \cmark & \cmark & \cmark & \cmark \\
\arrayrulecolor{green}\bottomrule\arrayrulecolor{black}
\end{tabular}
\centering
\caption{Comparison of existing datasets for Indian languages in terms of number of languages (\textbf{\#L}), number of hours of labeled (\textbf{Lb}) and unlabeled (\textbf{Ub}) data, number of speakers (\textbf{\#Sp}), number of domains (\textbf{\#D}), type of data (\textbf{R}ead/\textbf{E}xtempore/\textbf{C}onversation, wide (\textbf{WB}) or narrow band (\textbf{NB})).} 
\label{tab:related_work}
\end{table}

Keeping this in mind, we set out to collect data for each of these 22 languages while ensuring a comprehensive representation across various demographics such as gender, age, educational background, and geographic location, with specific quotas for each category. Additionally, we aimed for diversity in the vocabulary, content, and recording channels, including a mix of read speech, voice commands, extempore discussions, and both wide and narrow-band recordings. We also aim to maintain a balanced representation of urban and rural speakers and recording devices, and include a significant portion of data recorded in noisy environments representing everyday usage of ASR systems. This is indeed a one-of-its-kind effort which supersedes past efforts (Table \ref{tab:related_work}), in terms of number of languages covered, number of districts covered, diversity in content/topics, recording channels, nature of data (read, extempore, conversational) and so on. 

To facilitate data collection at this scale, we created a clear framework which can be replicated across languages and diverse geographical locations. First, we created a centralized repository, accompanied by a frontend mobile application for enabling data collection in a remote and distributed setup using a standard procedure. Second, to elicit meaningful responses from participants capturing local customs, traditions, beliefs, lifestyles, preferences, culture, etc., we created a repository of 2.5K questions, 46.6K prompts, and 1.1K (Dogri) to 4.1K (Hindi) role-play scenarios covering 21 domains and 28 topics of interest, and anchored in practical, day-to-day usage scenarios. The data collection was facilitated by a countrywide network consisting of agencies, local universities, local NGOs, and social sector professionals who acted as regional influencers to engage a variety of participants. To ensure quality in such a remote and distributed setup, we built an \textit{in-house} quality control team who listened to every audio collected from the field and ensured adherence to our stringent acceptance criteria. Lastly, we built a team of transcribers consisting of makers, checkers and supercheckers to transcribe the data collected from the field. We also create an elaborate two-level transcription guideline which addresses the unique normalisation challenges in Indian languages arising due to a significant gap between formal written language and colloquially spoken language. 

\dataset~is the result of this massive effort involving a total of 1893 personnel in a variety of roles such as language experts, local mobilisers, coordinators, quality control experts, transcribers, language leads and project managers. It contains a total of 7348 hours of read (9\%), extempore (74\%) and conversational (17\%) audio data from 16237 speakers covering 145 districts and 22 languages, of which, 1639 hours have already been transcribed (the rest is under progress). We also carve out a robust ASR benchmark from \dataset~ which would allow Indic ASR models to be evaluated on practical everyday usage scenarios across different demographics. Note that, while in this paper we mainly focus on ASR data, the collected data can be used for several other purposes such as speaker diarization, speaker identification, speaker verification, language identification, intent detection, entity extraction, query by example and audio denoising. We will discuss more about these tasks in subsequent work. While we are still far from our eventual goal of collecting $\sim$20000 hours of transcribed data for Indian languages, we are sharing our initial experiences with the community early on so that we can get feedback and incorporate changes in the remaining phase of the project. We hope that all the data, tools, guidelines and other material created as a part of this work serve as an open source framework for data collection projects in other multilingual regions of the world such as Africa, Southeast Asia and Latin America. 

\section{The Wishlist}
\label{sec:wishlist}
\begin{figure}
    \centering
    \includegraphics[width=0.8\linewidth]{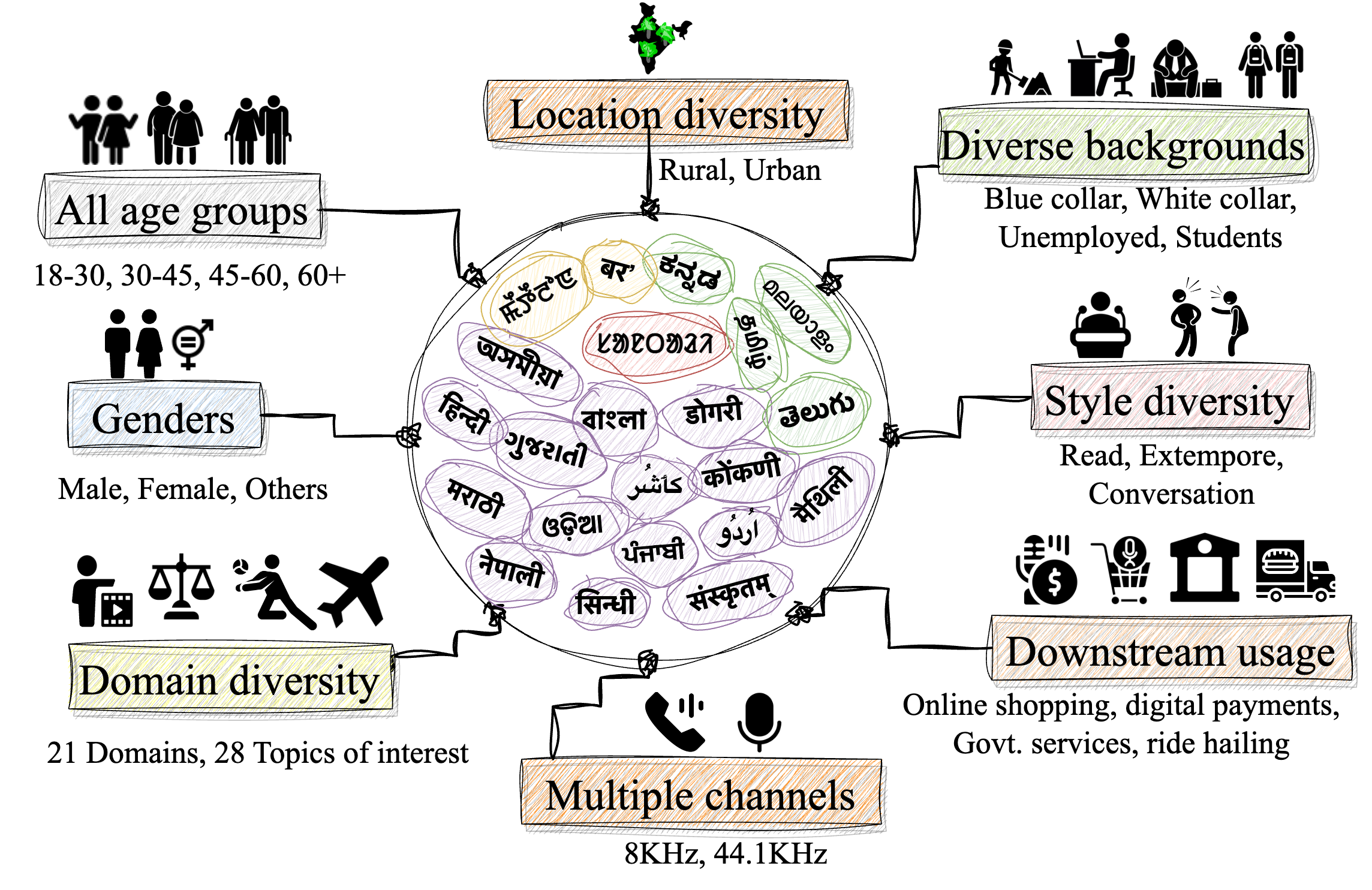}
    \caption{Demographic, geographical, domain, style and usage diversity wishlist for our data collection.}
    \label{fig:wishlist}
\end{figure}
The overarching goal of this project is to develop accurate speech technology for Indian languages. This is vital for ensuring digital inclusion by providing voice as a natural interface to a population of 1.2B+ speakers. The question then is ``What would it take to build accurate speech technology for a diverse set of Indian languages?'' The short answer is, of course, \textit{data}. This leads to the next question, ``How do you capture data which is inclusive and representative of the linguistic, demographic and cultural diversity of a country like India while being grounded in practical usage in downstream applications?" To answer this question, we outline a basic wishlist that we adhere to in our data collection effort.

\noindent\textbf{WL1. Diversity of speakers:} To ensure demographic diversity, we include speakers from various age groups, genders, educational backgrounds, and professions. We have four age brackets (18-30, 30-45, 45-60, 60+), with a minimum of 15\% representation from each. Male and female  genders are balanced, with efforts to include speakers of other genders. We consider four educational levels (no schooling, high school, graduates, postgraduates), ensuring at least 15\% from each category. Additionally, we include speakers from three professional backgrounds (blue collar, white collar, unemployed), with a minimum of 10\% from each. The quotas don't sum to 100\% to allow flexibility and practicality in representation, especially considering the varying availability of certain groups in different regions (e.g., in some rural areas it may be difficult to source enough post graduates). Thus, once the minimum quota for each sub-class is filled ensuring fair representation, there is flexibility to source remaining speakers from any of the sub-classes.

\noindent\textbf{WL2. Diversity of locations:} Given the geographical spread of different languages as shown in Figure \ref{fig:language_atlas_of_India} it is often the case that for a given language, the accents and colloquial usage varies from one region to another. To capture this diversity we ensure that we collect data from at least 60-80\% of the districts where a given language is spoken as the primary language. Apart from capturing multiple accents, this also allows us to capture regional variations in culture, interests, preferences, etc. This also ensures that speakers from both urban and rural districts are represented in our data.

\noindent\textbf{WL3. Diversity in content:} It is important that the spoken content is not repetitive and covers multiple domains. For example, it is not prudent to ask every speaker the same questions ``Tell us something about your district?'' as they are likely to say very similar things. Hence, it is important to have questions related to multiple domains and topics of interest so that the speakers could engage with meaningful  responses which capture regional knowledge.

\noindent\textbf{WL4. Diversity in styles:} We ensure that the data has a good representation of different styles of speech, viz., read speech, extempore speech and conversational speech involving two parties. 

\noindent\textbf{WL5. Broad vocabulary coverage:} While ensuring diversity in content as defined above would ensure coverage of multiple domains it may still not ensure broad coverage of vocabulary. For example, it is possible that very few participants speak about topics from the legal domain resulting in poor coverage of vocabulary from this domain. To ensure broad coverage of vocabulary, we collect a fraction of the data as read speech where participants simply read sentences shown to them. For such read speech, we select sentences from 13 domains, \textit{viz.},  legal, governance, history, geography, tourism, stem, religion, business, sports, entertainment, health, culture and news thereby ensuring broad coverage of vocabulary.

\noindent\textbf{WL6. Downstream Usage:} It is important that the data should have a good representation of scenarios encountered in everyday use cases in which speech recognition systems are likely to be used. Keeping this in mind, a fraction of the data is collected by requesting participants to perform every day interactions such as hailing a ride, making a digital payment, interacting with government services, ordering food and groceries and so on. This again ensures good coverage of entities such as brand names, numbers, alphanumeric codes, etc. which may not appear in regular extempore conversations. Further, keeping downstream usage in mind, we ensure that a fraction of the data is collected on a 8 KHz telephony channel. This ensures representation of the usage patterns of many low-income users in India who do not have smartphones and are likely to interact with Interactive Voice Response Systems on a 8KHz telephone channel. 

\noindent\textbf{WL7. Diversity in recording conditions:} Lastly, we take cognizance of the fact that in everyday usage, speech recognition systems are likely to be used in noisy environments, on a variety of devices with different quality of microphones. Hence, we ensure that the data collection happens in diverse environments such as farms, houses, open spaces, etc., using users' mobile phones instead of a standard high quality recording setup thereby ensuring that the data captures regular usage conditions of an average user and is not biased towards high quality setups in controlled environments.

\begin{figure}[!t]
    \centering
    \includegraphics[width=\linewidth]{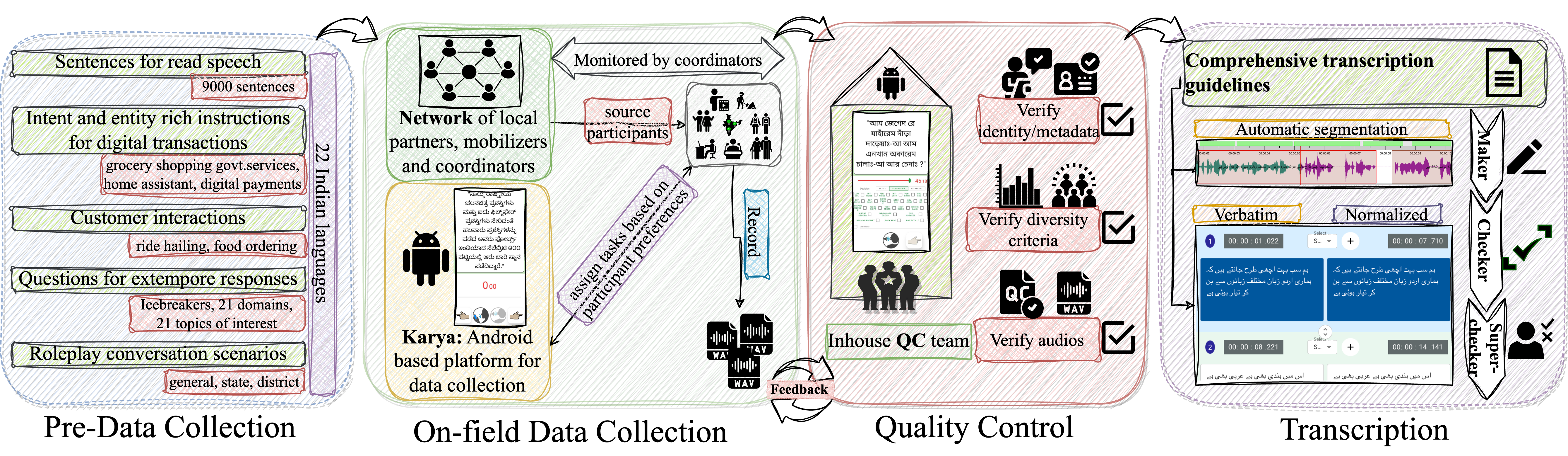}
    \caption{The four stages in the process used for collecting \dataset. The collection and quality control is done using our in-house tool, an extension of Karya. The transcription is done using our in-house tool, Shoonya.}
    \label{fig:overall_process}
\end{figure}

We now describe our process for data collection which is mindful of the above wishlist. At a high level, the process contains of four stages (see Figure \ref{fig:overall_process}), \textit{viz.}, (i) preparation before data collection (ii) on-field data collection (iii) quality control and (iv) transcription.  We describe each of these in detail below.

\section{Pre-collection Data Preparation}
Before collecting audio samples, some background preparatory work is needed to ensure that participants contribute meaningfully. For example, most participants do not respond well to simple prompts such as ``let us talk for 10 minutes'' or ``enact some everyday interactions that you carry out on your phone using apps or calls''. Hence, to elicit rich responses and collect diverse content as listed in our wishlist we need to create some sentences, questions and scenarios which would be read, answered or enacted by the participants. We list these resources below and also mention the specific item(s) in the wishlist that they cover.

\subsection{Sentences for read speech (WL4, WL5)} 
\label{subsec:3.1}
While read speech is not spontaneous, it is still useful to ensure coverage of specific words from different domains of interest. To do so, we first collected 9K sentences from Wikipedia articles belonging to 13 different domains, \textit{viz., news, entertainment, geography, health, education, sports, tourism, culture, government, industry, economy, legal} and \textit{religion}. These sentences were then manually translated to 22 languages with the help of in-house translators. This ensures words such as \textit{hydroelectricity, photosynthesis, etc.},  which may not be encountered in extempore interactions with participants, are covered in the data. We would then ask participants to read these sentences as it is to collect \textit{read speech}.

\subsection{Tasks involving voice assistants (WL6)} A critical application of speech recognition is found in digital voice assistants, which facilitate the execution of daily tasks and support interaction with frequently used services. To ensure that the collected data has a fair representation of such downstream usage we created task specific instructions encountered in important applications as listed below. These instructions would then be spoken by participants during data collection.

\noindent\underline{\textit{Everyday tasks:}} We refer to the MASSIVE \cite{fitzgerald-etal-2023-massive} dataset which contains human created utterances directed towards an in-home personal assistant. These utterances are questions or requests covering 60 intents and 18 domains. We sampled 9K English utterances from MASSIVE covering all domains and intents, and translated them to each of the 22 Indian languages with the help of human translators. The translators were instructed to do a colloquial translation which contains code-mixing to reflect the usage pattern of an average speaker of that language, as opposed to a very formal translation. They were also instructed to perform localisation by replacing Western locations, brands, cuisines, etc. by Indian entities. 

\begin{figure}[!t]
    \centering
    \includegraphics[width=\linewidth]{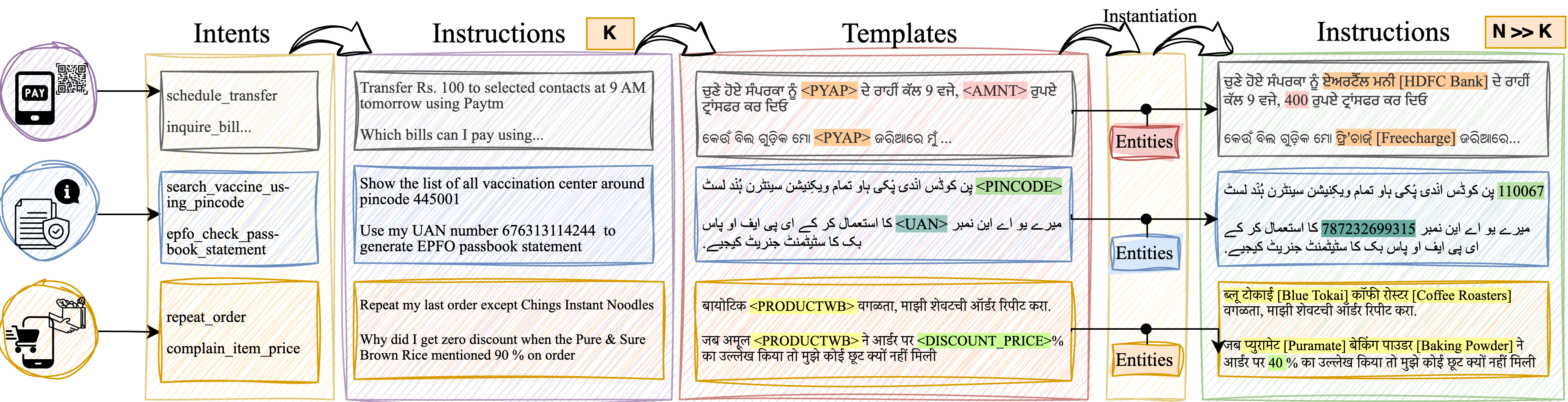}
    \caption{Standardised process for using a small number ($k$) of human-created instructions to create templates and then instantiate a large number of instructions ($N$) by replacing entities in the templates.}
    \label{fig:template_process}
\end{figure}

\noindent\underline{\textit{Digital financial transactions:}} In India, digital payments have experienced substantial growth due to UPI 
(Unified Payments Interface), widespread mobile usage, and the necessity for contactless transactions during the COVID-19 pandemic. However, for a more inclusive growth and ease of access, it is important to enable voice-based digital transactions. Therefore, accurately representing such transactions in the data we collect is imperative, especially given the distinctive characteristics of such data which contains (i) rich numeric vocabulary and long number sequences (ii) code-mixing to integrate everyday banking terminology in English with regional languages and (iii) regional and international bank names. To create representative instructions we use the following procedure (see Fig. \ref{fig:template_process}). We first enumerate intents encountered in a typical digital payments app such as \textit{transfer money}, \textit{check balance}, \textit{pay electricity bill} and so on. For each of these intents we ask human annotators to write 4-5 utterances which would convey the intent in colloquial everyday language. We then replace entities such as \textit{bank names}, \textit{account numbers}, \textit{transaction amounts}, etc. with placeholders resulting in templates of instructions. Next, we create a list of all entities such as multi-digit or alpha-numeric (fake) bank account numbers, small to large transaction amounts, a list of all bank names and so. We then create multiple instructions from the templates by replacing the placeholders by entities from above lists. In total, we create 9.4K single turn instructions in each of the 22 languages, which would then be spoken out by participants.

\noindent\underline{\textit{Online grocery transactions:}} Another important use-case is voice based interactions with online grocery websites and apps which many Indians engage with on a daily basis. The utterances in such interactions also have unique characteristics as they contain (i) brand names (ii) product names (iii) numeric quantities and (iv) units for specifying quantities. To ensure that this important use-case is represented in our data, we followed the same procedure as above to create such interactions. Specifically, we identified intents, asked humans to write utterances capturing these intents, converted the utterances to templates by replacing entities by placeholders, created a list of entities such as numbers, brand names, product names, etc. and lastly created multiple interactions by replacing the placeholders in the templates by these real world entities.  In total, we create 9.3K single turn instructions covering 31 intents and 1.7K unique entities in each of the 22 languages.

\noindent\underline{\textit{Digital government services:}} Another important use case is citizen interactions with government services. Umang is an all-in-on app launched by the Indian government which allows citizens to access services related to passport, unique financial id, unique national id, government welfare schemes, and so on. To ensure representation of this use-case we again created typical citizen interactions using the same template based approach as described above. These templates covered 77 intents and several entities such as vaccine names, welfare scheme names, college names, district names, document types, college names, etc. In total, we create 9.2K single turn instructions covering  294 unique entities in each of the 22 languages.

\subsection{Customer care interactions (WL6)} All the use cases described above involved scripted interactions which were created by us to ensure broad coverage of entities, intents and use case specific characteristics (such as code mixing). We now consider two more applications and create certain scenarios which would be enacted by participants. These scenarios involved interactions with customer care services of ride hailing apps and food delivery apps. For example, consider the following scenario: ``\textit{Imagine that you and your family are planning a road trip and you want to borrow a rental car to drive. Inquire with the car provider of your choice and ask about the prices per km, how many days you will be traveling, the name of the place you are planning to travel.}''. The participant would then be expected to enact these scenarios using their own imagination as oppose to following a script prepared by us. We created 35 and 68 such scenarios for ride hailing and food delivery respectively, and translated them to all the 22 languages with appropriate localisation. Note that there are several such other applications involving customer care interactions but we chose these 2 as they are easy to related to and do not involve any PPI (Private Personal Information) which could be accidentally leaked.  
\begin{figure}[!t]
    \centering
    \includegraphics[width=0.9\linewidth]{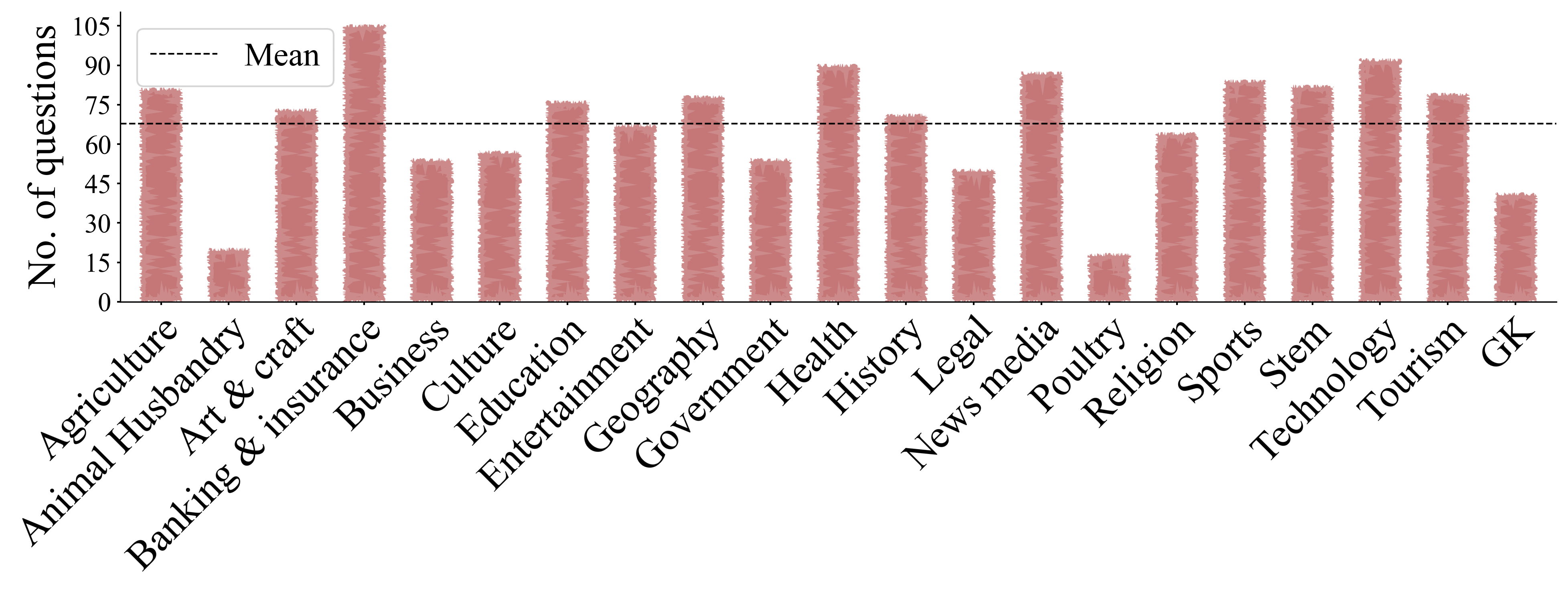}
    \includegraphics[width=0.9\linewidth]{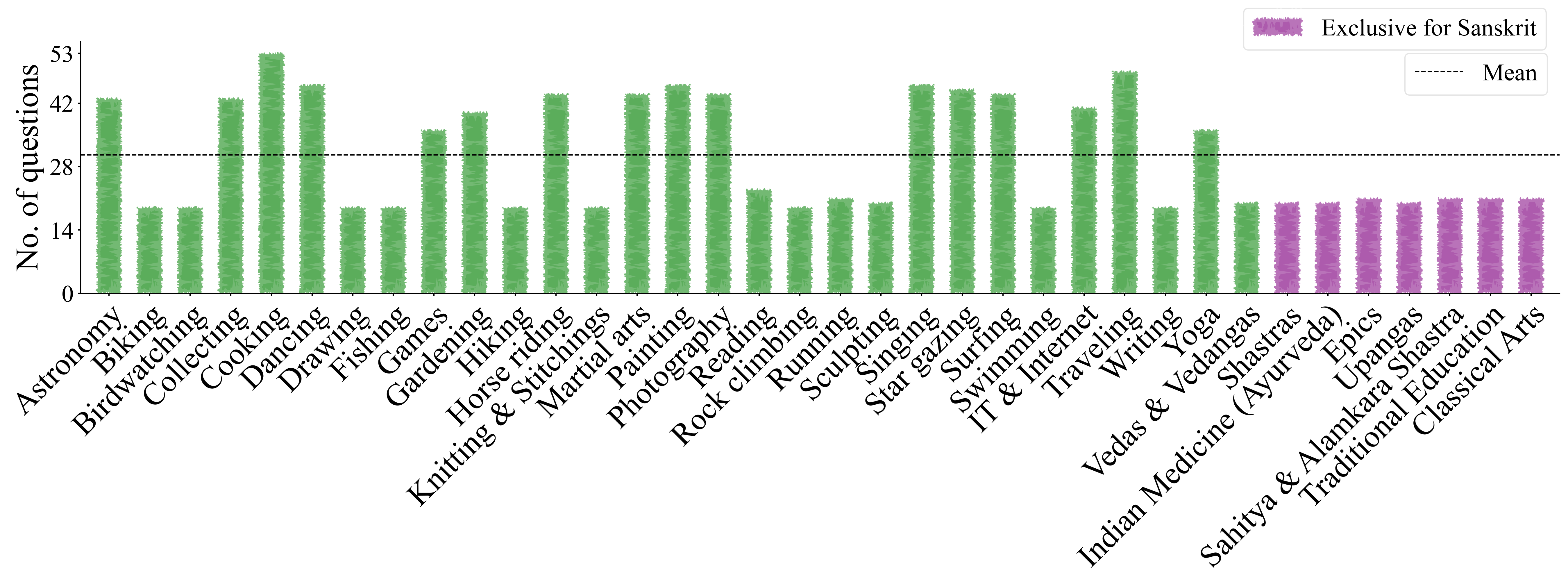}
    \caption{Number of engaging questions created for different domains (top) and topics of interest (bottom). These questions are contextualised and colloquially translated into 22 languages.}
    \label{fig:distribution_of_questions}
\end{figure}

\subsection{Extempore questions (WL3, WL4)} 
\label{sec:extempore-questions} 
Extempore conversations are crucial in capturing natural speech, reflecting the spontaneous and unscripted nature of daily communication. However, eliciting natural, engaging dialogue in a data collection setting poses significant challenges. Participants often struggle to connect with or articulate thoughts on unfamiliar topics, leading to hesitant or superficial responses. To mitigate this, we offered participants a choice from 21 domains and 28 topics of interest, aiming to cover a broad spectrum of relatable content. However, we quickly realized that broad prompts like ``talk about politics'' were too vague and did not effectively engage participants. To address this, we collaborated with journalists to formulate well-thought-out questions that resonate on a personal level. Despite this initial effort, feedback from our initial pilots indicated that some questions were still too technical or failed to resonate across diverse age groups and the urban-rural divide. This led us to further refine and simplify the existing questions and add more question with the assistance of linguists, ensuring the language and content were accessible and engaging to all demographics. The linguists often relied on assistive tools like ChatGPT to suggest or simplify questions on specific topics and domains. Despite this, when we did a pilot in a rural district, we found that further modifications and additions were need to make the questions relatable to a rural audience. We appropriately did so based on feedback and help from local partners on the ground. 

Even with these improvements, we noticed a tendency for repetitive answers, particularly when questions touched on well-known subjects. In Madurai, for instance, the prompt to describe a notable tourist attraction overwhelmingly led to mentions of the \textit{Meenakshi Amman Temple}, overshadowing other significant but less prominent landmarks. To encourage a more diverse range of responses, we introduced specific hints for each question, nudging participants to recall and discuss less dominant, yet equally significant topics. Through this iterative process of feedback, refinement, and the strategic use of hints to broaden the scope of discussion, we crafted a comprehensive set of 2.5K questions, each accompanied by 8-15 hints per question (see Figure \ref{fig:distribution_of_questions}). We also needed some special modifications for Sanskrit which is not a language spoken by the masses and we describe these in Appendix A.

\begin{figure}[!t]
    \centering
    \includegraphics[width=\linewidth]{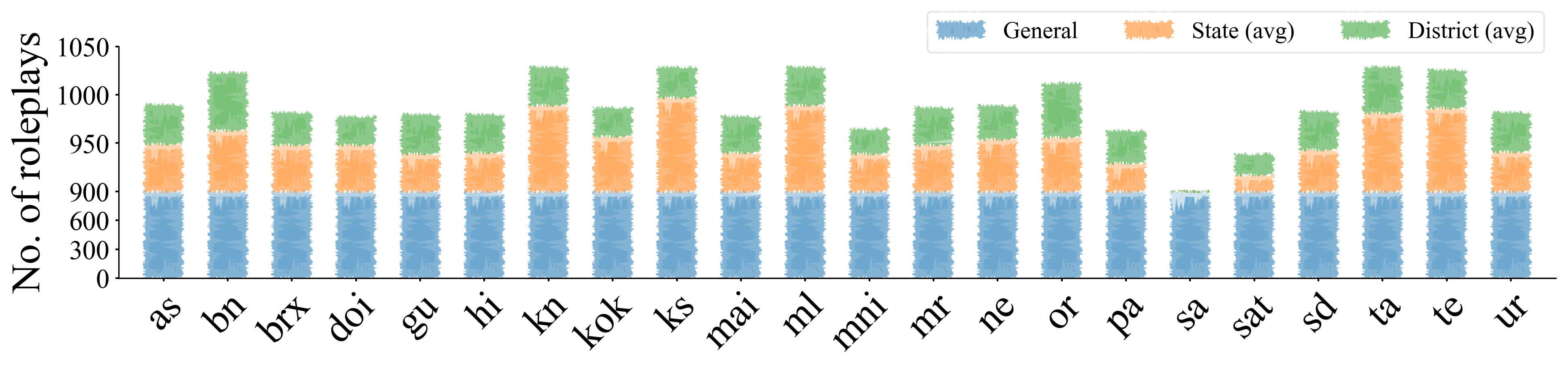}
    \caption{No. of general, state-specific and district specific roleplay conversational scenarios per language.}
    \label{fig:enter-label}
\end{figure}

\subsection{Icebreaker questions (WL3, WL4)}
During our initial pilots, we realised it was important to initiate conversations with an icebreaker for setting a comfortable environment, because the participants initially feel a bit reserved. To do so, we created 3 sets of easy-to-answer questions, with each set containing 50-70 questions. The first tier comprised of warm-up questions like ``Tell us about a typical day in  your life?'', ``Which is your favorite season and why?''. The second set comprised of questions related to everyday life such as ``Which electrical appliances do you use'', ``What are some measures you take to keep the areas around your house hygenic?'' and so on. The third set comprised of questions related to the participants mother tongue such as ``What are some tongue twisters in your language'', ``What are some idioms in your language?''. The participants were shown one randomly chosen question from each of these sets at the start of data collection to warm them up.

\subsection{Questions about named entities (WL6)}
\label{subsec:3.6}
Many downstream applications require speech recognition systems to recognise names, numbers, locations and dates. Given the geographical and cultural diversity of India these entities vary across the country (e.g., people names in South India are very different from those in North India). To represent these accurately in the data we had simple tasks such as ``Tell us 5 people names that come to your mind.'', ``Tell us 5 Indian cities, districts, states and 5 international cities'', ``Tell us any 5 dates'', ``Tell us any 5 numbers''.  

\begin{table}[!t]
\tiny
\centering
\begin{tabular}{l@{\hspace{0.7em}}|c@{\hspace{0.58em}}c@{\hspace{0.58em}}c@{\hspace{0.58em}}c@{\hspace{0.58em}}c@{\hspace{0.58em}}c@{\hspace{0.58em}}c@{\hspace{0.58em}}c@{\hspace{0.58em}}c@{\hspace{0.58em}}c@{\hspace{0.58em}}c@{\hspace{0.58em}}c@{\hspace{0.58em}}c@{\hspace{0.58em}}c@{\hspace{0.58em}}c@{\hspace{0.58em}}c@{\hspace{0.58em}}c@{\hspace{0.58em}}c@{\hspace{0.58em}}c@{\hspace{0.58em}}c@{\hspace{0.58em}}c@{\hspace{0.58em}}c}
\arrayrulecolor{green}\toprule
\textbf{Type} & \textbf{as} & \textbf{bn} & \textbf{brx} & \textbf{doi} & \textbf{gu} & \textbf{hi} & \textbf{kn} & \textbf{kok} & \textbf{ks} & \textbf{mai} & \textbf{ml} & \textbf{mni} & \textbf{mr} & \textbf{ne} & \textbf{or} & \textbf{pa} & \textbf{sa} & \textbf{sat} & \textbf{sd} & \textbf{ta} & \textbf{te} & \textbf{ur} \\
\arrayrulecolor{green}\midrule\arrayrulecolor{black}
Sentences for & 9146 & 9200 & 9199 & 8976 & 9200 & 9040 & 9006 & 9119 & 9135 & 9196 & 9200 & 8444 & 9065 & 9200 & 9199 & 8982 & 9199 & 9152 & 5036 & 9200 & 7721 & 9068 \\
read speech &  \\
\midrule
Everyday tasks & 9330 & 9359 & 9359 & 8000 & 8958 & 6363 & 9354 & 9093 & 5297 & 7962 & 9357 & 9356 & 9354 & 9350 & 9355 & 8171 & 9342 & 9346 & 7988 & 9359 & 9345 & 9302 \\
\midrule
Digital financial& 9448 & 9448 & 9448 & 9448 & 9448 & 9448 & 9448 & 9447 & 9448 & 9448 & 9448 & 9448 & 9448 & 9448 & 9448 & 9448 & 9448 & 9448 & 9444 & 9448 & 9448 & 9448 \\
transactions &  \\
\midrule
Online grocery & 9335 & 9335 & 9335 & 9335 & 9335 & 9335 & 9335 & 9334 & 9335 & 9335 & 9335 & 9335 & 9335 & 9334 & 9335 & 9335 & 9335 & 9335 & 9331 & 9335 & 9334 & 9335 \\
transactions & \\
\midrule
Digital govt. & 9274 & 9274 & 9274 & 9274 & 9274 & 9274 & 9274 & 9273 & 9274 & 9274 & 9274 & 9274 & 9274 & 9274 & 9274 & 9274 & 9274 & 9274 & 9274 & 9274 & 9274 & 9272 \\
services & \\
\midrule
Customer care& 103 & 103 & 103 & 103 & 103 & 103 & 103 & 103 & 103 & 103 & 103 & 103 & 103 & 103 & 103 & 103 & 103 & 103 & 103 & 103 & 103 & 103 \\
interactions & \\
\midrule
Extempore & 2401 & 2401 & 2401 & 2401 & 2401 & 2401 & 2401 & 2401 & 2401 & 2401 & 2401 & 2401 & 2401 & 2401 & 2401 & 2401 & 1216 & 2401 & 2401 & 2401 & 2401 & 2401 \\
questions & \\
\midrule
Icebreaker & 152 & 152 & 152 & 152 & 152 & 152 & 152 & 152 & 152 & 152 & 152 & 152 & 152 & 152 & 152 & 152 & 145 & 152 & 152 & 152 & 152 & 152 \\
questions & \\
\midrule
Questions about & 7 & 7 & 7 & 7 & 7 & 7 & 7 & 7 & 7 & 7 & 7 & 7 & 7 & 7 & 7 & 7 & 7 & 7 & 7 & 7 & 7 & 7 \\
named entities & \\
\midrule
Role-play & 1185 & 1911 & 201 & 196 & 887 & 3256 & 1355 & 448 & 423 & 439 & 662 & 199 & 1488 & 426 & 2700 & 329 & 0000 & 245 & 425 & 1882 & 1804 & 2722 \\
scenarios & \\
\arrayrulecolor{green}\bottomrule\arrayrulecolor{black}
\end{tabular} \\
\caption{Count of different kind of prompts and sentences, prepared as a part of our data collection for all the 22 languages.}
\label{tab:prompts-table-counts}
\end{table}

\subsection{Role-play scenarios (WL3, WL4, WL6)}
In addition to single-speaker extempore data, it is also important to record natural, free-flowing conversations between two participants. However, spontaneously initiating dialogue between strangers is challenging. Indeed, in our initial pilots we found that pairing two participants lead to very mundane get-to-know-me style conversations involving ``How are you? What do you do? Where do you live?'' and so on. To elicit rich conversations, we crafted structured roleplay prompts, assigning specific roles (like Customer and Shopkeeper) and topics to encourage organic interaction. These prompts are categorized into \textit{general}, \textit{state-specific}, and \textit{district-specific} roleplays to add depth and cultural context to the conversations. For example, general roleplays might involve scenarios like "\textit{Customer and Shopkeeper bargaining over vegetable prices}". State-specific roleplays could include interactions such as ``\textit{Kashmiri artisan and local Kashmiri discussing the impact of industries on handcrafted items}'', while district-specific roleplays might delve into more localized contexts like ``\textit{Rice dealer and customer discussing types of rice native to Palakkad}''. 
In creating these conversational prompts, we manually curated extensive lists of occupation-based pairs, each associated with unique scenarios for brief yet meaningful 2-3 minute interactions. Providing participants with context-rich scenarios, results in more natural and authentic speaking style compared to extempore speech. Moreover, the scenarios are designed to cover a range of emotions from disappointment and excitement to appreciation and advice. To minimise repetition, we retire a prompt once $k$ conversations have been collected for it in a given language.

\begin{table}[!h]
\begin{center}
\includegraphics[width=\linewidth]{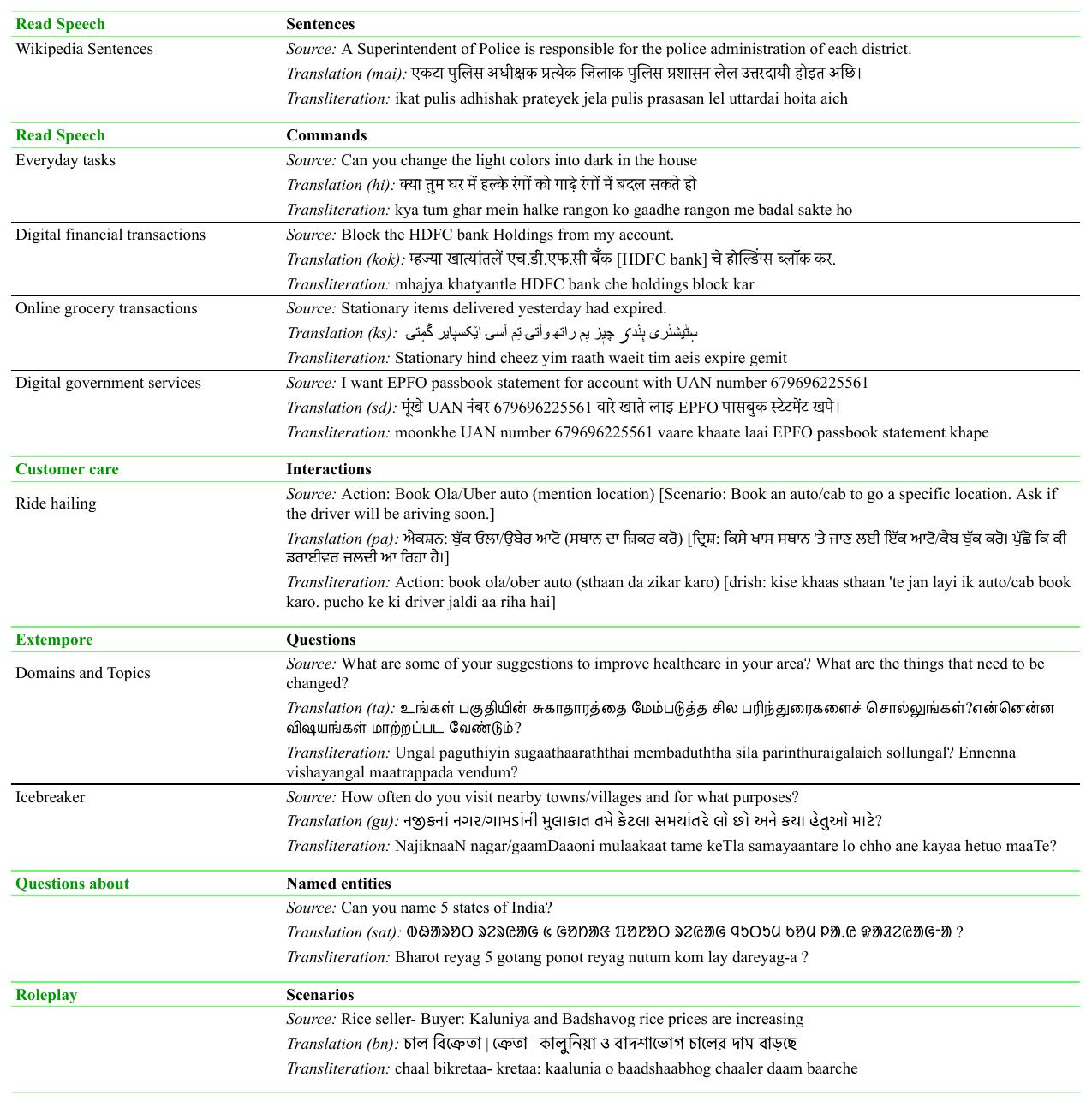}
\end{center}
\caption{Examples of different sentences, prompts, questions and role-pay scenarios created as a part of our pre-collection preparation. These are localised and colloquially translated to all the 22 languages.}
\label{tab:examples-tasks}
\end{table}

\noindent\textbf{Summary}
Table \ref{tab:prompts-table-counts} shows the number of sentences for read speech, number of instructions for different digital interactions, number of scenarios for customer care interactions, number of questions and role-play scenarios for two-party conversations. Table \ref{tab:examples-tasks} shows examples of the sentences, questions and prompts collected by us. This required a team of linguists experienced in fieldwork for eliciting meaningful responses and translators for colloquially translating and contextualising the material to all the 22 languages. All of this material will be released as a part of this work and can be adapted with minimal effort for data collection in other languages.

\section{On-field Data Collection}
We now describe the process of collecting data focusing on (i) process of building a countrywide team (ii) platform used for data collection and (iii) procedure used for collecting data.

\subsection{Countrywide Network}
As mentioned above, this work aims to collect data from 408 out of the 742 districts in India, covering the 22 scheduled languages. To do so, we need a team of district-specific influencers/mobilisers who are resourceful enough to source participants meeting our diversity criteria covering age, gender, educational backgrounds and professions as outlined in Section \ref{sec:wishlist}. Additionally, we need local coordinators who can be trained in our data collection method and be physically present alongside participants to ensure seamless collection. This required us to establish a network of collaborative partners with connections at the grassroot level, who could help in recruiting such local mobilisers and coordinators. Given the unique culture, geographical and demographic challenges of each language, a one-size-fits-all approach does not suffice. We thus had to evaluate several options by conducting small scale pilots in different parts of the country.

During these pilots we found that for languages such as Hindi, Tamil, Telugu, etc. that are widely spoken in more developed regions with prosperous urban centers, locating professional data collection agencies with the requisite resources and connections to mobilize participants and secure proficient coordinators is relatively straightforward. However, for languages such as Dogri, Manipuri, and Santali, which are predominantly spoken in remote and less-accessible areas, suitable commercial agencies often lack the necessary local presence and capabilities required to support effective data collection. For such languages, we identified three distinct categories of dependable partners: (i) foundations dedicated to the preservation and promotion of regional languages (ii) social sector professionals operating independently, with prior experience in on-field surveys and (iii) local universities with robust linguistics departments and prior involvement in research requiring fieldwork. Eventually, we decided to partner with data collection agencies for 10 languages, foundations for 5 languages, universities for 3 languages and social sector professionals for 4 languages.

\begin{figure}
    \centering
    \includegraphics[width=0.9\linewidth]{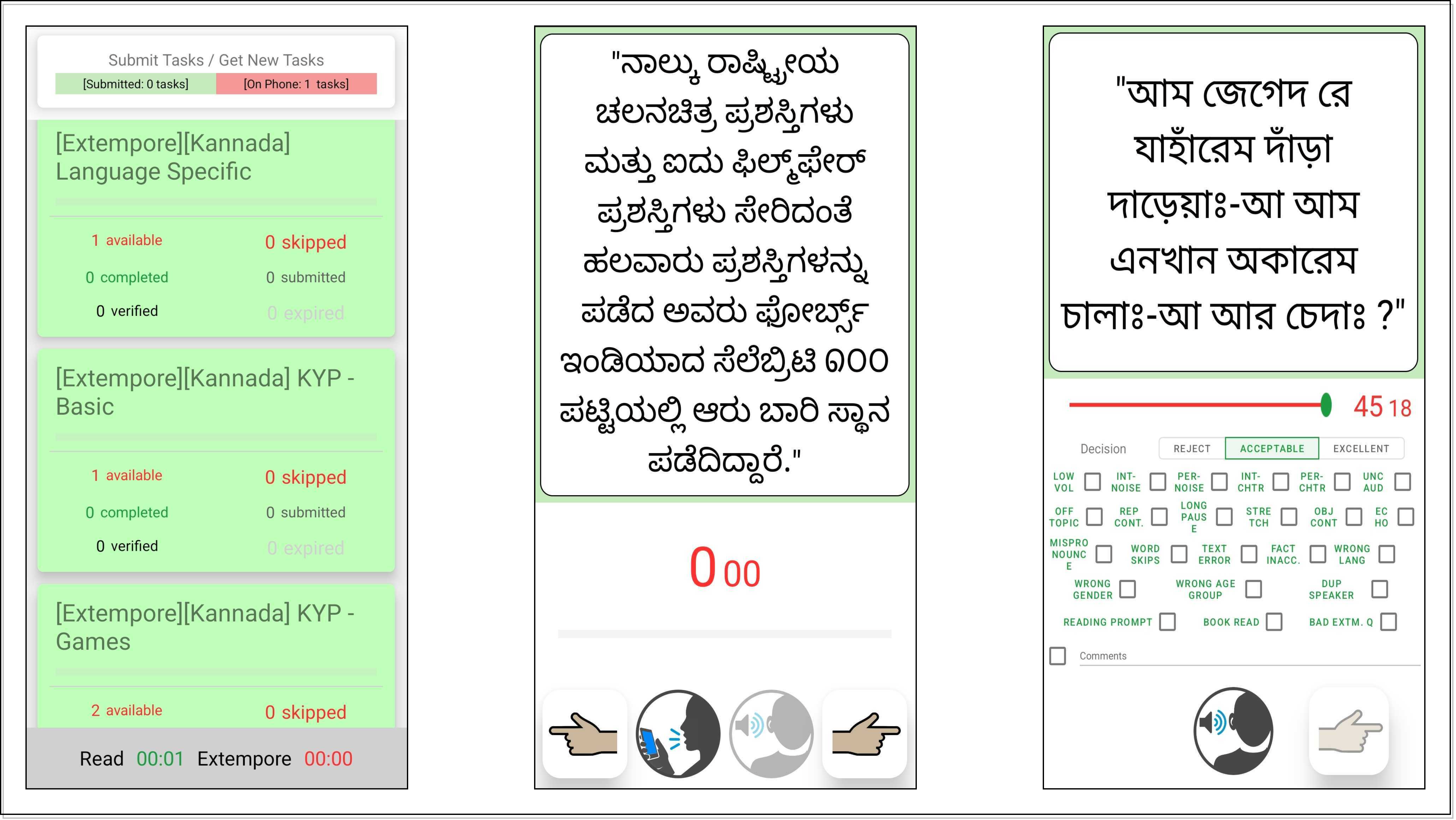}
    \caption{Karya's Home screen (left), micro-task screen (middle) and QC screen (right)}
    \label{fig:karya-screen}
\end{figure}

\subsection{Platform for data collection}
To manage data collection in a remote set up distributed across the country, we needed a tool that could help us standardise the entire procedure. For this, we chose Microsoft’s open source crowd-sourcing platform called Karya which has been used by other teams for collecting speech data \cite{Svarah,10.1609/aaai.v37i11.26521,karya}. Karya is available as an open source Android application, which can operate even in offline environments and synchronize with a backend server periodically, to upload accumulated data. This makes it particularly well-suited for Indian settings where uninterrupted connectivity may at times be a challenge.

The app allows the users to perform micro-tasks where a micro-task could involve reading a sentence, responding to a question, or enacting a scenario. For example, Figure \ref{fig:karya-screen} show a micro-task which requires the participant to read a prompt. The participant can see the prompt and then press the record button to start recording their response. Once they are done, they can press the stop button and listen to their response once. If they are satisfied with the response they can submit it, else they can record it again. Once submitted, the audio file of their recording gets stored in .wav format. We made a few changes to the base version of the app to (i) display micro-tasks grouped into categories defined in section \ref{subsec:3.1} to \ref{subsec:3.6} (ii) display the total recording duration and (iii) display hints for questions to avoid repetitive answers (See section \ref{sec:extempore-questions}).

While the mobile app was used to collect data from a single participant over a 44.1kHz channel, to enable conversations between two participants over a 8kHz channel we used a cloud telephony service provider. More specifically, we set up a telephone bridge and connected two participants on a single call and recorded their conversation.

\subsection{Mobilisation and Training}
In every district, we recruit 1-2 mobiliser(s) with the help of our local partners. These mobilisers are local influencers who help to spread awareness about the data collection effort and clearly inform potential participants about the compensation involved and the purposes for which the data would be used. In addition, we also hired 5-7 coordinators in each district. These were typically young graduates who were native speakers of the language and were good at interpersonal skills so that they could make participants feel comfortable and elicit good responses from them. We also ensured that roughly half the coordinators were females to ensure that female participants felt comfortable. We trained the coordinators in using the Karya app and explained the various micro-tasks. We also explained the acceptance/rejection criteria for responses as outlined in our quality control guidelines (Section \ref{sec:QC}). This helped them guide the participants and get acceptable responses. 

\subsection{Collection Procedure}
Based on regional factors, we considered two options (i) renting a centrally located space in the district where the participants could assemble or (ii) visiting participants at their home in case they have any reluctance/restrictions in traveling. In some districts, we used both the options. 

\noindent{\textbf{Onboarding participants:}} Once a willing participant has been identified by the mobiliser and contacted by the coordinator (either at the central assembly point or at their home), we start the process of onboarding the participant. As a first step, the coordinator again explains the task, the purpose of data collection, the amount of time it will take and the compensation involved. The coordinator then asks for their consent by requesting them to sign a consent form. The coordinator then helps the participant to fill a registration form capturing phone, name, age, gender, primary language, language proficiency, occupation, location, education, domains and topics of interest. The form also captures the coordinator's id for tracking and accountability and is finally submitted along with the soft copy of the consent form. Capturing this information is important so that the tasks shown in Karya match the interests of the participants. For example, if the participant selects ``Politics'' as a topic of interest then they will be asked questions related to politics during the extempore conversations. Similarly, if the participant states their language proficiency as speak-only, then they will not be shown any read speech tasks. Lastly, the meta-data about age, gender, education, occupation, etc. helps us in ensuring that we meet all the diversity criteria in our wishlist.

\noindent{\textbf{Assigning tasks:}} Once the participant is onboarded, the coordinator helps them in installing the Karya app on their phone. We prefer using the participant's phone for recording as this allows us to cover a wider range of devices and microphones of varying quality. The participant then logs into the app using a unique access code that was generated during the onboarding process. Once the participant logs in, the app connects to the backend and assigns a list of tasks based on the preferences shared by the participant in the registration form. Table \ref{tab:assigning_tasks} summarizes the list of tasks seen by a participant.

\begin{table}[!h]
\begin{center}
\scriptsize
\begin{tabularx}{\textwidth}[t]{sz}
\arrayrulecolor{green}\toprule
\textbf{Task} & \textbf{Description}\\
\arrayrulecolor{green}\midrule
Sentences for read speech & 4 sentences from Wikipedia that need to be read as it is. \\
\arrayrulecolor{black}\midrule
Everyday tasks & 4 commands used for interaction with in-home personal assistant. \\
\arrayrulecolor{black}\midrule
Digital financial transactions & 4 interactions that are typically encountered in digital transactions \\
\arrayrulecolor{black}\midrule
Online grocery transactions & 4 interactions that are typically encountered in such transactions \\
\arrayrulecolor{black}\midrule
Digital government services & 4 interactions that are typically encountered while interacting with such services \\
\arrayrulecolor{black}\midrule
Keywords & 10 keywords to be spoken as it is by the participant \\
\arrayrulecolor{black}\midrule
Customer care interactions & 2 scenarios to enact, one each from ride hailing and food delivery \\
\arrayrulecolor{black}\midrule
Extempore questions & 4 questions, two each from the participant's interested  domain and topic of interest. The task requires participants to answer the questions in their natural way. \\
\arrayrulecolor{black}\midrule
Icebreaker questions & 3 questions, one from each tier, viz. warm-up, everyday-life and questions on participant's mother-tongue. \\
\arrayrulecolor{black}\midrule
Product reviews & One positive review for a product that they have purchased recently and one negative review for a product that they have purchased recently. \\
\arrayrulecolor{black}\midrule
Questions about named entities &  7 questions asking about entities like names of countries, international cities etc. \\
\arrayrulecolor{black}\midrule
Role-play scenarios &  3 scenarios for conversations between two participants, one each from general, state-specific and district-specific. These are not done through the app but by connecting two participants on a telephone call through a bridge. \textbf{Note} that it is not necessary that the same person gets paired with the same partner for all the 3 conversations. \\
\arrayrulecolor{green}\bottomrule

\end{tabularx}
\end{center}
\caption{Examples of different sentences, prompts, questions and role-pay scenarios created as a part of our pre-collection preparation. These are localised and colloquially translated to all the 22 languages.}
\label{tab:assigning_tasks}

\end{table}

\noindent{\textbf{Recording data:}} Each of the above is a separate micro-task. For example, the 4 Wikipedia sentences correspond to 4 different micro-tasks, each recorded, verified and submitted independently. Similarly, the 4 commands for the everyday tasks correspond to 4 independent micro-tasks and so on. The coordinator is physically present with the participant during the entire procedure. Apart from guiding the participant through the app, the coordinator also does the level 0 quality control. In particular, once the participant records a response for a given micro-task, the coordinator checks if (i) the response is audible despite ambient noise (ii) for read speech, the response matches the given text verbatim (iii) for questions/prompts, the response is relevant to the question/prompt being displayed and (iv) the response is of sufficient length but is not unnecessarily repetitive or stretched. The participant is requested to repeat the micro-task if the response is not found to be satisfactory. The coordinator also ensures that the participant is comfortable through the entire process and takes a break whenever required.

\noindent{\textbf{Logging out:}} 
Throughout the session, the coordinator keeps monitoring the amount of data recorded by the participant, as displayed on the home screen of the app. We mandate at least 20 minutes of recorded data per participant and to account for rejection during QC, the coordinator ensures that at least 25-35 minutes gets recorded per participant. Once all the tasks are finished the coordinator ensures that all the data gets uploaded to the backend server as confirmed by a message displayed on the app. The participant is then requested to delete the app so that it does not consume any space on their device. The entire collection procedure would takes anywhere between 1-4 hours depending on the participant. We would count this as half-day's work and pay the participant accordingly based on prevailing half day wages in that region and as agreeable to the participant.

\section{Quality Control}
\label{sec:QC}
It was important to have a centralised quality control team to ensure integrity and consistency of data being collected by local partners in remote and widespread areas of the country. To do so, we set up an in-house quality control team comprising of 3-5 experts per language. This team was responsible for verifying the meta-data as well as the audio files as described below. 
\begin{figure}
    \centering
    \includegraphics[width=\linewidth]{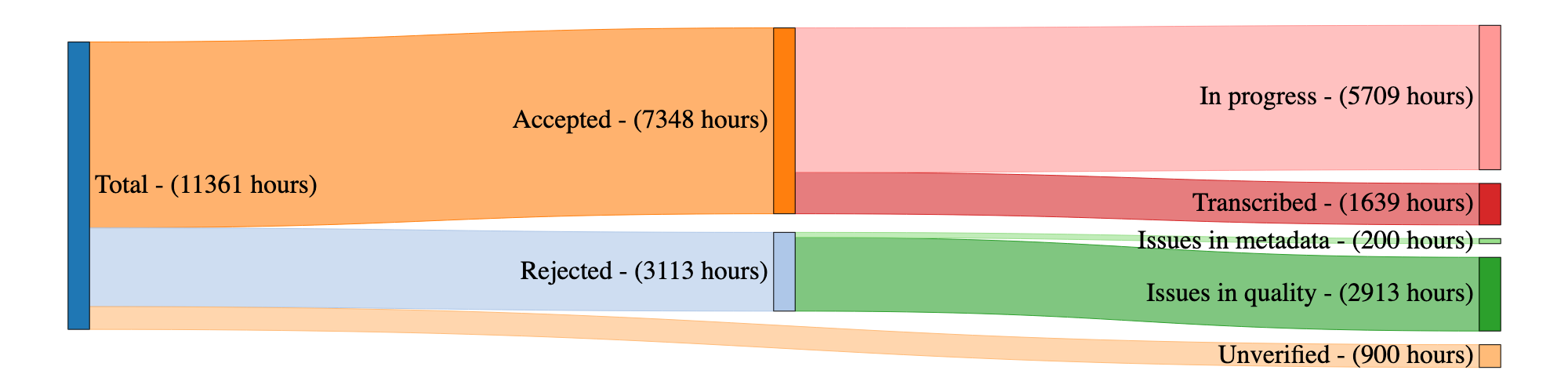}
    \caption{Number of hours of data at different stages in the collection pipeline}
    \label{fig:sankey}
\end{figure}

\begin{figure}[!t]
    \centering
    \includegraphics[width=0.8\linewidth]{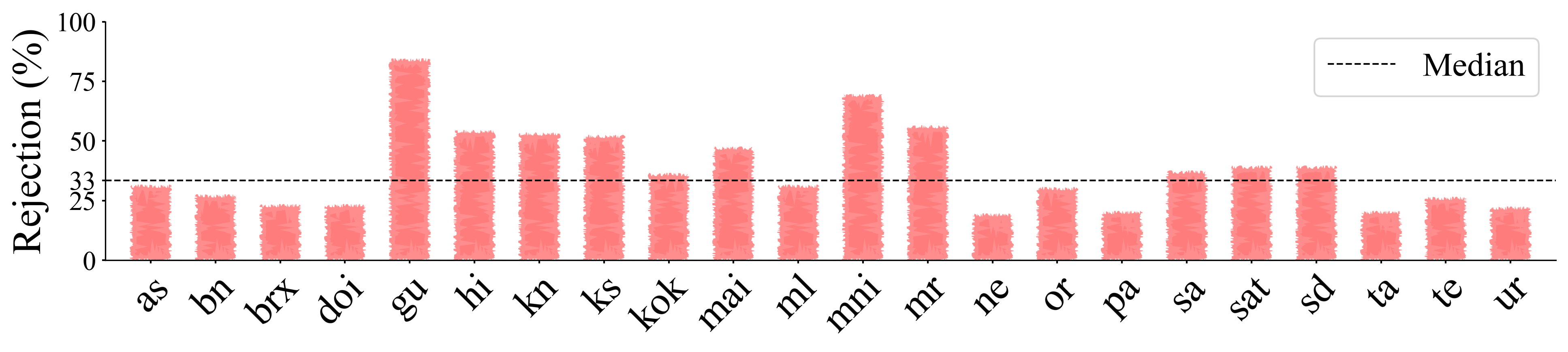}
    \includegraphics[width=0.8\linewidth]{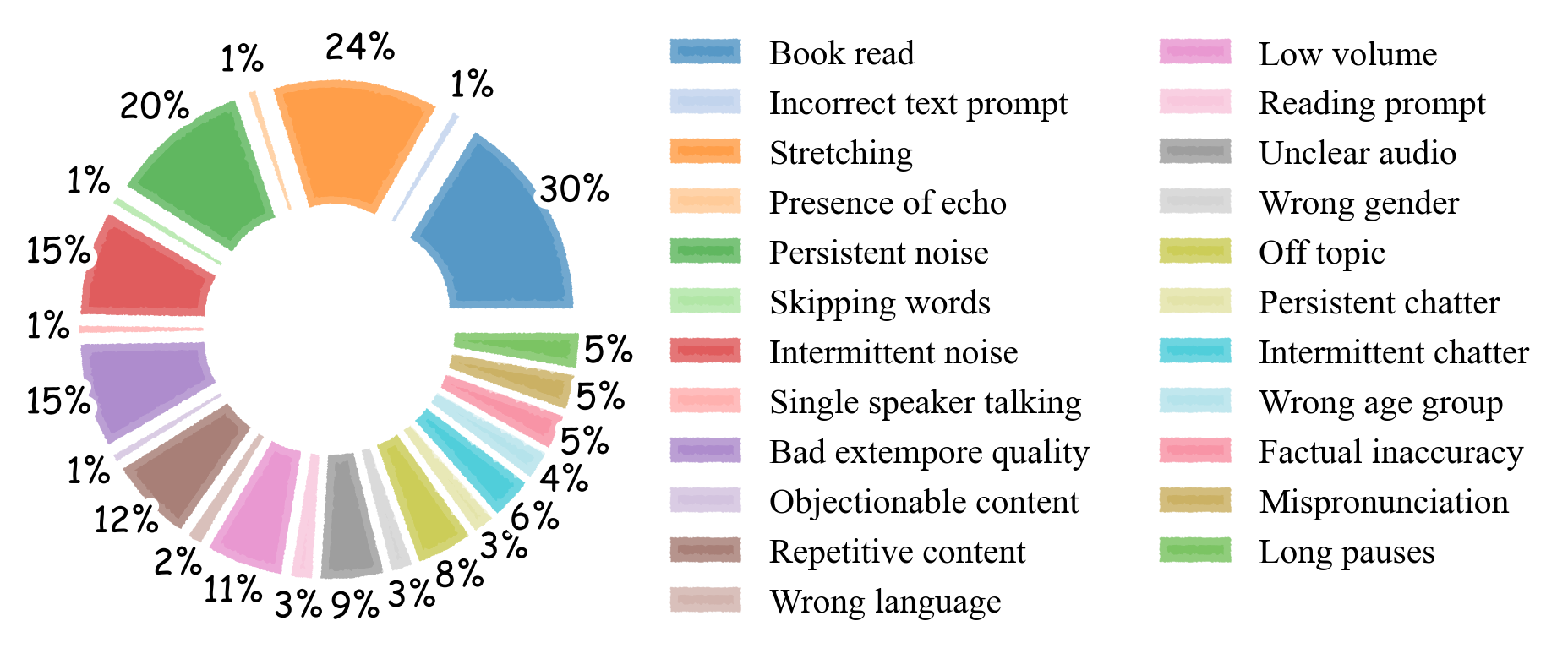}
    \caption{\textbf{Top:} Percentage of data that was rejected by our QC team for each language. The higher numbers are for languages where multiple pilots were done before we could find the right local partners. \textbf{Bottom:} Distribution of errors across the 23 categories.}
    \label{fig:rejection-percentage}
\end{figure}

\noindent\textbf{Verifying meta-data:} During the initial stages of data collection, some of our team members raised a concern that the voice of a participant did not match the corresponding age and/or gender information. On further investigation, we found that indeed in some cases this information was being entered incorrectly intentionally (to cheat and meet the criteria for demographic diversity) or unintentionally (human error while filling the registration form). We also found that in some cases, the voice was not consistent across audio files corresponding to the same participant, indicating that multiple participants were contributing data under the same participant id (perhaps because the original participant had to leave and a family member continued in their place). Lastly, we also found some cases where the same voice was heard across different participant ids, indicating that the same individual was posing as different participants. 

To mitigate these issues related to authenticity of participant's identity, we added an additional micro-task where the participant would record a video speaking some trivial content such as ``I am hear to contribute to this initiative. I live in district X and my mother tongue is Y''. Our in-house QC verifies the age and gender information entered by the participant by comparing it with their appearance in the video. If the age or gender doesn't match then this participant's data is rejected (of course, the age can only be approximately inferred from the video and hence the participant was rejected only if there was a huge disparity). The QC team also rejects a participant's data if the voice of the participant as heard in this video does not match the voice in the audio samples contributed by this participant. Further, the QC team also flags cases where the same participant appears in multiple videos using different ids. All video verification tasks were done only by female members of the QC team. To maintain privacy, these videos could not be downloaded or shared in any away and were deleted once the verification was done. Some female participants were not comfortable in recording a video. Respecting these cultural sensitivities, in such cases, we did not force them to record the video but instead connected them to a female member from our QC team for live verification over a WhatsApp call. 

\noindent\textbf{Verifying diversity criteria:} For each language, one member from the in-house QC team was responsible for checking that for every district we meet all the diversity criteria outlined in Section \ref{sec:wishlist}. In case of a gap, the local partners were asked to source participants from specific underrepresented categories. The QC team  also ensured that there was enough diversity in the domains and topics across participants. For example, in the initial pilots, we saw that participants often pick up easy topics such as ``Sports'' and ``Entertainment''. To avoid this and to ensure that all the 21 \textit{domains} and 28 \textit{topics of interest} have some representation in the data, the QC team would inform the local partners to source participants who could speak about specific domains and topics which were under-represented in the given district. Lastly, the QC team would also monitor that easy conversation scenarios such as ``teacher-student'', ``customer-shopkeeper'' were not repeated by disabling such options in the registration form.

\noindent\textbf{Verifying content:}
In the early stages of data collection, we found that despite the level 0 QC by the coordinator, there were still some issues with the audio files. For example, some audio files had very low volume making it difficult to understand what was being said. Similarly, some audio files had constant loud background noise making it impossible to hear the participant. 
We hypothesize that certain audio samples might bypass the initial quality control (level 0 QC) due to potential oversights. These oversights may occur if the coordinator is not vigilant or because being directly engaged in the conversation, the coordinator can comprehend the audio content despite challenges like low volume or significant background noise. To identify and reject such audio files, our in-house QC team first analysed several batches of audio files and came up with a exhaustive list of error categories as listed in Table \ref{apx:tab:error-categories-table} in Appendix \ref{apx:qc-noise-tags}. Every audio file collected from the field was then verified and marked with one of the following categories: Excellent, Acceptable (has some errors but a human could still transcribe it), NotAcceptable (has some errors and even a human could not transcribe it). For audio files falling into the last two categories, the QC team  identified and marked each specific error present in the file, ensuring a comprehensive record of \textit{all} issues detected. Figure \ref{fig:rejection-percentage} shows the percentage of data that was rejected for each language and the distribution of errors across different categories.
 
\section{Transcription}
We now discuss the process of transcription, focusing on audio segmentation, transcription guidelines and transcription workflow.

\begin{figure}
    \centering
    \includegraphics[width=\linewidth]{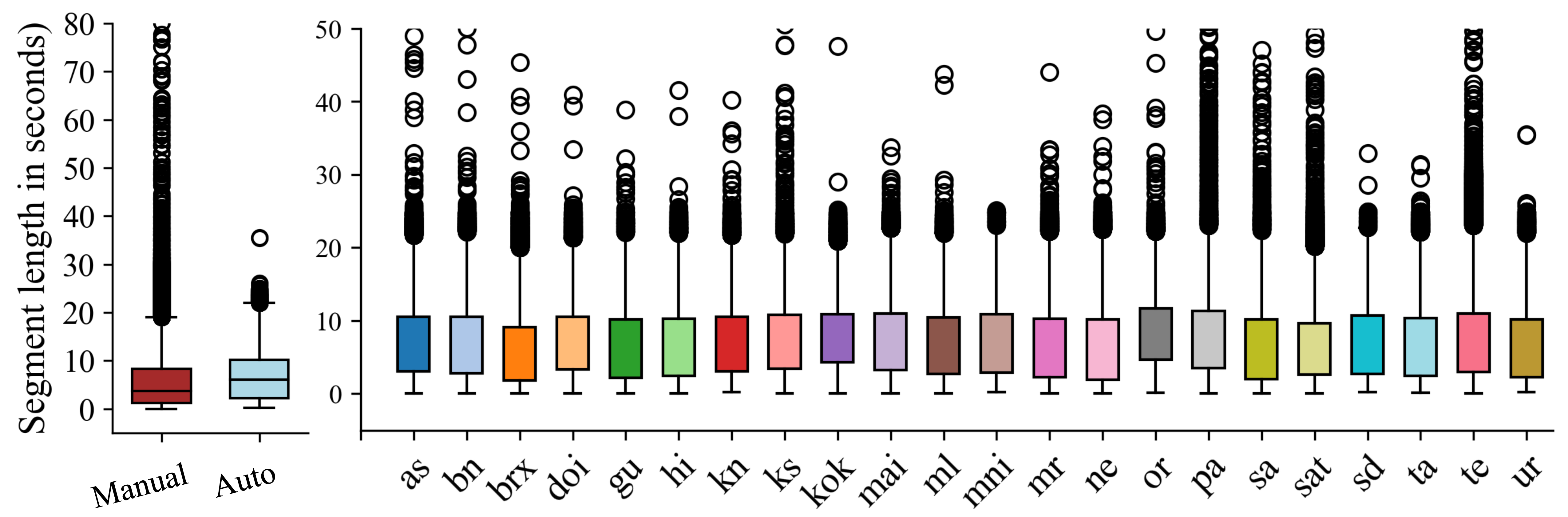}
    \caption{\textbf{Left:} Comparison of the distribution of segment lengths generated by manual segmentation and automated segmentation. On average, the manual segmentation process leads in very short segments (small mean) but also in results some excessively large segments. \textbf{Right:} With the automated approach we are able to control the distribution of the length of audio segments ensuring a similar average length across languages and very few outliers. The outliers typically correspond to very fast spoken content where pauses are hard to identify.}
    \label{fig:segment_lengths}
\end{figure}

\noindent\textbf{Audio segmentation:} Our data includes read, extempore, and conversational audio data, with read speech typically lasting 6 to 8 seconds, and extempore and conversational data extending up to several minutes. The extended duration necessitated segmentation due to the computational constraints of Automatic Speech Recognition (ASR) systems, optimized for 20 to 30-second audio contexts.  Initially, we relied on annotators to listen to the entire audio and then delineate logical segments using an annotation tool. However, defining what constitutes a ``logical segment'' proved ambiguous, leading to considerable variability in the segmentation outcomes, with some annotators creating excessively short or long segments (as shown in Figure \ref{fig:segment_lengths}). 

To standardize segmentation and reduce subjectivity, we shifted to an automated approach using Silero Voice Activity Detection (VAD \cite{SileroVAD}. This method determines voiced segment boundaries based on a specified minimum silence duration ($S_{min}$). The selection of $S_{min}$ directly influences segment length, with shorter $S_{min}$ values generating more segments by identifying every minor pause, and longer $S_{min}$ values requiring more substantial pauses for segmentation. We followed an iterative approach, wherein we varied the value of $S_{min}$ from 1000 ms to 40 ms. During each iteration, we retained segments which were less that 20 seconds and removed them from further processing. In subsequent iterations, longer audio files were further segmented by relaxing the minimum silence duration. Further, to prevent skew towards brief segments, we merged shorter segments, achieving an average segment length of 9 seconds. Figure \ref{fig:segment_lengths} shows the distribution of the segments length across languages when using the automated approach.

\begin{table}
    \centering
    \includegraphics[width=\linewidth]{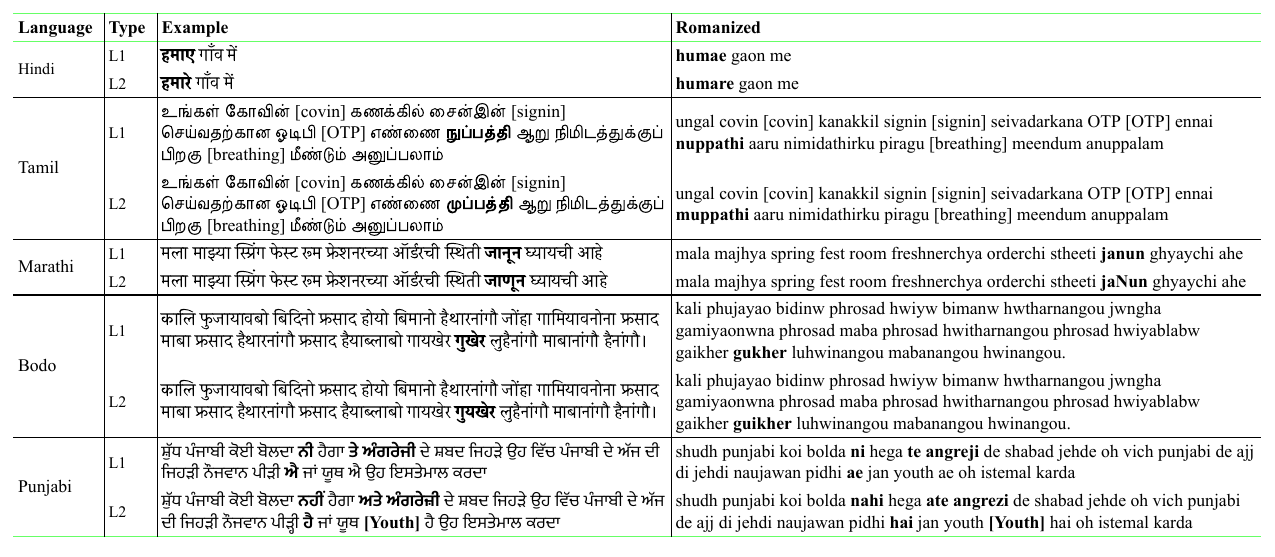}
    \caption{Examples demonstrating differences between L1 and L2 transcripts.}
    \label{tab:l1l2examples}
\end{table}

\noindent\textbf{Transcription guidelines:} A significant challenge that we encountered early on was the absence of clear transcription guidelines for Indian languages. This was further exacerbated by the divergence between spoken and written forms due to colloquialisms, accent variations, mispronunciations, and contractions in rapid everyday speech, as shown in Table \ref{tab:l1l2examples}. This issue necessitated a decision between two transcription approaches: (i) verbatim transcription, capturing spoken language as it is, despite deviations from standard written forms, or (ii) standardized transcription, ensuring that all transcriptions conform to textbook representations, regardless of spoken variations. For instance, the Hindi word ``mujhe'', where the correct phoneme is an aspirated 'jha', is often lazily pronounced as ``muje'' with an unaspirated 'ja' in everyday speech. Transcribing the spoken "muje" as "mujhe" may not align with the actual pronunciation, causing dissatisfaction among those prioritizing speech-text coherence. Conversely, transcribing it as "muje" could frustrate those expecting textbook spellings, as it may not be recognized in dictionaries or by downstream applications.

To address this dilemma, we adopted a two-level transcription strategy. In Level 1, transcription is performed exactly as spoken, aligning with those emphasizing verbatim transcriptions. Level 2 involves standardizing words to their textbook form, satisfying both spelling purists and practical users. Given the high correspondence between grapheme and phonemes sets for Indian languages, coming up with Level 1 transcription guidelines was relatively easier as the transcribers just need to listen to the sound and transcribe it using the corresponding grapheme. An additional guideline here was to add standard tags such as [\textit{baby\_crying}], [\textit{barking}], [\textit{hmm}], [\textit{horn}], etc. We had a total of 89 such tags, as mentioned in Table \ref{apx:tab:noise-tags-table} in Appendix \ref{apx:qc-noise-tags}. However, Level 2 guidelines required more careful consideration, for which we collaborated with linguists to develop detailed rules for standardization. For each of the 22 languages, we crafted guidelines incorporating examples from the collected audio, undergoing several rounds of review to achieve consensus and finalize the guidelines, as detailed in Appendix \ref{apx-transcription-guidelines}.

\noindent\textbf{Transcription workflow:} To ensure quality we used a maker-checker-superchecker workflow, \textit{i.e.}, the initial transcription was followed by two rounds of review. For this, we hired a team of transcribers comprising of makers, checkers and super-checkers. There are a total of 700 transcribers involved in the process, most of them working remotely, and some even stationed inside data collection agencies. The super-checkers are in-house team members who are experts in the language and have vast experience in various language tasks such as creative writing, translation and transcription. To manage the workflow across partners and languages, we use an in-house annotation platform called Shoonya \cite{shoonya} which supports all the 22 languages and allows us to assign, monitor and track tasks and set up a maker-checker-superchecker workflow.

We posit that Level 2 transcription demands certain expertise and should be undertaken by specialists. Therefore, the initial transcription (Level 1) is conducted by transcribers and reviewers, while super-checkers perform a secondary review of Level 1 transcriptions and execute the Level 2 standardization process. 

\section{IndicVoices}
\begin{figure}[!t]
    \centering
    \includegraphics[width=\linewidth]{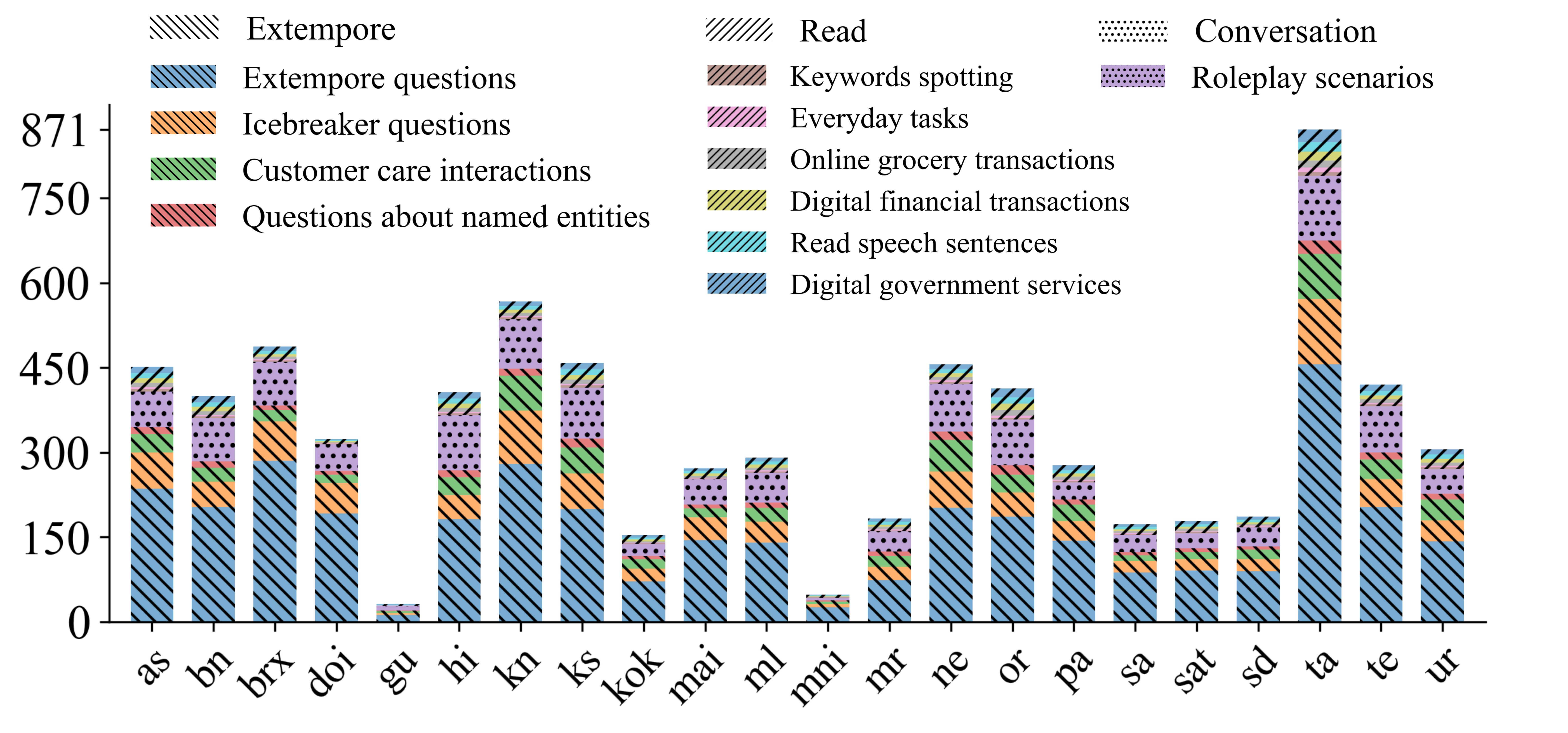}
    \caption{No. of hours of Read, Extempore and Conversation speech collected for each language along with splits for each sub-category.}
    \label{fig:re_ext_conv_hours}
\end{figure}

\begin{figure}[!b]
    \centering
    \includegraphics[width=\linewidth]{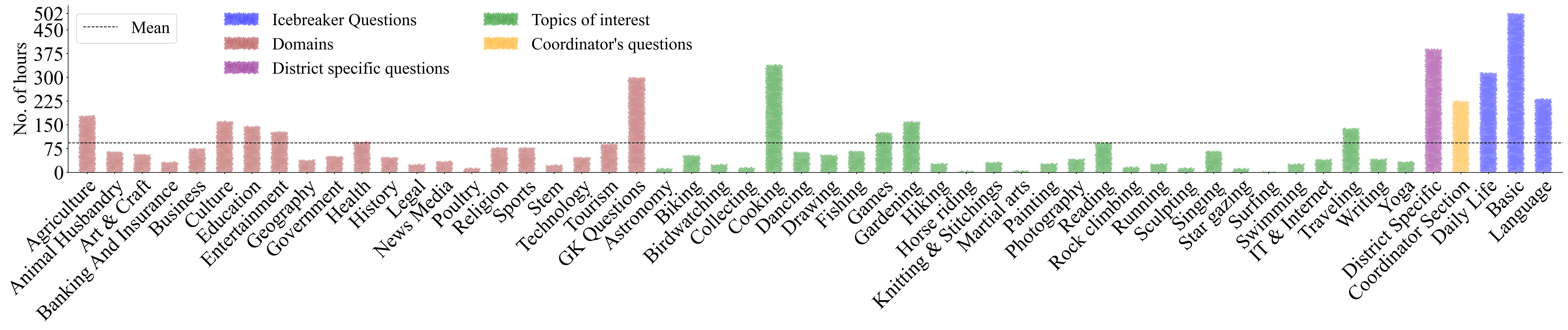}
    \caption{No. of hours of audio data collected from each of the domains and topics of interest summed up across languages. All domains have a good minimum representation ensuring high diversity in the collected data.}
    \label{fig:diversity-qa}
\end{figure}

\begin{figure}
    \centering
    \includegraphics[width=\linewidth]{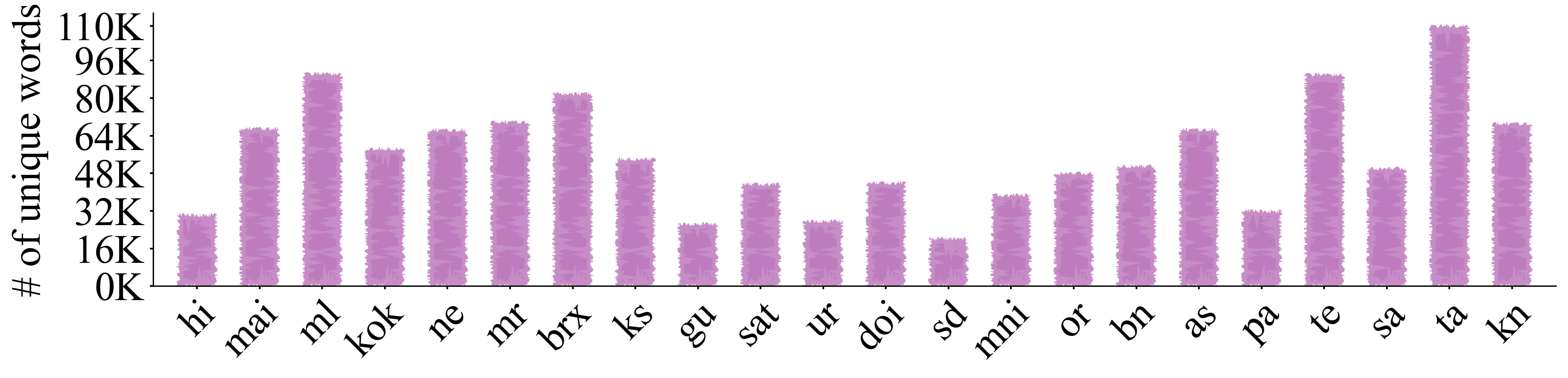}
    \caption{No. of unique tokens in the transcribed data across languages. The smaller bars correspond to languages for which transcribed data is very less.}
    \label{fig:unique_tokens}
\end{figure}

\begin{figure}%
    \centering
    \subfloat[]{{\includegraphics[width=6.5cm]{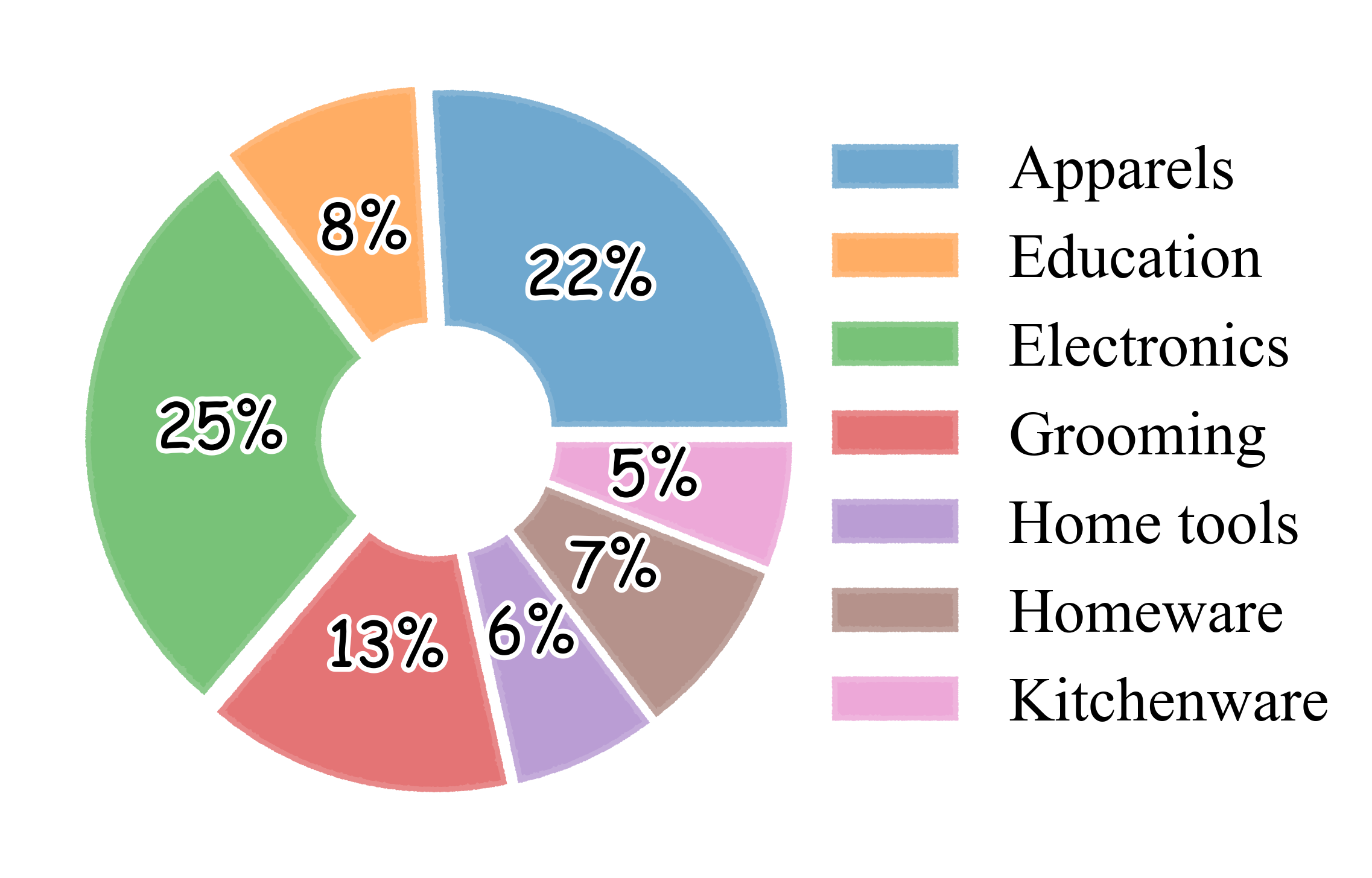} }}%
    \qquad
    \subfloat[]{{\includegraphics[width=6.5cm]{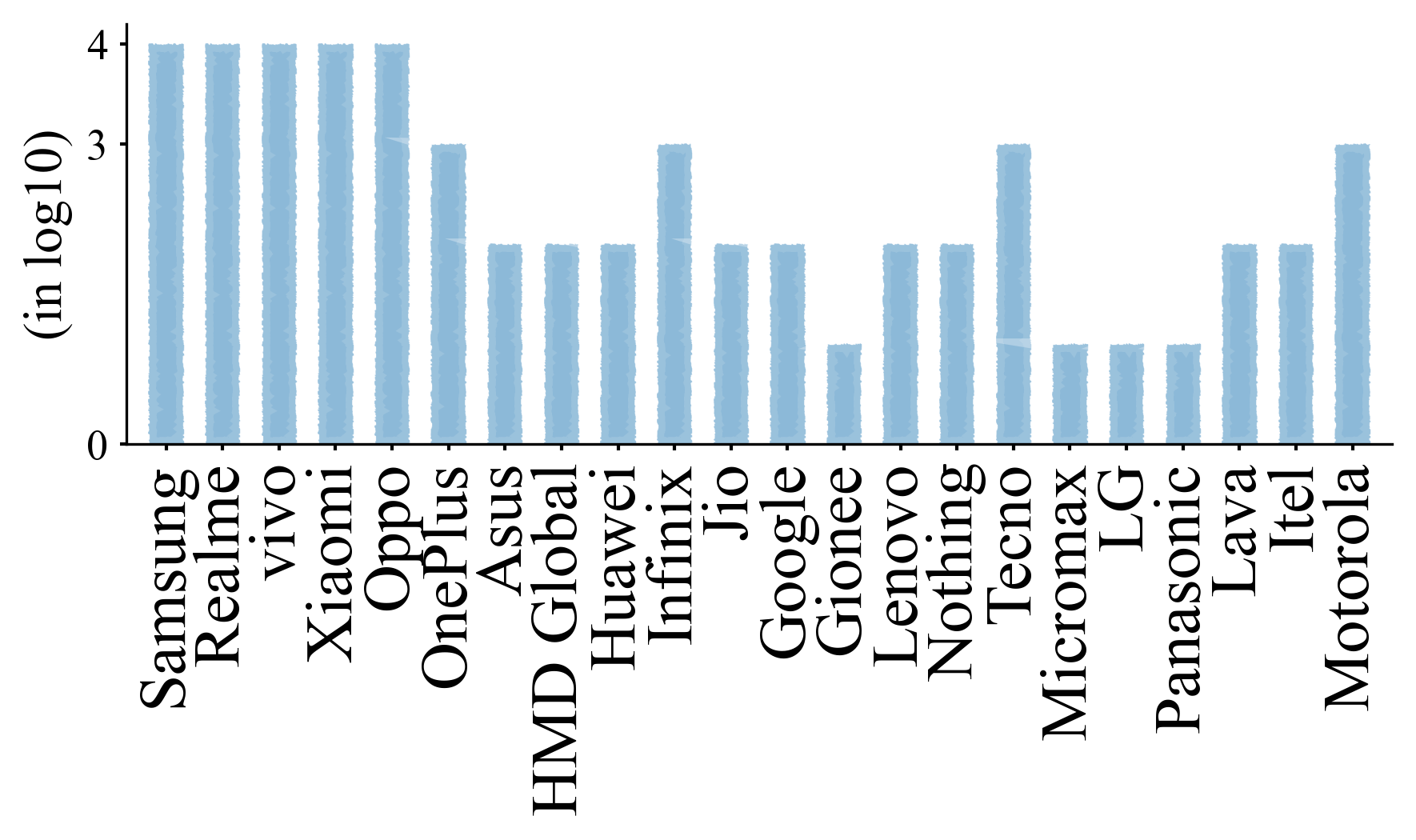} }}%
    \caption{The LHS plot shows different categories in the product reviews spoken by participants. The RHS plot shows the number of recording devices from different brands that were used in data collection.}%
    \label{fig:mobile-and-product}%
\end{figure}

\begin{figure}
    \centering
    \includegraphics[width=\linewidth]{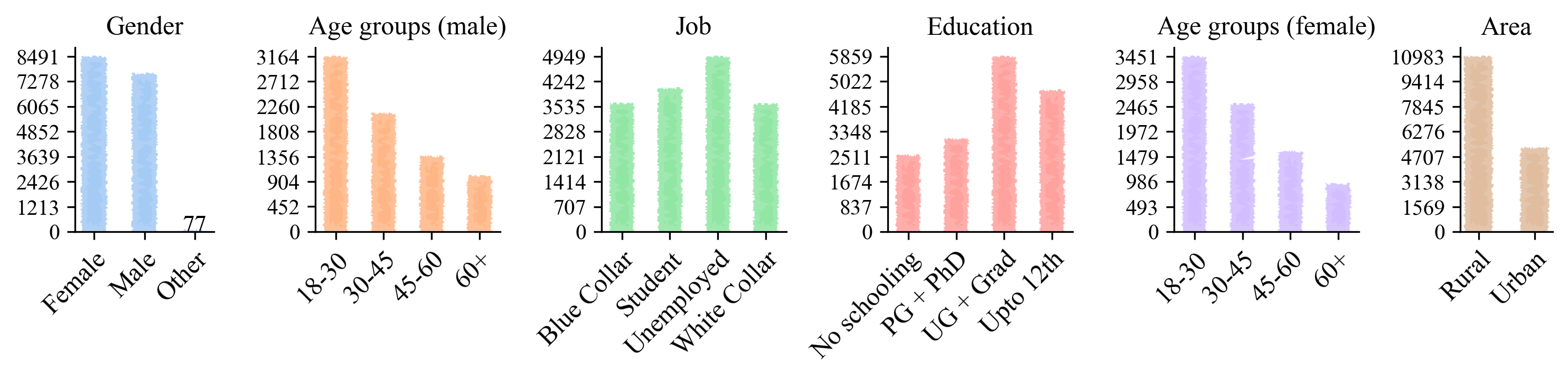}
    \caption{Demographic distribution of participants across categories summed up across all 22 languages.}
    \label{fig:demographics-avg}
\end{figure}

We collected a total of 7348 hours of data summed up across the 22 languages, covering 145 districts. Of these, 1639 hours of data have already been transcribed with the rest in the pipeline. Figures \ref{fig:re_ext_conv_hours} to \ref{fig:demographics-avg} summarise different statistics of the collected data which showcase its rich diversity. 

\begin{figure}
    \centering
    \includegraphics[width=\linewidth]{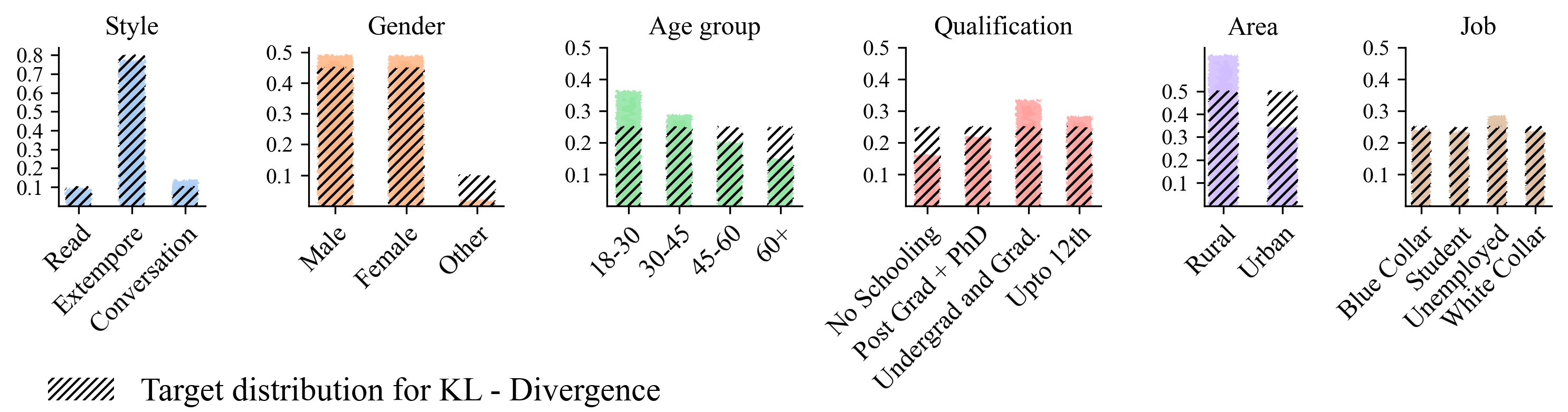}
    \includegraphics[width=\linewidth]{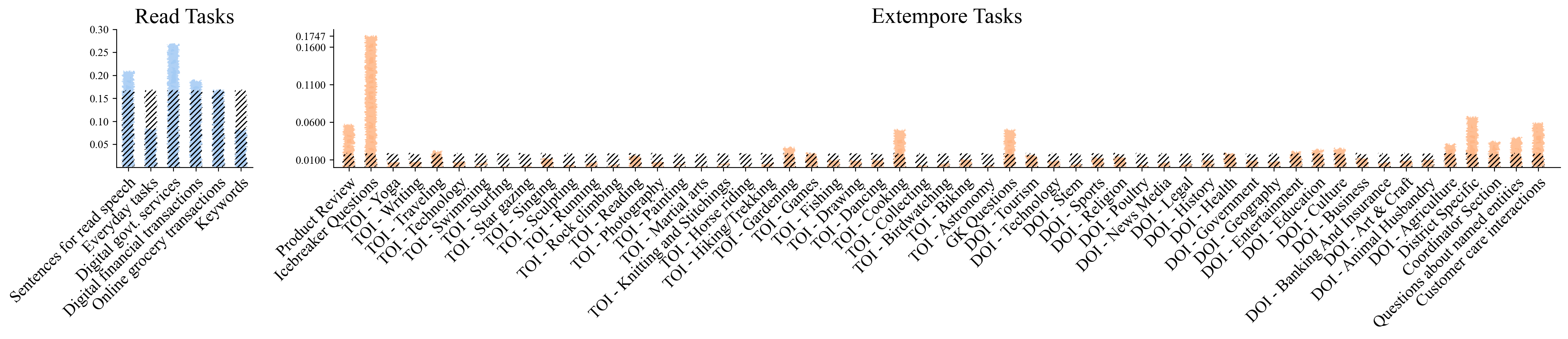}
    \caption{Difference between the desired distribution for each of the criteria considered and the distribution of the actual collected data.}
    \label{fig:kld}
\end{figure}

\begin{table}[!b]
\centering
\scriptsize
\begin{tabular}{l|ccccc||l|ccccc}
\arrayrulecolor{green}\toprule\arrayrulecolor{black}
\multicolumn{1}{l|}{\multirow{2}{*}{\textbf{Lang}}} & \multicolumn{2}{c}{\textbf{Train}} & \multicolumn{2}{c}{\textbf{Test}} & \multicolumn{1}{c||}{\multirow{2}{*}{\textbf{Valid}}} & \multicolumn{1}{l}{\multirow{2}{*}{\textbf{Lang}}} & \multicolumn{2}{c}{\textbf{Train}} & \multicolumn{2}{c}{\textbf{Test}} & \multicolumn{1}{c}{\multirow{2}{*}{\textbf{Valid}}} \\ 
\multicolumn{1}{l|}{} & \textbf{\#sp} & \multicolumn{1}{l}{\textbf{\#h}} & \textbf{\#sp} & \multicolumn{1}{l}{\textbf{\#h}} & \multicolumn{1}{l||}{} & \multicolumn{1}{l}{} & \textbf{\#sp} & \multicolumn{1}{l}{\textbf{\#h}} & \textbf{\#sp} & \multicolumn{1}{l}{\textbf{\#h}} & \multicolumn{1}{l}{} \\
\arrayrulecolor{green}\midrule\arrayrulecolor{black}
as & 684 & 100 & 163 & 5 & 1 & mni & 88 & 15 & 19 & 4 & 1 \\
bn & 590 & 100 & 125 & 5 & 1 & mr & 407 & 85 & 100 & 5 & 1 \\
brx & 694 & 100 & 164 & 5 & 1 & ne & 496 & 100 & 114 & 5 & 1 \\
doi & 321 & 65 & 84 & 5 & 1 & or & 391 & 75 & 92 & 5 & 1 \\
gu & 71 & 15 & 14 & 5 & 1 & pa & 274 & 71 & 61 & 5 & 0.4 \\
hi & 287 & 65 & 63 & 5 & 1 & sa & 176 & 45 & 33 & 5 & 1 \\
kn & 390 & 50 & 112 & 5 & 1 & sat & 237 & 86 & 53 & 5 & 1 \\
ks & 317 & 50 & 86 & 5 & 1 & sd & 108 & 10 & 30 & 5 & 1 \\
kok & 234 & 55 & 54 & 5 & 1 & ta & 863 & 100 & 195 & 5 & 1 \\
mai & 421 & 100 & 101 & 5 & 1 & te & 477 & 102 & 107 & 5 & 1 \\
ml & 332 & 60 & 67 & 5 & 1 & ur & 491 & 65 & 99 & 5 & 1 \\
\arrayrulecolor{green}\bottomrule\arrayrulecolor{black}
\end{tabular}
\caption{Number of speakers (\textbf{\#sp}) and hours (\textbf{\#h}) in the train, validation and test and splits across languages.}
\label{tab:train-testsplit}
\end{table}

\noindent\textbf{Creating train-test splits:} 
While the collected data can be used for training models for various speech tasks, in this work we focus on releasing standardised train-test splits for automatic speech recognition. While creating train-test splits, we need to adhere to a few basic principles (i) the test set should not contain any speakers which are already seen in the training data (ii) the test set should have the desired distribution of age, gender, education levels, professions, etc. as outlined in section \ref{sec:wishlist} (iii) the test set should have fair representation of speakers from every district and (iv) the test set should keep growing as more data gets collected in later phases of the project to allow for representation from newer districts, while adhering to conditions listed above. Creating such a balanced test set is challenging as it requires satisfying multiple demographic constraints simultaneously.

To efficiently manage this complexity, we create train-test splits using a sampling based method. More specifically, from the data collected for a given district, we draw a large number of samples that include all data from a random set of 20\% of speakers from that district (referred to as test speakers). Each such sample may have a different set of speakers with its own demographic distribution. We then evaluate the demographic distribution of each sample, selecting the sample whose demographic distribution has the highest entropy and the lowest Kullback-Leibler (KL) divergence from the target demographic profile. We ensure that the data for none of the speakers in the chosen sample is present in the training data. Figure \ref{fig:kld} shows the difference between the desired distribution for each of the criteria considered and the distribution of the actual collected data for that district. Our process thus ensures a fair and robust benchmark ensuring speaker exclusivity between train-test splits, inclusion of all districts and a balanced demographic distribution, to the extent possible.

\begin{table}[h]
\scriptsize
\centering
\begin{tabular}
{@{}l@{\hspace{0.2em}}|@{\hspace{0.2em}}c@{\hspace{0.56em}}c@{\hspace{0.56em}}c@{\hspace{0.56em}}c@{\hspace{0.56em}}c@{\hspace{0.56em}}c@{\hspace{0.56em}}c@{\hspace{0.56em}}c@{\hspace{0.56em}}c@{\hspace{0.56em}}c@{\hspace{0.56em}}c@{\hspace{0.56em}}c@{\hspace{0.56em}}c@{\hspace{0.56em}}c@{\hspace{0.56em}}c@{\hspace{0.56em}}c@{\hspace{0.56em}}c@{\hspace{0.56em}}c@{\hspace{0.56em}}c@{\hspace{0.56em}}c@{\hspace{0.56em}}c@{\hspace{0.56em}}c@{}}
\arrayrulecolor{green}\toprule
\textbf{Model} & \textbf{as} & \textbf{bn} & \textbf{brx} & \textbf{doi} & \textbf{gu} & \textbf{hi} & \textbf{kn} & \textbf{kok} & \textbf{ks} & \textbf{mai} & \textbf{ml} & \textbf{mni} & \textbf{mr} & \textbf{ne} & \textbf{or} & \textbf{pa} & \textbf{sa} & \textbf{sat} & \textbf{sd} & \textbf{ta} & \textbf{te} & \textbf{ur} \\
\midrule
USM & 74.8 & - & - & - & 22.4 & 20.5 & 46.5 & - & - & - & 78.2 & - & 30.7 & 38.8 & - & 35.7 & - & - & 99.8 & 58.9 & 40.3 & 29.1 \\
Azure & - & 33.6 & - & - & 54.7 & 27.2 & 57.3 & - & - & - & 62.3 & - & 44.6 & 57.0 & - & 39.9 & - & - & - & 61.2 & 53.3 & 31.3 \\
Whisper & 127.2 & 93.2 & - & - & 60.5 & 34.0 & 96.5 & - & - & - & 148.6 & - & 95.2 & 100.8 & - & 89.1 & 99.0 & - & 92.6 & 78.4 & 151.9 & 29.7 \\
MMS & 51.0 & 44.4 & - & 77.6 & 42.1 & 38.9 & 70.7 & - & - & 77.1 & 76.8 & - & 52.7 & - & 54.6 & 44.0 & - & - & - & 75.4 & 62.4 & 43.6 \\
\arrayrulecolor{black}\midrule
IndicASR & 20.4 & 15.9 & 25.9 & 33.3 & 21.4 & 15.0 & 30.3 & 31.4 & 39.3 & 33.6 & 40.5 & 26.1 & 18.2 & 16.4 & 23.4 & 12.9 & 22.8 & 35.4 & 29.6 & 31.2 & 26.8 & 14.4 \\
\arrayrulecolor{green}\bottomrule\arrayrulecolor{black}
\end{tabular}
\caption{Performance of our IndicASR model trained only on the \dataset{} dataset compared with different models on the \dataset{} benchmark.}
\label{tab:results}
\end{table}

\noindent\textbf{Evaluation:}  We train a multilingual 130M conformer based model (IndicASR) following the same architecture as proposed by \cite{model-rnnt}, using only the \dataset{} train set. We evaluate this model on the \dataset{} test set and compare it with existing state of the art models. As seen in Table \ref{tab:results}, IndicASR is the first model to support all the 22 languages and it outperforms all existing models on supported languages.

\noindent\textbf{Other tasks:} While in this work, we focus only on ASR, the collected data can be used for training and evaluating multiple speech and language tasks. More specifically, using the meta-data such as district id, speaker id, language id, list of entities in a given interaction, etc., the collected data can be used for the following tasks: speaker diarization, speaker identification, speaker verification, language identification, intent detection, entity extraction, query by example and audio denoising, crosslingual transfer in multilingual ASR systems, learning multimodal speech-text embeddings and so on. Further, given the data is being collected from multiple districts in phases, it can be used for evaluation of zero-shot transfer across districts with new accents. We also plan to create a framework for training and evaluating continual learning methods for ASR using this data as each district can be considered as a new episode of data. 
This comprehensive dataset, thus holds the potential to significantly advance a broad spectrum of speech and language processing tasks for multiple low resource languages from the Indian sub-continent.

\section{Conclusion}
We presented \dataset{}, a comprehensive speech dataset adhering to an inclusive diversity wishlist with fair representation of demographics, domains, languages and applications. The materials developed as a part of this work will serve as a good starter kit for data collection for other languages. These materials include (i) Wikipedia sentences from multiple domains (ii) prompts for digital interactions, (iii) questions from different domains and topics of interest, (iv) conversational role-play scenarios (v) elaborate transcription guidelines (vi) android application for on-field data collection and verification and (vii) a web based platform for transcription workflow management. These can be adapted with minimal effort to bootstrap large scale data collection efforts for other languages. 

\section{Limitations}
We list down the limitations of our work below.

\noindent\textbf{Number of districts:} Due to budget constraints, for some languages such as Hindi and Urdu which are spoken in multiple districts we will have to skip about 40\% of the districts (see Figure \ref{fig:language_atlas_of_India} for Hindi). 

\noindent\textbf{Number of speakers:} Currently, we target 100-150 speakers per district. Ideally, we would have liked to target 4-5 times more speakers per district but this would have significant cost implications. We made a choice to collect minimum 20 minutes per speaker to ensure that there is enough diversity in the content as opposed to just collecting 4-5 minutes per speaker and targeting 4x-5x more speakers. The latter had two disadvantages: (i) the first 2-3 minutes of data would just end up being related to icebreaker questions resulting in very little content diversity (ii) even if a speaker contributes just 4-5 minutes we still have to pay half day's wage which would increase the total budget by 4x-5x.

\noindent\textbf{Number of languages:} In this work, we focus on 22 scheduled languages. Going forward we would want to extend this effort to many more languages. 

\noindent\textbf{Secondary districts:} In this work, for a given language, we collect data only from districts where that language is identified as the primary language (i.e., spoken by majority of the population). Ideally, in subsequent phases we would want to collect data from districts even if that language is spoken only by a small population in that district.

\noindent\textbf{Topic Diversity:} Despite our best efforts some topics such as \textit{horse riding, sculpting}, etc. were underrepresented. This is understandable as very few participants could relate to such topics. Going forward, we will explicitly source participants who can speak on such underrepresented topics.

\noindent\textbf{Two-party Conversations:} Currently two-party conversations account for only 17\% of our data. We would want this number to be at least 25\%. The lower fraction is mainly due to the logistical difficulty in extracting meaningful conversations from two unrelated participants.

\noindent\textbf{Fine-grained evaluation:} Given the limited compute we have, we were not able to report a fine-grained evaluation across demographics and applications and just report average WERs over the entire test set. We will fix this by conducting a shared task where we invite the community to participate and report fine-grained results for multiple models.

\section{Ethics}
The data collection process underwent rigorous ethical review and approval by the Institute Ethics Committee. Their recommendations, including the provision of all instructions in participants' native language, were duly implemented to ensure clarity and comprehension. Prior to recording, participants were fully informed about the purpose of data collection, the expected effort from their side, and the potential uses of their data. Their consent was obtained explicitly before proceeding. Participants were compensated appropriately, in line with the prevailing daily wages in their respective districts. This compensation was intended to recognize their time, effort, and contribution to the research endeavor. Careful attention was paid to participants' comfort and well-being throughout the data collection process. Bets efforts were made to provide refreshments such as tea, coffee, water, and biscuits to ensure a hospitable environment conducive to participation. Participants were explicitly informed of their right to skip any task they felt uncomfortable with. The privacy and confidentiality of participants' personal identifiable information (PII) were rigorously safeguarded. No PII data will be shared externally, and measures were implemented to anonymize and protect sensitive information. Participants were given the option to have their videos deleted immediately after verification or to opt for a live call instead of recording a video for identity verification purposes. This choice was provided to accommodate individual preferences and ensure participant comfort and privacy. Lastly, all transcribers and other personnel employed in the project were also compensated appropriately according to prevailing salaries and wages. All tools will be released with a MIT license\footnote{https://opensource.org/licenses/MIT} and the dataset will be released with CC-BY-4.0 license\footnote{https://creativecommons.org/licenses/by/4.0/}, allowing commercial usage.

\section{Author Contributions}
\dataset~ is the result of collective efforts from several team members. To list down the contributions of the authors, we document the areas and list the authors contributing significantly to each of these areas. In each area, the contributors are listed sorted by last name.\\\\
\textbf{Field Research and Pilot Studies:} Tahir Javed, Mitesh M. Khapra, Janki Atul Nawale \\\\
\textbf{Data Preparation:} Hafsah Faquih, Kunal Sharad Gandhi, Eldho Ittan George, Tahir Javed, Sakshi Joshi, Sunjay Murali, Janki Atul Nawale, Pratiti Palit, Tripura Panchagnula, Sneha Ravishankar, Saranya Sukumaran \\\\
\textbf{Development and Maintenance of Karya:}  Eldho Ettan George, Tahir Javed\\\\
\textbf{Development and Maintenance of Shoonya:} Ishvinder Virender Sethi, Aparna Ananthanarayanan, Tahir Javed\\\\
\textbf{Management of Translation teams:}  Tahir Javed, Janki Atul Nawale\\\\
\textbf{Quality Control:} Tahir Javed, Manickam K M, Janki Atul Nawale \\\\
\textbf{Transcription guideline creation:} Hafsah Faquih, Tahir Javed, Sunjay Murali, Janki Atul Nawale, Pratiti Palit, Ambujavalli R \\\\
\textbf{Management of Transcription teams:} Hafsah Faquih, Kunal Sharad Gandhi, Tahir Javed, Sunjay Murali, Janki Atul Nawale, Pratiti Palit, Tripura Panchagnula, Ambujavalli R, Sneha Ravishankar, Saranya Sukumaran\\\\
\textbf{Model Training:} Kaushal Santosh Bhogale, Tahir Javed \\\\
\textbf{Model Evaluation:} Kaushal Santosh Bhogale, Tahir Javed, Sakshi Joshi, Deovrat Mehendale \\\\
\textbf{Management of data collection:} Hafsah Faquih, Kunal Sharad Gandhi, Eldho Ittan George, Tahir Javed, Sakshi Joshi, Krishnan Srinivasa Raghavan Karunganni, Sunjay Murali, Janki Atul Nawale, Pratiti Palit, Tripura Panchagnula, Sneha Ravishankar, Saranya Sukumaran, C Venkata Vaijayanthi\\\\
\textbf{Project Conceptualization and Direction:} Mitesh M. Khapra \\

\section{Acknowledgments}
Embarking on this mission was only possible due to the support of numerous organizations, individuals and members of the Indian language technology ecosystem. We would like to take a few sentences to thank all of them.\\\\
\textbf{Sponsors/Donors:} First and foremost, we thank the Ministry of Electronics and Information Technology (MeitY), Government of India, for setting up the ambitious Digital India Bhashini Mission with the goal of advancing Indian language technology. The human infrastructure comprising of a large team of translators, transcribers, reviewers and language experts who worked on this project were supported by the generous grant given by Digital India Bhashini Mission to IIT Madras to serve as the Data Management Unit for the mission.\\
We are indebted to Shri Nandan Nilekani and Shrimati Rohini Nilekani for believing in us and supporting our work through generous grants from EkStep Foundation and Nilekani Philanthropies. These grants were used for (i) supporting many of the students, research associates, and developers who worked on this project, (ii) fulfilling many of our compute needs, and (iii) recruiting project managers to oversee the massive pan-India data collection activity undertaken as a part of this work.\\
We thank Microsoft for their grant to support the creation of benchmarks for Indian languages.
We thank the Centre for Development and Advancement of Computing, Pune (CDAC Pune) for access to its ParamSiddhi super-computer which was used for mining bitext pairs at scale.\\\\
\textbf{IIT Madras:} We thank Prof. V Kamakoti (Director, IIT Madras), Prof. Mahesh V Panchagnula (Dean, IIT Madras), Prof. Ravindra Gettu (Dean, IIT Madras) and Prof. Manu Santhanam (Dean, IIT Madras) for their constant encouragement and administrative support. In particular, we are thankful for the office space provided to AI4Bharat which houses some of our students, researchers, language experts and administrative team.\\\\
\textbf{Indian language technology community:} We extend our heartfelt gratitude to the expansive Indian language technology community, comprising academia, startups, and the deep tech industry, both within India and across the globe. It is with immense gratitude that we acknowledge the incredible foundation laid by the giants of this community, whose pioneering work has paved the way for our endeavors. We are truly grateful for the knowledge, insights, and advancements that we have built upon, as we stand on the shoulders of these remarkable contributors. In particular, we thank Prof. Rajeev Sangal (Professor Emeritus, IIIT Hyderabad), Prof. Pushpak Bhattacharyya (IIT Bombay), Prof. Dipti Mishra (IIIT Hyderabad), Prof. Hema Murthy (IIT Madras), Prof. Umesh S (IIT Madras), Prof. Rajat Moona (IIT Gandhinagar), Prof. Ganesh Ramakrishnan (IIT Bombay), Partha Talukdar (Google Research India), Dr. Swaran Lata (MeitY), Dr. Sobha L (AU-KBC) and Dr. Ritesh Kumar (Dr. B.R. Ambedkar University) for their critical insights and constructive feedback in improving the translation guidelines used for creating the datasets released as a part of this work (we apologize if we have missed anyone). \\\\
\textbf{Language Experts:} We express our deepest gratitude to our exceptional and highly dedicated team of language experts, including translators, transcribers, reviewers, quality control experts, coordinators,  linguists and data leads whose invaluable contributions have been instrumental in the creation of \dataset. Their unwavering commitment to adhering to guidelines and their remarkable ability to work seamlessly as a cohesive unit, despite being geographically dispersed, is truly commendable. The quality and accuracy of the manual datasets developed as part of this endeavor owes much to their unwavering efforts. We extend our heartfelt thanks to every member of our remarkable language team for their outstanding dedication and invaluable contributions. \\\\
\textbf{Administration Team:} We are profoundly thankful to the remarkable individuals, Krishnan Karunganni S, Ambujavalli R, C V Vaijayanthi, Ravishankar Venkateswaran, for their exceptional dedication, patience, and extraordinary leadership in managing such an expansive team of data leads, language leads, coordinators and transcribers. Their unwavering commitment to orchestrating and guiding this diverse group of language experts is truly commendable. Through their exceptional organizational skills and expertise, they ensured
seamless coordination and maintained the highest standards of quality throughout the data collection process. We also thank our support staff Shanthi S, Bhanumathy M, Bhavana R, Suganya Kumaresan, and Kalaivanan A, who helped with recruitment and procurement. \\\\
\textbf{Development Team:} We also thank our development team comprising of our in-house engineers, as well as, engineers from Tarento for building Shoonya which enabled all the manual translation work. In the absence of Shoonya, it would have been impossible to manage such a diverse team spread across the country working towards a common goal. We thank members of our development team for their patience in working with the language experts and building features that helped improve both the speed and quality of translation. \\\\
\textbf{Partners:} We would also like to thank our partners, viz. Kashmir University, Goa University, Aripana Foundation, Korou Foundation, Pragyam Foundation, Calcutta Foundation, Suchana, Desicrew, Shaip, Navana Tech, Samskrita Bharati, Nava Data, Rekhta Foundation and Dogri Sanstha who helped in the data collection and transcription process. \\\\
Last, but not the least, we thank the Almighty for giving us the courage to embark on this mission!

\subsection{The Team Behind the Scenes}
This work was possible because the efforts put in by all the remarkable individuals listed below.
\begin{table}[!h]
    \begin{tabular}{p{0.35\linewidth}|p{0.45\linewidth}|p{0.1\linewidth}}
    \multicolumn{3}{@{}l}{\textbf{Administrative Team}} \\
    \arrayrulecolor{green}\toprule
    \textit{Name} & \textit{Designation} & \textit{Affiliation} \\
    \arrayrulecolor{green}\midrule
     Krishnan Karunganni S & Chief of Operations and Delivery & AI4Bharat \\
    Ravishankar Venkateswaran & Delivery Head & AI4Bharat \\
    Bhanumathy M & Recruitment & AI4Bharat \\
    Suganya Kumaresan & Recruitment & AI4Bharat \\
    Shanthi S & Operations & AI4Bharat \\
    Mohanarangan A & Operations & AI4Bharat \\
    Kalaivannan A & Operations & AI4Bharat \\
    Saranya B & Operations & AI4Bharat \\
    \arrayrulecolor{green}\bottomrule
    \end{tabular}
\end{table}

\begin{table}[!h]
    \begin{tabular}{p{0.35\linewidth}|p{0.45\linewidth}|p{0.1\linewidth}}
    \multicolumn{3}{@{}l}{\textbf{Project Managers}} \\
    \arrayrulecolor{green}\toprule
    \textit{Name} & \textit{Involvement} & \textit{Affiliation} \\
    \arrayrulecolor{green}\midrule
    C Venkata Vaijayanthi &  Data Collection & AI4Bharat \\
    Ambujavalli R & Data Transcription & AI4Bharat \\
    Manickam K M & Quality Control & AI4Bharat \\
    \arrayrulecolor{green}\bottomrule
    \end{tabular}
\end{table}

\begin{table}[!h]
    \begin{tabular}{p{0.83\linewidth}|p{0.1\linewidth}}
    \multicolumn{2}{@{}l}{\textbf{Data Leads}} \\
    \arrayrulecolor{green}\toprule
    \textit{Name} & \textit{Affiliation} \\
    \arrayrulecolor{green}\midrule
    Hafsah Faquih  & AI4Bharat \\
    Pratiti Palit  & AI4Bharat \\
    Sneha Ravishankar & AI4Bharat \\
    Saranya Sukumaran  & AI4Bharat \\
    Tripura Panchagnula & AI4Bharat \\
    Sunjay Murali  & AI4Bharat \\
    Kunal Sharad Gandhi & AI4Bharat \\
    Faizan Qadri & AI4Bharat \\
    Bishnu Prasad Barman & AI4Bharat \\
    Janki Atul Nawale  & AI4Bharat \\
    \arrayrulecolor{green}\bottomrule
    \end{tabular}
\end{table}

\begin{table}[!h]
    \begin{tabular}{p{0.35\linewidth}|p{0.3\linewidth}|p{0.25\linewidth}}
    \multicolumn{2}{@{}l}{\textbf{Karya Team}} \\
    \arrayrulecolor{green}\toprule
    \textit{Name} &\textit{Designation} & \textit{Affiliation} \\
    \arrayrulecolor{green}\midrule
    Eldho Ittan George & Project Associate & AI4Bharat \\
    Sakshi Joshi & MS Student & AI4Bharat, IIT Madras \\
    Tahir Javed & PhD Student & AI4Bharat, IIT Madras \\
    \arrayrulecolor{green}\bottomrule
    \end{tabular}
\end{table}

\begin{table}[!h]
    \begin{tabular}{p{0.35\linewidth}|p{0.45\linewidth}|p{0.1\linewidth}}
    \multicolumn{2}{@{}l}{\textbf{Shoonya Team}} \\
    \arrayrulecolor{green}\toprule
    \textit{Name} &\textit{Designation} & \textit{Affiliation} \\
    \arrayrulecolor{green}\midrule
    Aparna Ananthanarayanan &Manager& AI4Bharat\\
    Ishvinder Virender Sethi & Manager &AI4Bharat \\
    Kunal Tiwari & Backend Developer&AI4Bharat \\
    Kartik Virendra Rajput & Full Stack Developer & AI4Bharat \\
    Ayush Panwar & Full Stack Developer & AI4Bharat \\
    \arrayrulecolor{green}\bottomrule
    \end{tabular}
\end{table}

\begin{table}[!h]
\begin{tabular}{p{0.3\linewidth}|p{0.35\linewidth}|p{0.25\linewidth}}
\multicolumn{2}{@{}l}{\textbf{Data Collection Partners}} \\
\arrayrulecolor{green}\toprule
\textit{Language} &\textit{Name} & \textit{Affiliation} \\
\arrayrulecolor{green}\midrule
Assamese &  Bikash Chandra   &  Pragyam Foundation  \\ 
\arrayrulecolor{black}\midrule
Bengali &  Saumya Verma   &  Calcutta Foundation \\
 &  Mahima Verma   &  Calcutta Foundation \\
\arrayrulecolor{black}\midrule
Bodo &  Bikash Chandra   &  Pragyam Foundation  \\ 
\arrayrulecolor{black}\midrule
Dogri &  Dr. Preeti Dubey   & Dogri Sanstha  \\
\arrayrulecolor{black}\midrule
Gujarati & Rajan KM   &  Desicrew  \\
\arrayrulecolor{black}\midrule
Hindi & Rajan KM   &  Desicrew  \\
\arrayrulecolor{black}\midrule
Kannada & Rajan KM  &  Desicrew  \\
\arrayrulecolor{black}\midrule
Kashmiri & Dr. Adil Amin Kak & Kashmir University \\
 &  Nazima Mehdi  & Kashmir University \\
 \arrayrulecolor{black}\midrule
Konkani & Pradnya Bhagat   &  Goa University  \\
\arrayrulecolor{black}\midrule
Maithili & Avinash Kumar   &  Aripana Foundation  \\
\arrayrulecolor{black}\midrule
 & Dheeraj Kumar   &  Aripana Foundation  \\
 \arrayrulecolor{black}\midrule
Malayalam & Rajan KM   &  Desicrew  \\
\arrayrulecolor{black}\midrule
Marathi & Rajan KM   &  Desicrew  \\
\arrayrulecolor{black}\midrule
Manipuri &  Yasin   &  Korou Foundation \\
\arrayrulecolor{black}\midrule
Nepali &  Bikash Chandra   &  Pragyam Foundation  \\
\arrayrulecolor{black}\midrule
Odia &  Saumya Verma  &  Calcutta Foundation \\
 &  Mahima Verma   &  Calcutta Foundation \\
 \arrayrulecolor{black}\midrule
Punjabi &  Dr. Sudarshan Iyengar  &  IIT Ropar  \\
 &  Khushi Bhamra   &  IIT Ropar  \\
 &  Nimisha Mahajan   &  IIT Ropar  \\
 \arrayrulecolor{black}\midrule
Sanskrit &  Shashanka Hatwar   &  Sanskrit Bharati  \\
\arrayrulecolor{black}\midrule
Santali &  Sunil Murmu    &  Suchana  \\
 &  Kamala Murmu    &  Suchana  \\
 \arrayrulecolor{black}\midrule
Sindhi & Nilima Motwani    &  Rekhta Foundation  \\
 & Nida Fazli    &  Rekhta Foundation  \\
 \arrayrulecolor{black}\midrule
Tamil & Rajan KM   &  Desicrew  \\
\arrayrulecolor{black}\midrule
Telugu & Rajan KM   &  Desicrew  \\
\arrayrulecolor{black}\midrule
Urdu &  Nida Fazli    &  Rekhta Foundation  \\
\arrayrulecolor{green}\bottomrule
\end{tabular}
\end{table}

\clearpage

\setlength{\LTleft}{0pt}
\begin{longtable}{p{0.2\linewidth}|p{0.35\linewidth}|p{0.35\linewidth}}
\arrayrulecolor{black}
\multicolumn{3}{@{}l}{\textbf{Translation Team}}\\
\arrayrulecolor{green}\toprule \textit{Language} &\textit{Name} & \textit{Designation} \\
\arrayrulecolor{green}\midrule
\endfirsthead
\arrayrulecolor{black}
\multicolumn{3}{@{}l}%
{{\textbf{Translation Team} (continued)}} \\
\arrayrulecolor{green}\midrule \textit{Language} &\textit{Name} & \textit{Designation} \\
\arrayrulecolor{green}\midrule
\endhead
\arrayrulecolor{black}
\arrayrulecolor{green}\midrule\multicolumn{3}{r@{}}{{Continued on next page}} \\\arrayrulecolor{green}\midrule
\endfoot
\arrayrulecolor{black}
\endlastfoot

\arrayrulecolor{black}
Assamese  &  Devanga Pallav Saikia &  Language Lead, Senior Translator \\
 & Bikash Chandra & Senior Project Manager \\
 & Bishnu Prasad Barman & Translator \\
 & Dimpi Sarma  &  Translator \\
 & Bonya Baruah & Translator \\
 & Bikash Chetia & Translator \\
 & Kangkana Deka & Translator \\
 & Lelina Barman & Translator \\
 \midrule
Bengali & Sounak Dutta & Language Lead, Senior Translator \\
 & Shambhobi Ghosh & Senior Translator \\
 & Srija Mukherjee & Translator \\
 & Shreerupa Chattopadhyay & Translator \\
 & Natasha Ahmed & Translator \\
 & Kathakali Bhoumik Das & Translator \\
 & Atrayee Dutta & Translator \\
 \midrule
 Bodo & Prafulla Basumatry & Language Lead, Senior Translator \\
 & Bihung Brahma & Senior Translator \\
 & Bikash Chandra & Senior Project Manager \\
 & Sidwma Brahma & Translator \\
 & Sansuma Brahma & Translator \\
 & Jeetumoni Basumatry & Translator \\
 & Ria Borah Sonowal & Translator \\
 \midrule
Dogri & Preeti Dubey & Senior Project Manager \\
 & Lalit Mangotra & Senior Translator \\
 & Veena Gupta & Senior Translator \\
 & Shashi Pathania & Senior Translator \\
 & Anju Bala & Translator \\
 & Monika chandel & Translator \\
 & Kulbhushan Jasrotia & Translator \\
 \midrule
Gujarati & Pranav Pandya & Language Lead, Translator \\
 & Jayesh Adhyaru & Translator \\
 & Naresh Kapadia & Senior Translator \\
 & Faiz Masi & Translator \\
 & Jimal Patel & Translator \\
 \midrule
Hindi & Jaya Sarawati & Language Lead, Senior Translator \\
 & Sufiya Pathan & Senior Translator \\
 & Deepika Agarwal & Senior Translator \\
 & Aakansha Dubey & Translator \\
 & Neha Bhakal & Translator \\
 & Ayesha Pereira & Translator \\
 & Veda Bharti & Translator \\
 \midrule
Kannada & Anagha H. N. & Language Lead, Senior Translator \\
 & Adithi Raveendranath & Translator \\
 & Abhigna Joshi & Translator \\
 & Shivakumar R. M. & Translator \\
 & Arun Kumar & Translator \\
 & Goutham M & Translator \\
 & T.R. Nagesh & Translator \\
 \midrule
 Kashmiri & Vijay Wali & Senior Translator \\
 & Shafi Shauq & Senior Translator \\
 & Ambreen Farooq & Translator \\
 & Meer Bismah & Translator \\
 & Syed Samreen & Translator \\
 & Sumaya Jehangir & Translator \\
 & Nazima Mehdi & Senior Project Manager \\
 & Ishfaq Nisar & Translator \\
 \midrule
Konkani & Pradeep Padgaonkar & Senior Translator \\
 & Pradnya Bhagat & Senior Project Manager \\
 & Sandesh Prabhudesai & Senior Translator \\
 & Sharat Raikar & Senior Translator \\
 & Anwesha Singbal & Translator \\
 & Cia Fernandes & Translator \\
 & Ashwini Kamat & Translator \\
 \midrule
Maithili & Sanjay Jha & Language Lead, Translator \\
 & Avinash Kumar & Senior Project Manager \\
 & Yogendra Pathak & Senior Translator \\
 & Dr. Chandramani Jha & Senior Translator \\
 & Vikas Vineet Jha & Translator \\
 & Priyeshi Kumari & Translator \\
 & Rahul Kumar Jha & Translator \\
 & Vijay Deo Jha & Translator \\
 & Manoj Kumar Pathak & Translator \\
 & Tulika Swati & Translator \\
 & Prashant Kumar Jha & Translator \\
 & Nandan Kumar & Translator \\
 & Kishore Keshav & Translator \\
 & Sanjeev Kumar Jha & Translator \\
 & Deepak Kumar & Translator \\
 & Juli Jha & Translator \\
 & Swati Jha & Translator \\
 & Aditya Bhushan Mishra & Translator \\
 \midrule
Malayalam & Jebi Mariam Kurian & Language Lead, Translator \\
 & Manoj Varma & Senior Translator \\
 & C. V. Sudheendran & Senior Translator \\
 & Jihad M. & Translator \\
 & Jiza Mariam Kurian & Translator \\
 & Ann Mary Thomas & Translator \\
 & Srilekha Padmakuma Nambiar & Translator \\
 \midrule
Marathi & Kunal Gandhi & Language Lead,Translator \\
 & Paresh Prabhu & Senior Translator \\
 & Vrinda Sarkar & Senior Translator \\
 & Ranjana Pathak & Senior Translator \\
 & Saee Kodolikar & Senior Translator \\
 & Prasad Jog & Translator \\
 & Shweta Deshmukh & Translator \\
 & Bhushan Oke & Translator \\
 & Neha Satish Bandekar & Translator \\
 \midrule
 Manipuri & Reena Ashem & Language Lead, Senior Translator \\
 & Yasin Khan & Senior Project Manager \\
 & Chingtham Diana  Devi & Senior Translator \\
 & Diana Thingujam & Translator \\
 & Jahir Hussain & Translator \\
 & Sanju Pukhrambam & Translator \\
 & Alfina Khaidem & Translator \\
 & Kshetrimayum Momo & Translator \\
 & Padmabati Achom & Translator \\
 \midrule
Nepali & Sunita Dahal & Language Lead, Senior Translator \\
 & Bikash Chandra & Senior Project Manager \\
 & Dhaka Ram Kafle & Senior Translator \\
 & Lekhnath Chhetri & Senior Translator \\
 & Tika Ram Rai & Senior Translator \\
 & D. Ghimiray & Translator \\
 & Dr Srijana Sharma & Translator \\
 & Dr Khagen Sharma & Translator \\
 \midrule
Odia & Satyabrata Barik & Senior Translator \\
 & Pramodini Pradhan & Senior Translator \\
 & Sai Sudeep Das & Translator \\
 & Abhishek Parija & Translator \\
 & Suchishraba Sarangi & Language Lead \\
 & Bhimasena Bhol & Translator \\
 & Surendra Chandra Tripathy & Translator \\
 \midrule
Punjabi & Armin Virk & Language Lead, Translator \\
 & Pallavi Kaushal & Translator \\
 & Shallu Rani & Translator \\
 & Parneet Kaur & Translator \\
 \midrule
Sanskrit & Harisha H. M. & Language Lead, Senior Translator \\
 & Dr. Suresha & Senior Translator \\
 & Suprith S. & Translator \\
 & Sailaja Nittala & Translator \\
 & Vasudev Aital & Translator \\
 & Vivaswini & Translator \\
 & Dr. Narayan Dutt Mishra & Senior Translator \\
 \midrule
Santali & Kamala Murmu & Senior Project Manager \\
 & Baren Kisku & Senior Translator \\
 & Prasanta Kumar Hansda & Senior Translator \\
 & Baburam Murmu & Senior Translator \\
 & Sripati Tudu & Senior Translator \\
 & Urmila Murmu & Translator \\
 & Raju Mardi & Translator \\
 & Churki Hansda & Translator \\
 & Promila Hansda & Translator \\
 & Sova Tudu & Translator \\
 & Sanjiban Murmu & Translator \\
 & Satya Hembram & Translator \\
 & Guna Hembram & Translator \\
 & Sagen Murmu & Translator \\
 \midrule
Sindhi & Armin Virk & Language Lead, Translator \\
 & Dr. Nalini & Senior Translator \\
 & Prakash Tejwani & Translator \\
 & Bharati Chainani & Translator \\
 & Karan Vanni & Translator \\
 \midrule
 Tamil & Shakir Azeem & Language Lead, Senior Translator \\
 & Leema Rajavarman & Senior Translator \\
 & Shivapriya Murali & Translator \\
 & Sharmila Grahadurai & Translator \\
 & V Sayeelakshmi Rajaganapathy & Translator \\
 \midrule
Telugu & Shakir Azeem & Language Lead, Senior Translator \\
 & Karuna Vempati & Senior Translator \\
 & N. Sujatha & Senior Translator \\
 & Srimoukthika & Translator \\
 & Srilakshmi B. & Translator \\
 \midrule
Urdu & Dr. Irfan Ahmed & Senior Translator \\
 & Nazima Mehdi & Senior Project Manager \\
 & Aishwarya Diwakar & Translator \\
 & Anwar Wajhiuddin & Translator \\
 & Muhammad Anzar & Translator \\
 & Hasan Akram & Translator \\
 & Dr. Javaid Aziz Bhat & Translator \\
 & Hafsah Faquih & Translator \\
 & Habeebunnisa & Translator \\
 & Mohammad Afaan & Translator \\
 & Naziya Rasool & Translator \\
\arrayrulecolor{green}\bottomrule\arrayrulecolor{black}
\end{longtable}

\setlength{\LTleft}{0pt}
\begin{longtable}{p{0.25\linewidth}|p{0.35\linewidth}|p{0.3\linewidth}}
\multicolumn{3}{@{}l}{\textbf{Transcription Team}}\\
\arrayrulecolor{green}\toprule \textit{Language} &\textit{Name} & \textit{Designation} \\
\arrayrulecolor{green}\midrule
\endfirsthead
\arrayrulecolor{black}
\multicolumn{3}{@{}l}%
{{\textbf{Transcription Team} (continued)}} \\
\arrayrulecolor{green}\midrule \textit{Language} &\textit{Name} & \textit{Designation} \\
\arrayrulecolor{green}\midrule
\endhead
\arrayrulecolor{black}
\arrayrulecolor{green}\midrule\multicolumn{3}{r@{}}{{Continued on next page}} \\\arrayrulecolor{green}\midrule
\endfoot
\arrayrulecolor{black}
\endlastfoot
\arrayrulecolor{black}
Assamese  &   Daisy Devi   &   Superchecker \\
& Rima Saikia & Superchecker \\
& Angelina B. Dihingia & Superchecker \\
& Dewashree Ojah & Superchecker \\
& Harshita Talukdar & Superchecker \\
& Munmun Saikia & Superchecker \\
& Neelakshi Das &  Superchecker \\
& Parag Kumar Deka & Superchecker \\
& Shilpi Gogoi & Superchecker \\
& Mukti Duwara & Superchecker \\
& Suni Jyoti Kalita	& Superchecker \\
& Kangkana Deka	& Superchecker \\
\midrule
Bengali & Maamritha nandi &	Superchecker \\
& Ayan Chatterjee & Superchecker \\
& Sayantani Banerjee & Superchecker \\
& Subhadeep Bhattacharjee	& Superchecker \\
\midrule
Bodo & Pronita Basumatary & Superchecker \\
& Jwngsar Brahma & Superchecker \\
& Bithika Khaklary	& Superchecker \\
& Kenak Basumatary	& Superchecker \\
& Pabita Basumatary	& Superchecker \\
& Donald King Narzary	& Superchecker \\
& Mainu Basumatary	& Superchecker \\
& Mili Brahma	& Superchecker \\
\midrule
Dogri & Bansi Lal Sharma	& Superchecker \\
& Rakesh Kumar	& Superchecker \\
& Sunil Choudhary	& Superchecker \\
& Atul Sharma	& Transcriber \\
& Kamlesh Kumari & Transcriber \\
& Kumari Rita	& Superchecker \\
& Rajesh Manhas	& Superchecker \\
& Shagun Singh	& Transcriber \\
& Skindya Devi	& Transcriber \\
& Sunil Kumar	& Transcriber \\
& Varinder Kumar	& Transcriber \\
\midrule
Gujarati & Charmi Soni & Superchecker \\
\midrule
Hindi & Lalita Manoj Gehlot & Superchecker \\
& Sudeep Kumar Mishra & Superchecker \\ 
& Humera Begum	& Superchecker \\ 
& Rakshit Ghai	& Superchecker \\
\midrule
Kannada & Anusha & Superchecker \\
& Sandhya Gopal Shetty	& Superchecker \\
& Bharathi R Nayak & Superchecker \\
\midrule
Kashmiri & Rinko Ji Koul &	Superchecker \\
& Mohammad Asjad Khan & Superchecker \\
& Zargar Adil Ahmad	& Superchecker \\
& Naira Farooq &	Transcriber \\
& Uzma Nisar &	Transcriber \\
& Sheeba Shafi	& Transcriber \\
& Rehana Qasim Shah &	Transcriber \\
& Rafia Nabi & Transcriber \\
\midrule
Konkani & Anwesha Sigbal & Superchecker \\
& Sujatha kambli & Superchecker \\
& Kapila Surendra Desai &	Superchecker \\
& Sumeda Ankit Sheldekar & Transcriber \\
& Ankita Anand Zamburlikar & Transcriber \\
& Cialini Fernandes	& Transcriber \\ 
\midrule
Maithili & Sanjeev Kumar Jha &	Superchecker \\
& Juli Jha & Superchecker \\
& Sabita mishra &	Superchecker \\
& Nimmi Kumari	& Superchecker \\
& Pankaj Kumar Jha &	Superchecker \\
& Baiju Kumar Jha &	Transcriber \\
& Kanchan mishra & Transcriber \\
& Nandita Mishra &	Transcriber \\
& Prabhat Kumar Jha &	Transcriber \\
& Prabhu Nath Mishra &	Transcriber \\
& Raghunath Mukhiya	& Transcriber \\
\toprule
Malayalam & Niveditha Varma	& Superchecker \\
& Amritha Krishnan	& Superchecker \\
& Ann Mary Thomas & Superchecker \\
& Rajitha K V & Superchecker \\
& Lipi Pushpakaran & Superchecker \\
\midrule
Manipuri & Tongbram Jimi Singh & Transcription Reviewer \\
& Mousami Oinam	& Transcriber \\
& Anil Chingakham	& Transcriber \\
\midrule
Marathi & Suma G & Superchecker \\
& Rashmi Arun Sathe &	Superchecker \\
& Shraddha P. Prabhu & Superchecker \\
& Shweta Deshmukh & Superchecker \\
\midrule
Nepali & Suraj Sharma	& Superchecker \\
& Padam Parajuli	& Superchecker \\
& Ram Jiwan Rai	& Superchecker \\
& Bhakti Rai	& Superchecker \\
& Mausam Sharma	& Superchecker \\
& Bikash Chetri	& Transcriber \\
& Furtengi Sherpa	& Transcriber \\
& Keshav Sapkota	& Transcriber \\
& Menuka Bhujel	& Transcriber \\
& Sharmila Sharma	& Transcriber \\
\midrule
Odia & Rupanjali Badajena & Superchecker \\
& Abhipsa Mohanti	& Superchecker \\
& Surendra Chandra Tripathi &	Superchecker \\
& Pragyansmita Sharma	& Superchecker \\
& Nibedita Sahoo	& Superchecker \\
\midrule
Punjabi & Sandeep Kaur &	Superchecker \\
& Gurjeet Kaur	& Superchecker \\
& Jaspal Singh & Superchecker \\
& Sandeep Singh	& Superchecker \\
\midrule
Sanskrit & Shriganesh Devaru & Superchecker \\
& Nagaranjan V	& Superchecker \\
& N Sridhar & Superchecker \\
\midrule
Santali & Suku Murmu & Superchecker \\
& Sunil Murmu & Superchecker \\
& Prashantha	& Superchecker \\
& Sagen	& Superchecker \\
& Ladam Murmu	& Superchecker \\
& Biplab Tudu	& Transcriber \\
& Ganesh Saren	& Transcriber \\
& Sadhuram Hembram	& Transcriber \\
& Sandhya Mandi &	Transcriber \\
\midrule
Sindhi & Heeral Kavlani &	Superchecker \\
& Jyoti Bhatia	& Transcription Reviewer \\
\midrule
Tamil & Sumithra & Superchecker \\
& Savitha V	& Superchecker \\
& Krishna Arvind & Superchecker \\
& Madhu Vanisree &	Superchecker \\
& Hemalatha Venkatesh &	Superchecker \\
& Usha M K V & Superchecker \\
& Petchiammal	& Superchecker \\
\midrule
Telugu & Ranjith	& Superchecker \\
& Prudhvi Sagar	& Superchecker \\
& N Janaki	& Superchecker \\
& Herugu Deepak Iyengar Amrutha Lakshmi &	Superchecker \\
& Radhe Shyam Salopanthula	& Superchecker \\
& Suguru Sri Harsha Rao &	Superchecker \\
\midrule
Urdu & Syed Raqib Siraj	& Superchecker \\
& Shahnawaz Alam	& Superchecker \\
& Saad Ahmad Mirani	& Superchecker \\
& Faizur Rehman	& Superchecker \\
\arrayrulecolor{green}\bottomrule\arrayrulecolor{black}
\end{longtable}

\setlength{\LTleft}{0pt}
\begin{longtable}{p{0.2\linewidth}|p{0.5\linewidth}|p{0.2\linewidth}}
\multicolumn{3}{@{}l}{\textbf{Quality Control Team}}\\
\arrayrulecolor{green}\toprule \textit{Language} &\textit{Name} & \textit{Designation} \\
\arrayrulecolor{green}\midrule
\endfirsthead
\arrayrulecolor{black}
\multicolumn{3}{@{}l}%
{{\textbf{Quality Control Team} (continued)}} \\
\arrayrulecolor{green}\midrule \textit{Language} &\textit{Name} & \textit{Designation} \\
\arrayrulecolor{green}\midrule
\endhead
\arrayrulecolor{black}
\arrayrulecolor{green}\midrule\multicolumn{3}{r@{}}{{Continued on next page}} \\\arrayrulecolor{green}\midrule
\endfoot
\arrayrulecolor{black}
\endlastfoot
\arrayrulecolor{black}
Assamese & Tonmoyee Bhuyan & QC Analyst \\
& Anima Chetry & QC Analyst \\
\midrule
Bengali & Sunanda Sarkar	& QC Analyst \\
& Anamika Das &	QC Analyst \\
& Aditi Ghosh	& QC Analyst \\
\midrule
Bodo & Ria Borah Sonwal &  QC Analyst \\
& Bashiram Basumatary &  QC Analyst \\
\midrule
Dogri & Rahul Singh	& QC Analyst \\
& Kuldeep Kumar &	QC Analyst \\
\midrule
Gujarati & Miraba Yogendrasinh Chudasama & QC Analyst \\
& Yutika Amitkumar Chauhan &	QC Analyst \\
& Richa Thaker Bhatt &	QC Analyst \\
\midrule
Hindi & Afifa Anjum	& QC Analyst \\
& Rimsha Jairajpuri &	QC Analyst \\
\midrule
Kannada & Asha N &	QC Analyst \\
& Jayaseetha M N & QC Analyst \\
& Roopkamal S	& QC Analyst \\
\midrule
Kashmiri & Tahir Ahmad Sheikh	& QC Analyst \\
& Fatima Ashraf	& QC Analyst \\
\midrule
Konkani   & Varsha Vishnu Garad         & QC Analyst \\
& Pooja C Tople                   & QC Analyst \\
\midrule
Maithili  & Chitralekha Anshu               & QC Analyst \\
\midrule
Malayalam & Haritha M S                     & QC Analyst \\
& Anjali P Nair                   & QC Analyst \\
& Hanan A Vaheed                  & QC Analyst  \\
\midrule
Manipuri  & Khumanthem Yuremba Meitei       & QC Analyst \\
\midrule
Nepali & Anima Chetry & QC Analyst \\
\midrule
Marathi   & Prachi Dnyaneshwar Dharashivkar & QC Analyst \\
& Akshay Sarjerao Talekar         & QC Analyst   \\
& Manpritkaur Gurmitsingh Ragi    & QC Analyst  \\
& Nikita Parolekar                & QC Analyst  \\
& Vishwanath Arjun Bhandwale      & QC Analyst  \\
\midrule
Odia      & Mohammed Irfan             & QC Analyst \\
& Aliva Pradhan                   & QC Analyst  \\
& Pranita Das                     & QC Analyst  \\
& Ulka Antariksha                 & QC Analyst   \\
\midrule
Punjabi   & Amandeep Singh                  & QC Analyst \\
& Kanchan Bala                    & QC Analyst \\
\midrule
Sanskrit  & Sujatha Bala                  & QC Analyst \\
\midrule
Santali & Aditi Ghosh &  QC Analyst \\
\midrule
Tamil     & Ramya                           & QC Analyst \\
& Muthathal Subramanian           & QC Analyst \\
& Elumalai Ellappan               & QC Analyst   \\
& Suganthi V                      & QC Analyst   \\
& Hemavardhini R                  & QC Analyst   \\
& Yuvaraj R                       & QC Analyst   \\
& Gnanasoundari A                 & QC Analyst  \\
\midrule
Telugu & Peddareddygari Praneeth Reddy   & QC Analyst \\
& Vakkapati Monika Chandana       & QC Analyst  \\
& Purnag                          & QC Analyst  \\
& Srilakshmi Garimella            & QC Analyst  \\
\midrule
Urdu      & Sabiya Jan            & QC Analyst  \\
& Imtiyaz Ahmad Dar               & QC Analyst  \\
& Tariq Ahmad Sheikh              & QC Analyst \\

\arrayrulecolor{green}\bottomrule\arrayrulecolor{black}
\end{longtable}

\setlength{\LTleft}{0pt}
\begin{longtable}{p{0.2\linewidth}|p{0.5\linewidth}|p{0.2\linewidth}}
\multicolumn{3}{@{}l}{\textbf{Field Coordinators}}\\
\arrayrulecolor{green}\toprule \textit{Language} &\textit{Name} & \textit{Designation} \\
\arrayrulecolor{green}\midrule
\endfirsthead
\arrayrulecolor{black}
\multicolumn{3}{@{}l}%
{{\textbf{Field Coordinators} (continued)}} \\
\arrayrulecolor{green}\midrule \textit{Language} &\textit{Name} & \textit{Designation} \\
\arrayrulecolor{green}\midrule
\endhead
\arrayrulecolor{black}
\arrayrulecolor{green}\midrule\multicolumn{3}{r@{}}{{Continued on next page}} \\\arrayrulecolor{green}\midrule
\endfoot
\arrayrulecolor{black}
\endlastfoot
\arrayrulecolor{black}

Assamese & Manas Pratim Kalita & Coordinator \\
 & Srijana Ojha & Coordinator \\
 & Juhi Bayan & Coordinator \\
 & Madhushree Saud & Coordinator \\
 & Abinash Bordoloi & Coordinator \\
 & Pangkhi Das & Coordinator \\
 & Mayuri Deka & Coordinator \\
 & Tanmoy Jyoti Bhuyan & Coordinator \\
 & Tonmoyee Bhuyan. & Coordinator \\
 & Tarun Chetia & Coordinator \\
 & Munmi Borpatra Gohain & Coordinator \\
 & Niloy Pratim Kashyap & Coordinator \\
 & Gitika Barman & Coordinator \\
 & Rikshikha Kashyap & Coordinator \\
 & Sadhana Devi & Coordinator \\
 & Biswajyoti Deka & Coordinator \\
 & Mahesh Sharma & Coordinator \\
 & Tarun Chetia & Coordinator \\
 & Bhabani Sarmah & Coordinator \\
 & Jadab Kalita & Coordinator \\
 & Jayashree Saikia & Coordinator \\
 & Bhaskar Gogoi & Coordinator \\
 & Pujashree Saikia & Coordinator \\
 & Luckey Lahon & Coordinator \\
 & Rupjyoti Narah & Coordinator \\
 & Pankaj Medhi & Coordinator \\
 & Nilutpal Saikia & Coordinator \\
 & Pranab Borah & Coordinator \\
 & Rupam Hazarika & Coordinator \\
 \midrule
Bodo & Narbilash Brahma & Coordinator \\
 & Didwm Basumatary & Coordinator \\
 & Nisha Daimary & Coordinator \\
 & Punpun Basumatary & Coordinator \\
 & Dhanjita Swargiary & Coordinator \\
 & Nisha Daimary & Coordinator \\
 & Punpun Basumatary & Coordinator \\
 & Dhanjita Swargiary & Coordinator \\
 & Kriti Deepa Brahma & Coordinator \\
 & Didwm Basumatary & Coordinator \\
 & Eragdao Brahma & Coordinator \\
 & Albina Islary & Coordinator \\
 & Hungama Narzary & Coordinator \\
 & Jasmine Hajoary & Coordinator \\
 & Chinki Narzary & Coordinator \\
 & Aloka Muchahary & Coordinator \\
 \midrule
Dogri & Navedika Mishra & Coordinator \\
 & Rakesh Kumar & Coordinator \\
 \midrule
Hindi & Sanjay Kumar Paswan & Coordinator \\
 & Ashok Kumar Jha & Coordinator \\
 & Ajay Shankar Jha & Coordinator \\
 & Anjali Jha & Coordinator \\
 & Menka Minu & Coordinator \\
 & Bhuneshwar Tiwari & Coordinator \\
 & Shrishti Pathak & Coordinator \\
 & Shailesh Kumar Shukla & Coordinator \\
 & Shivani Singh & Coordinator \\
 & Anjani & Coordinator \\
 & Akanksha Yadav & Coordinator \\
 & Pooja Mishra & Coordinator \\
 & Ram Kumar Dubey & Coordinator \\
 & Nishant Dubey & Coordinator \\
 & Anjali Mishra & Coordinator \\
 & Tanu Yadav & Coordinator \\
 & Sushil Kumar Ray & Coordinator \\
 & Sanjay Kumar Paswan & Coordinator \\
 & Vijay Shankar Jha & Coordinator \\
 & Pooja Mishra & Coordinator \\
 & Atul Sharma & Coordinator \\
 & Shatrudhan Gupta & Coordinator \\
 & Vikas Gir & Coordinator \\
 & Ankit Singh & Coordinator \\
 & Vineeta Shukla & Coordinator \\
 & Rajiv Thakur & Coordinator \\
 & Gautam Govind & Coordinator \\
 & Nitish Kumar & Coordinator \\
 \midrule
Kashmiri & Faizan Qadri & Coordinator \\
 & Aneesa Khan & Coordinator \\
 & Rayees Ahmad Lone & Coordinator \\
 & Zubair Ahmad Bhat & Coordinator \\
 & Faisal Ahmad Malik & Coordinator \\
 & Rizwan Uz Zaman Wan & Coordinator \\
 & Idrees Rehman Mir & Coordinator \\
 & Yasir Mustaq Bhat & Coordinator \\
 & Mehjabeena Nisar & Coordinator \\
 & Sakeena Mohi Ud Din & Coordinator \\
 \midrule
Konkani & Manthan H Dessai & Coordinator \\
 & Snehal Devanand Prabhu & Coordinator \\
 & Santoshi Mahendra Bakal & Coordinator \\
 & Venkatesh Prabhu & Coordinator \\
 \midrule
Maithili & Rajiv Thakur & Coordinator \\
 & Gautam Govind & Coordinator \\
 & Nitish Kumar & Coordinator \\
 & Sandeep Kumar Jha & Coordinator \\
 & Saurabh Jha & Coordinator \\
 & Geethanjali Mishra & Coordinator \\
 & Vijay Shankar Jha & Coordinator \\
 & Hirendra Kumar Jha & Coordinator \\
 & Saket Kumar & Coordinator \\
 & Shankara Nand Jha & Coordinator \\
 & Gajendra Narayan Jhan & Coordinator \\
 & Mukesh Kumar Mishra & Coordinator \\
 & Neetu Kumari & Coordinator \\
 & Devashish Jha & Coordinator \\
 & Sushil Kumar Ray & Coordinator \\
 & Ashish Kumar Jha & Coordinator \\
 & Rakesh Kumar Ray & Coordinator \\
 & Dhiraj Kumar Jha & Coordinator \\
 & Premalata Kumari & Coordinator \\
 & Raj Kumar Jha & Coordinator \\
 & Raghunath Mukhiya & Coordinator \\
 & Alia Ali & Coordinator \\
 & Krishna Mohan Kumar & Coordinator \\
 & Shital Kumari & Coordinator \\
 & Sanjay Kumar & Coordinator \\
 & S & Coordinator \\
 & Narendra Kumar Mandal & Coordinator \\
 & Ashwni Kumar & Coordinator \\
 & Rajan K Pandey & Coordinator \\
 & Abha Kumari & Coordinator \\
 & Kiran Kumari & Coordinator \\
 & Mohammad Sirajuddin & Coordinator \\
 & Naveen Kumar & Coordinator \\
 & Ranjan Kumar & Coordinator \\
 \midrule
Nepali & Susmita Gurung & Coordinator \\
 & Dinesh Kharga & Coordinator \\
 & Pukar Rai & Coordinator \\
 & Tulshi Rai & Coordinator \\
 & Chanda Khawas & Coordinator \\
 & Pranoy Rai & Coordinator \\
 & Ritu Rai & Coordinator \\
 & Ajit Biswa & Coordinator \\
 & Nishal Sharma & Coordinator \\
 & Manisha Subba & Coordinator \\
 & Sarita Lama Rai & Coordinator \\
 & Bishal Cheetri & Coordinator \\
 & Ranjeeta Lama Chhetri & Coordinator \\
 & Bivek Sarki & Coordinator \\
 \midrule
Sanskrit & Shashanka Hatwar T V & Coordinator \\
 & Divya Pandey & Coordinator \\
 & Haripriya R Kulkarni & Coordinator \\
 & Vidyadhare & Coordinator \\
 & Dr. Rukmangadha & Coordinator\\
\arrayrulecolor{green}\bottomrule\arrayrulecolor{black}
\end{longtable}
We also acknowledge the efforts of transcribers, translators, administrative teams, field coordinators that are working with our external partners

\clearpage

\bibliographystyle{ieeetr}
\bibliography{custom}

\clearpage


\appendix

\section{The Journey} 
\label{apx:the-journey}
\dataset~has been a life changing experience for all of us involved in the project. Below we share some of our experiences with the reader.

\noindent\textbf{Humble Beginnings in the Holy City of Madurai.}
Our journey commenced in Madurai, Tamil Nadu, famous for the Meenakshi Amman Temple. Seeking blessings from the \textit{Devi}, we commenced our journey with high hopes and a clear vision. However, this pilot quickly became a reality check, challenging our assumptions at every turn, especially regarding participant mobilization. Despite the dense population of India, finding willing participants became unexpectedly difficult. Trust was a major hurdle; many were hesitant to share personal information during registration, fearing potential fraud, especially when digital transactions were mentioned. This skepticism slowed down the mobilization process significantly, making it challenging to achieve the desired diversity in age and gender ratios.

Additionally, the time commitment required from participants—sometimes extending up to four hours to complete the recording process—added another layer of complexity. This duration, much longer than anticipated, tested the patience and commitment of our participants. Hesitancy in speaking freely was another obstacle; many participants showed reluctance in opening up, leading to numerous retakes to capture responses that were natural and usable. This reluctance often resulted in responses that lacked depth and spontaneity, necessitating multiple attempts to elicit more meaningful dialogue. The culmination of these challenges not only extended the duration of our pilot but also highlighted the importance of building trust and ensuring clarity in communication to facilitate smoother data collection processes in the diverse linguistic landscape of India.

\noindent\textbf{The Forgotten Generation.} Right from the beginning, we were very clear that we wanted sufficient participant from the senior citizen age group to capture their rich life experiences and tap into their repository of knowledge about Indian customs, traditions and beliefs. However, addressing the underrepresentation of the senior demographic emerged as a significant hurdle. The limited presence of individuals aged 60 and above necessitated a reevaluation of our outreach efforts.  Engaging with old age homes, senior citizen clubs, and conducting home visits became essential strategies to include this vital segment of the population. This challenge highlighted the importance of inclusivity and the need for tailored approaches to ensure diverse demographic participation. Our concerted efforts to involve senior citizens bore fruit in an unexpectedly delightful way, particularly notable in the diverse regions of India. A striking example of this success was observed in Jammu, where the older population displayed remarkable enthusiasm towards contributing to the project. This enthusiasm stemmed not just from their fluency in their native language (Dogri, in this case) but also from a deep-seated urge to preserve and pass on their linguistic heritage. Contrary to the challenges faced with other age groups and languages, senior citizens in Jammu and beyond became invaluable participants. Their eagerness to contribute not only enriched our dataset with authentic, nuanced language use but also underscored the critical role of senior citizens in safeguarding linguistic diversity. 

\noindent\textbf{The Need for a Guiding Hand.} In the initial pilots, our fears about ‘What if people don't speak’ were confirmed. We learnt the hard way that eliciting fluent speech with good quality content from individuals in interactions with strangers posed an intriguing challenge. Recognizing the pivotal role of participant’s comfort in encouraging natural dialogue, we acknowledged the importance of assigning dedicated coordinators to accompany participants throughout the data collection process. Through targeted training, these coordinators were equipped with effective communication skills to engage participants authentically. Consequently, we established a procedural framework to ensure coordinators' proficiency in guiding participants through the data collection procedure. The coordinators also helped the participants in overcoming technical challenges in installing the app, getting accustomed to the ``record-verify-submit'' workflow on the app and clarifying the expectation with respect to each microtask.

\noindent\textbf{Back to the drawing board.} After these initial pilots, we took a pause to critically reassess our data collection strategy. Insights from our initial pilots revealed that participants often provided repetitive answers, and everyday conversations failed to yield a diverse vocabulary and had very little coverage of names, numbers, entities, and brand names critical for downstream Automatic Speech Recognition (ASR) applications. The pilots also offered us a glimpse into people's interactions with technology and their expectations from it, such as placing orders or completing digital transactions. We also realised that simple prompts like ``talk about politics'' fell short, and sparking a meaningful dialogue between strangers over a phone call proved challenging, often resulting in mere exchanges of pleasantries without covering a wide array of topics.

Recognizing these gaps, we returned to the drawing board for an in-depth pre-collection phase. Our goal was to collect sentences with rich vocabulary, craft engaging questions spanning various domains, and create scenarios that simulate everyday digital interactions. We also refined our process to include tasks that would elicit responses rich in sequences of numbers, named entities, locations, dates, etc., and add role-play scenarios with detailed narratives to encourage dynamic conversations between two parties. 

\noindent\textbf{Weather Plays Spoilsport.}
After refining our approach, we headed to the extreme north of India, to the beautiful land of Kashmir. Excited as we were, we encountered the formidable challenge of the region's harsh winter. Starting our pilot in mid-November in Srinagar, we were keenly aware of the narrow operational window before the onset of heavy snowfall, which renders data collection nearly impossible from December to February. The unique weather conditions and the limited daylight hours significantly restricted our daily operations, allowing us only a brief period between 10 a.m. and 5:30 p.m. for recordings. This time constraint meant that each coordinator could only manage sessions with a maximum of two participants per day, highlighting the need for a more adaptable approach to meet our productivity and deadline goals. As we moved forward, it became crucial to innovate our data collection methods, shifting towards conducting recordings within the warmth and accessibility of participants' homes in the subsequent districts. 
This was a good learning for us and an early realisation that given the diverse geographical landscape of India, we will have to be mindful of weather conditions: be it self-imposed afternoon curfews during summer in West Bengal, mobility issues during monsoon in Kerala and Goa, floods in Northeast India, harsh winters in the northern parts of Punjab, Delhi, Kashmir, Jammu and so on. On a lighter note, while working on \dataset, we have become experts on weather conditions in different parts of the country. 

\noindent\textbf{A Journey to the Remotest Parts of India.} After covering Kashmir (Kashmiri) and Jammu (Dogri), our journey took us to the north-eastern parts of India to conduct pilot studies covering Assamese, Bodo, Manipuri, and Nepali. Initially filled with enthusiasm, our venture into Assam's relatively tranquil settings soon led us into the more secluded and challenging terrains of Bodoland. Here, the sparse population and limited access to resources questioned the feasibility of our project, presenting a stark contrast to our prior experiences. Bodoland, with its serene yet isolated landscape of traditional tribal huts and quiet, dimly lit pathways, intimidating silence with no vehicular noise offered a unique set of challenges. The initial low turnout at our designated collection site in Kokrajhar prompted us to rethink our approach to engaging with remote tribal communities. Their concerns over privacy and the openness to share opinions reminded us of the delicate balance required to ensure inclusivity while being mindful of socio-political or geographical hurdles.

Our journey further led us to Kalimpong in West Bengal, a region where the linguistic landscape shifts dramatically to predominantly Nepali speakers, diverging significantly from the Bengali culture of the state. This diversity within a single state highlighted the intricate patchwork of India's linguistic heritage. In response, we tailored our data collection approach, creating district-specific hints to engage participants in a manner that resonated with their unique linguistic identity.

Following this, the endeavor in Imphal West, Manipur, introduced us to a different set of challenges marked by remoteness and the complexities of operating amidst constant disruptions. The initiation of the pilot in the compact confines of a local hotel was just the beginning of a journey punctuated by curfews, riots, and internet shutdowns, which significantly delayed our progress. The decision to romanize text data to include older participants unfamiliar with the local script, and continuing annotation work offline during internet shutdowns, were necessary to adapt to the ever changing conditions in the state.

This expedition across the remote parts of India was not just a journey through diverse geographical landscapes but a deep dive into the heart of India's linguistic diversity. Each region, with its unique challenges, taught us the importance of resilience, adaptability, and the profound value of including voices from every region of the country, no matter how remote.

\noindent\textbf{The Rural Urban Divide.} Following this, we did some pilots in rural districts in West Bengal covering two languages, Bengali and Santali. This illuminated the stark urban-rural divide, presenting unique challenges and learning opportunities at every turn. As we ventured into the tribal regions to engage with Santali-speaking communities, the serene yet complex rural setting offered a vivid contrast to the bustling urban environments we had previously navigated. Conducting recording sessions in the midst of forests, under the canopy of trees, or in the humble backyards of huts, we were confronted with the realities of rural life: limited internet connectivity, the scarcity of participants fitting specific demographic profiles, unpredictable weather, and the reticence of women to participate.

These experiences shed light on the significant divide between urban and rural contexts, especially in terms of technological accessibility and the relevance of certain questions and use cases. It became apparent that some of our initial questions, designed with an urban mindset, were not resonant with the daily experiences of rural participants. For instance, the concept of ``hailing a cab'' was alien to many, revealing a disconnect in the applicability of our queries. This insight prompted us to revisit our approach and make our questions more closely aligned with the realities of rural life. We modified scenarios to involve ``arranging for transport for cattle or food grains,'' among other adjustments, ensuring our questions and use-cases were relevant and relatable to the lives of our rural participants.

This re-calibration was not merely about changing the wording of questions but about cultural and contextual sensitivity in linguistic data collection. This journey through the contrasting landscapes of India reinforced the notion that to ensure inclusivity we need to constantly change our assumptions, adapt and improve our processes.

\noindent\textbf{The Silence Amidst the Noise.} While rural areas offered a raw and unfiltered glimpse into India's linguistic diversity, urban settings introduced a different set of challenges. In bustling metropolises like Mumbai, the search for tranquil venues for recording sessions became a Herculean task, particularly in densely populated areas. The urban clamor and the scarcity of quiet spaces escalated the costs and complexities of data collection, underscoring the logistical hurdles unique to urban centers. Moreover, the enthusiasm for participating in data collection efforts was noticeably muted among the urban populace, especially among professionals leading busy lives. This apathy necessitated innovative mobilization strategies to engage a demographic that seemed distant from the cause. Despite these obstacles, the endeavor to capture the linguistic essence of India's urban centers was as crucial as that of its rural counterparts. Overcoming this silence and capturing voices amidst the noise became an essential part of our mission.

\noindent\textbf{India, a land of many festivals.} Navigating the vibrant maze of India's festivals proved to be one of the most colorful challenges in our data collection journey. In a country where each region celebrates its own set of festivals with fervor and devotion, scheduling work around these celebrations was akin to finding a needle in a haystack of holidays. From Durga Puja in West Bengal and Odisha, Ramzaan in Kashmir, Bihu and Pongal across other parts, to the universally celebrated Diwali, our calendar was a mosaic of cultural festivities.

The complexity of scheduling was humorously encapsulated in a conversation with one of our partners, which turned into a comedic back-and-forth of date dodging. 

\begin{verbatim}
Partner: We can't start in May as it is too hot that time of the year!

We: Ok, let's start in June then.

Partner: No, that would be difficult due to the monsoon season. 
Both June and July would be washed out, quite literally!

We: That looks bad. Then we should definitely start in August.

Partner: But then we would have Ganesh Chaturthi which is a very 
important festival here. No participants would turn up during that time.

We: Phew, what about September?

Partner: Schools and colleges will have exams so we will not be able to use
them as venues (other options would be expensive).

We (frustrated):  Okay, then I guess after that we would have to wait for 
Dusshera (October), Diwali (November), Christmas and New Year (December) to 
also pass by.

Partner (with a straight face): Yes, that would be ideal! 
\end{verbatim}
This humorous exchange underscored a significant reality of executing a prioject of this scale in India. It taught us the importance of flexibility, patience, and the ability to laugh at the seemingly impossible task of scheduling around the endless cycle of festivals. In the end, these challenges just became a part of our journey, making every successfully covered district feel like a festival in its own right.

\noindent \textbf{Everything that could go Wrong.} Amidst all the celebrations, we soon realised that in a remote and distributed setup involving a large number of people, ensuring quality is a challenging task. Early in the pilot phases, we observed participants abruptly stopping mid-speech or drifting into unrelated conversations with coordinators without stopping the recording, leading to fragmented audio and content corruption. Despite comprehensive guidelines and thorough training, the human element introduced unpredictability in task execution, with participants sometimes simply reading the questions/prompts instead of answering/enacting them. It became very clear that we need to have an in-house QC team whose task would be to listen to every audio file collected on the ground and tag such errors. We iteratively refined our error categorization, adapting to new types of errors as they were discovered. 

In their eagerness to assist participants, coordinators sometimes went above and beyond, inadvertently scripting entire responses or conversations. These were then merely read aloud by the participants, transforming what was supposed to be a spontaneous exchange into a rehearsed performance! This would diminish the authenticity of spontaneous speech. To combat these issues and preserve data integrity, we introduced errors categories like `Bad extemporaneous,' and `Book read,' so that such content could be tagged. Similarly, in telephonic conversations, a tendency emerged for one participant to dominate, leading to minimal contributions from the other party, a phenomenon tagged as `SST' (Single Speaker Talking) by our QA team. Categories like `Stretching,' `Repeating Content,' and `Long Pauses' were introduced to counter verbosity and repetition, ensuring the recordings were content-rich.

In several places, the authenticity of participants' identities emerged as a significant challenge. Concerns were raised when the voice of a participant didn't seem to align with their reported age and/or gender, leading to discoveries of intentional misinformation or unintentional errors in registration. Some instances revealed inconsistency in voices under the same participant ID, hinting at multiple individuals sharing a single ID, while others showed the same voice across different IDs, indicating individuals masquerading as multiple participants. To address these authenticity issues, we introduced a micro-task requiring participants to record a video stating basic information, allowing our QC team to verify age and gender visually. Disparities led to data rejection, while voice mismatches in audio samples triggered further scrutiny. Recognizing privacy concerns and cultural sensitivities, particularly among female participants reluctant to record videos, we offered alternatives like live verification through WhatsApp calls, conducted by female QC members, ensuring a respectful and secure verification process.

Data collection across diverse settings—outdoors, in public schools, small hotels, and participants' homes—brought the challenge of background noise interference. Distant ambient noises were less intrusive compared to the constant buzz of fans in closed spaces. Distinguishing between unavoidable natural background sounds and disruptive persistent noises was essential. 

An unexpected challenge arose with the capture of highly expressive, albeit profane, reactions to daily frustrations, necessitating a balance between authenticity and appropriateness. This led to the creation of an `Objectionable Content' category to carefully screen for hate speech or inappropriate content. Through this iterative process of reviewing audio files our QA team came up with 23 error categories which comprehensively captured everything that could go wrong!

\noindent \textbf{The Subtle Art of Transcription.} While we continued on our journey across the country, little did we know that our greatest challenge lay not in the fieldwork, but in the nuanced art of transcription. Initially perceived as a straightforward task of converting speech to text, the complexity of transcribing the diverse speech styles of thousands of individuals soon became apparent. The main issue was the different between colloquially spoken language and standardised language found in textbooks. The former contains words which may not have any spellings in standard textbooks or dictionaries but still cannot be ignored, simply because this is how people talk! This required us to choose between the pure phonetic representation of speech resulting in non-standard spellings on one hand, and pure textbooks representations which deviated from what was being said on the other. To address this we ended up with a two-level transcription approach wherein the first level the transcribers were asked to transcribe verbatim without worrying about correctness of spellings (thus emphasising only on phonetic fidelity). In the second level the transcribers were asked to standardise the transcription to convert the phonetically correct representations to nearest standard spellings in textbooks. Thus the lazily spoken Hindi word ``muje'' would be transcribed verbatim as ``muje'' in Level 1 ensuring phonetic fidelity and then standardised to ``mujhe'' in Level 2 ensuring spelling accuracy. However, aligning all language experts and transcribers with this novel framework proved challenging. Many transcribers initially resisted typing verbatim spellings, feeling it betrayed the standard writing style. Through extensive discussions, iterative rounds of feedback, and a collaborative effort across language teams, we established a set of guidelines that balanced standardized systems with fidelity to the actual sound wave. This meticulous process underscored transcription not just as a task but as an art form, requiring a deep understanding of linguistic nuances, cultural context, and the delicate balance between preserving the integrity of spoken language and adhering to standard linguistic conventions.

\noindent\textbf{Stories from the Heart of India.} Despite the hurdles, the journey was largely filled with numerous positive experiences. The diversity we encountered in weather, languages, and cultures was overwhelming, yet it filled our hearts with an indescribable warmth and a profound appreciation for India's rich cultural tapestry. The love and hospitality offered by the people, their eagerness to share their stories, and their enthusiasm for preserving their linguistic heritage were truly heartening. Our journey underscored the power of language as a bridge to understanding people, their cultures, and the nation at a deeper level.

In a world where technology often seems to isolate us, our project brought people closer together, allowing them to connect and share their lives through their native languages. This technical endeavor became a conduit for genuine human connection, enabling people from various backgrounds to express their lives, traditions, and experiences. From a Tamil Nadu entrepreneur sharing her journey to success, to a Kashmiri woman finding solace in prayer during tumultuous times; from a young boy in Kashmir divulging his secret recipe, to a Manipuri girl aspiring for higher education amidst challenges, each story added a unique voice to the rich mosaic of IndicVoices.

The diversity in stories we collected — ranging from personal achievements and cultural narratives to expressions of socio-political concerns — highlighted the importance of the dataset we were building. Whether it was an old man reminiscing about life seventy years ago, a professor discussing the significance of a local folk culture, or a young Nepali girl playfully sharing tales of her dates, these narratives painted a vivid picture of the diverse life across India. Such stories not only enrich our understanding but also celebrate the myriad facets of Indian life, from its challenges to its triumphs.

\noindent\textbf{Miles to go.} Despite the vast disparities in lifestyle, language, and social circumstances across different regions, a common thread of respect for linguistic heritage and a passion for technology united participants from all walks of life. This shared enthusiasm underscores a collective commitment to preserving India's linguistic diversity, bridging the gap between tradition and modernity. IndicVoices, thus, stands as a testament to the enduring spirit of India and its people, weaving together the voices of its many inhabitants into a vibrant tapestry of stories that resonate with authenticity, diversity, and a profound sense of belonging. Yet, as extensive as our journey has been, it is but a chapter in a much larger story. With over 15,000 hours of recordings still ahead, our expedition through the heart of India's linguistic landscape is far from over. Like the timeless verse, ``miles to go before I sleep'' our path stretches onward, promising more voices to be heard, more stories to be shared, and an ever-deepening appreciation for the rich mosaic of Indian culture and language. This journey has only just begun, and the road ahead is filled with the promise of discovery, understanding, and the celebration of India's incredible diversity.

\section{Instructions for participants}
The following instructions were given to the participants:

\noindent\textbf{Aim of the Project:} The aim of this project is to collect data for the development and evaluation of speech technology specifically tailored to your language. [The details of what is entailed by ``speech technology'' was colloquially explained by the coordinators to the participants using everyday applications.]

\noindent\textbf{Purpose of Data Collection:} Your data will be utilized for both commercial and non-commercial purposes to enhance speech technology for your language.

\noindent\textbf{Amount of Time:} The entire process is expected to take between 1 to 4 hours.

\noindent\textbf{Compensation:} You will receive compensation of INR X for your participation. The exact value of X will be communicated to you by the coordinator.

\noindent\textbf{Consent:} By participating, you agree to the terms and understand that your data will be used as mentioned above. Please proceed only if you consent to these terms and sign the consent form.

\noindent\textbf{Registration:} Your details will be collected during registration, ensuring your privacy is preserved. None of this information will be shared with third parties. You should also upload your signed consent form as a part of the registration process.

\noindent\textbf{App Installation:} Search for the ``<Anonymous>'' application on the Google Play Store using your Android smartphone and download the application.

\noindent\textbf{Login:} Use the access code assigned to you during registration to log in. Confirm this code with your coordinator before proceeding further.

\noindent\textbf{Fetch Tasks:} Press the "Submit Tasks/Get New Tasks" button to access the list of tasks assigned to you.

\noindent\textbf{Recording:} Click on each task and carefully read the prompt provided. If you find any task unclear, consult your coordinator for clarification.
Take your time to formulate your response. If you are uncomfortable with a question, feel free to skip it. Take breaks as needed and complete the tasks at your own pace.

\noindent\textbf{Submit a Task:} After recording your response, click on "Stop". The application will then replay your response. If you and/or the coordinator are satisfied with the response, click on the "Next" arrow. Otherwise, click on "Record" to re-record your response.

\noindent\textbf{Two-party conversations:} Once you have finished the tasks on the App, the coordinator will pair you with a participant for conversation on a specific scenario. You can request to be paired with a participant of your choice and choose a role as well as scenario from a long list of roles scenarios that will be shared by the coordinator.

\noindent\textbf{Logging Out:} Once you have finished all tasks, return to the home screen and click on "Submit Tasks". Wait for a message confirming that all tasks have been submitted. You will then be prompted to record a video for identity verification purposes. This video will not be shared publicly and will be deleted once verified by our Quality Control team. If you are uncomfortable recording a video, inform your coordinator to arrange a WhatsApp call with a female member of our central team. This call will not be recorded, ensuring your privacy.

\clearpage

\section{QC Categories and Noise tags}
\label{apx:qc-noise-tags}
\begin{table}[!h]
\centering
\scriptsize
\begin{tabularx}{\textwidth}[t]{sz}
\arrayrulecolor{green}\toprule
\textbf{Error Type} & \textbf{Explanation} \\
\arrayrulecolor{green}\midrule
Low volume & Audio not loud enough for clarity. \\ 
\arrayrulecolor{black}\midrule
Intermittent noise & Background sounds disrupting audio at intervals. \\ 
\arrayrulecolor{black}\midrule
Persistent noise & Constant background noise throughout the audio. \\ 
\arrayrulecolor{black}\midrule
Intermittent chatter & Background conversations at intervals. \\ 
\arrayrulecolor{black}\midrule
Persistent chatter & Continuous background conversations throughout the audio. \\ 
\arrayrulecolor{black}\midrule
Unclear audio & Poor audio clarity. \\ 
\arrayrulecolor{black}\midrule
Off topic &  Response deviating from the given prompt. \\ 
\arrayrulecolor{black}\midrule
Repetitive content & Unnecessary repetition of nuances in the audio. \\ 
\arrayrulecolor{black}\midrule
Long pauses & Unusually long silences in the audio. \\ 
\arrayrulecolor{black}\midrule
Mispronunciations & Incorrect pronunciation of words. \\ 
\arrayrulecolor{black}\midrule
Reading prompt & Participant reading the prompt instead of responding with an answer. \\ 
\arrayrulecolor{black}\midrule
Book read & Monotonous, scripted reading detected. \\ 
\arrayrulecolor{black}\midrule
Single speaker talking & Only one speaker actively participating in the roleplay task.\\ 
\arrayrulecolor{black}\midrule
Stretching & Unnatural elongation of words or sounds. \\ 
\arrayrulecolor{black}\midrule
Objectionable content & Inappropriate or offensive use of words in the audio. \\ 
\arrayrulecolor{black}\midrule
Incorrect text prompt & Prompt provided to the participant was faulty and the participant has corrected it while speaking. \\ 
\arrayrulecolor{black}\midrule
Factual inaccuracy & Response recorded by a speaker contains factually incorrect information.  \\ 
\arrayrulecolor{black}\midrule
Skipping words & Omission of crucial parts of the prompt. \\ 
\arrayrulecolor{black}\midrule
Wrong language & Incorrect language used for the task. \\ 
\arrayrulecolor{black}\midrule
Presence of echo & Echo affecting sound quality. \\ 
\arrayrulecolor{black}\midrule
Bad extempore quality & Poor naturalness and fluency in extempore tasks. \\ 
\arrayrulecolor{black}\midrule
Wrong gender (metadata) & Discrepancy between provided information and spoken content regarding gender \\
\arrayrulecolor{black}\midrule
Wrong age group (metadata) & Inconsistency between entered age details and information verified through video verification. \\
\arrayrulecolor{black}\midrule
Additional Comments & Other noteworthy observations not covered above. \\
\arrayrulecolor{green}\bottomrule
\end{tabularx}
\caption{Error categories identified by the QC team}
\label{apx:tab:error-categories-table}
\end{table}

\begin{table}[!h]
\scriptsize
\centering
\begin{tabular}
{|@{\hspace{0.3em}}l@{\hspace{0.5em}}|@{\hspace{0.3em}}l@{\hspace{0.5em}}|@{\hspace{0.3em}}l@{\hspace{0.5em}}|@{\hspace{0.3em}}l@{\hspace{0.5em}}|@{\hspace{0.3em}}l@{\hspace{0.5em}}|@{\hspace{0.3em}}l@{\hspace{0.5em}}|@{\hspace{0.3em}}l@{\hspace{0.5em}}|@{\hspace{0.3em}}l@{\hspace{0.5em}}|@{\hspace{0.3em}}l|}
\arrayrulecolor{green}\toprule\arrayrulecolor{black}
TV & animal & baby & baby\_crying & tsk & barking & beep & bell & sniffle \\
\midrule
child & child\_crying & child\_laughing & child\_talking & dishes & child\_yelling & children & children\_talking & tones \\
\midrule
click & clicking & clink & clinking & cough & child\_whining & door & footsteps & gasp \\
\midrule
horn & hum & inhaling & laughter & meow & motorcycle & music & nose\_blowing & noise \\
\midrule
phone\_vibrating & popping & pounding & screeching & rattling & ringing & rustling & scratching & printer \\
\midrule
smack & sneezing & sniffing & bird\_squawk & snorting & squawking & squeak & stammers & static \\
\midrule
throat\_clearing & thumping & tone & children\_yelling & trill & baby\_talking & typewriter & ugh & uhh \\
\midrule
wheezing & whispering & whistling & yawning & yelling & buzzer & clanking & phone\_ringing & hmm \\
\midrule
unintelligible & buzz & clanging & persistent-noise-start & tapping & singing & talking & umm & hiss \\
\midrule
breathing & chiming & groan & persistent-noise-end & sigh & swallowing & uh-huh & siren & \\
\arrayrulecolor{green}\bottomrule
\end{tabular}
\caption{Different noise tags supported by Shoonya in transcription.}
\label{apx:tab:noise-tags-table}
\end{table}

\clearpage



\section*{Supplementary material}

# D  Level 2 Transcription Guidelines for Hindi

To access L2 Transcription guidelines for all 22 Indian languages, please login to the following link[8]

**Rules for correcting consonant errors**

1. If aspirated and unaspirated consonants can be interchanged then interchange it for the standardized spelling. For example,
    a. -b (ब) and -$b^h$ (भ)
        i. sabi (सबी) -> $sab^h$i (सभी)
    b. -t (ट ) and -th (ठ ):
        i. meeta (मीटा ) -> $meet^h$a (मीठा)
    c. -j (ज) and -jh (झ):
        i. janDaa (जंडा) -> $j^h$anDaa (झंडा)
    d. -g (ग) and -$g^h$ (घ)
        i. gar (गर) -> $g^h$ar (घर)F
    e. -k (क) and -KH (ख)
        i. $d^h$oka (धोका) -> $d^h$o$k^h$aa (धोखा)
    f. -f (फ़ ) and -$p^h$ (फ):
        i. fool (फ़ूल) -> $p^h$ool (फूल)
        ii. pheslaa (फैसला) -> feslaa (फ़ैसला)

2. If consonants are spoken in a way that is very close to the correct form (can be common or uncommon in spoken form) then change it to the correct form. For example,
    a. -s and -sh
        i. siksha( सिक्षा) -> shikshaa (शिक्षा)
    b. -d`h and - r`h
        i. pad`hnaa (पढना) -> par`hnaa(पढ़ना)
    c. -sh  and -s :
        i. $s^h$o$s^h$al (शोशल) -> so$s^h$al (सोशल)
        ii. saamil (सामिल) -> $s^h$aamil (शामिल)
        iii. chasmaa  (चस्मा) -> cha$s^h$maa (चश्मा)
    d. -d` and -r
        i. larka(लरका ) -> lad`ka (लड़का)
        ii. $b^h$eed` (भीर ) -> $b^h$eer` (भीड़)
        iii. bar`tan (बड़तन)  -> bartan (बर्तन)
        iv. sad`ak (सडक) -> sar`ak (सड़क)
    e. Interchange between -cch and -ksh
        i. ikshaa (इक्षा) -> icchaa (इच्छा)
        ii. cchetra (छेत्र) -> kshetra(क्षेत्र)

3. If the interchange of consonant clusters helps in achieving the standard spelling for a word (metathesis), such a transformation is allowed as long as it does not deviate the resultant word from the spoken-text. For example,
    a. Interchange between -tf  and -ft
        i. luft (लुफ़्त) -> lutf (लुत्फ़)
        ii. Ustah (उस्ताह) -> Utsah (उत्साह)

4. If consonants are mispronounced because of its adjoining letter, then correct it (phonological assimilation).  For example,
    a. Kuut.ta (कूत-ता)  -> Kuud.ta (कूदता)
    b. kitap parh ke (किताप पढ़ के) -> kitab parh ke (किताब पढ़ के))
    c. kus sunna (कुस सुना) -> kuch sunna (कुछ सुना)
    d. chammas se (चम्मस से ) -> chammach se (चम्मच से)

---

[8]https://ai4bharat.iitm.ac.in/indicvoices

**Rules for correcting vowel errors**

1. If vowel sounds are elongated or shortened or interchanged due to emphasis/modulation then change it to the standard form. For example,
    a. -a  and -aa
        i. kaahaanii (काहानी) -> kahaanii (कहानी)
        ii. bahaar (बहार) -> baahar (बाहर)
    b. -i  and -ii :
        i. miisaal (मीसाल)  -> misaal (मिसाल)
    c. -u  and -uu :
        i. uudhaar (ऊधार)  -> udhaar (उधार)
        ii. ungli (उंगली)  -> uungli (ऊँगली)
        iii. uncha(उंचा)  -> uuncha (ऊँचा)
    d. -o  and -u :
        i. Kyunki (क्यूँकि)  -> kyonki (क्योंकि)
    e. -e  and -a:
        i. bechen (बेचेन) -> bechain(बेचैन)
        ii. chain (चैन) -> chen (चेन) (गले की चेन)
    f. -a and -i :
        i. kataab (कताब) -> kitaab (किताब)
        ii. salaaii (सलाई)-> silaaii (सिलाई)

2. If interchanging monothongs (a, aa, e, i, ii, o, u, uu) with dipthongs (ai, au, etc) allows you to arrive at the standard spelling for the word, then please interchange. Note that diphthongs and triphthongs are combinations of two or three vowels respectively (aa + o = au). For example,
    a. hei(हे) -> hai (है) [e v/s ae]
    b. mei (में ) -> mein (मैं)  [e v/s ae]
    c. nao (नो) -> nau (नौ)  [o v/s au]

3. If removal or addition of a vowel helps in getting to the standard spelling of a word, such a transformation is allowed. For example,
    a. removal of -a (using halant)
        i. darad (दरद) -> dard (दर्द)
        ii. mulak (मुलक) -> mulk (मुल्क)
        iii. khatam (खतम) -> khatm (खत्म)
    b. removal of -i (using halant) :
        i. Kirya (किरया) -> Kriya (क्रिया)
    c. addition of -i
        i. Briyani ( ब्रियानी ) -> biryani (बिर्यानी)

4. If vowels are replaced by consonants because of accent, then change it to the standard form. For example,
    a. -gu and  -u :
        i. gunnees (गुन्नीस ) -> unnees (उन्नीस)
    b. -a  and  -h:
        i. haspataal (हस्पताल) -> aspataal (अस्पताल)

**Rules for dealing with errors at word beginnings**

1. If word beginning requires insertion/deletion of a vowel or a schwa, then change it to the standard spelling. For example,
    a. Deletion of इ:
        i. Istree ( इस्त्री) -> Stree (स्त्री)
        ii. Isthan (इस्थान) -> स्थान (Sthan)

**Rules for dealing with errors at word endings**

1. If word ending has an extra vowel then correct it to make it to standard form. For example,
    a. Himalayaa(हिमालया) -> Himaalya (हिमालय)
    b. kavii (कवी) -> kavi (कवि)
    c. matii (मती) -> mati (मति)

2. If additional schwa is added at the end of words ending in consonant sounds then remove it to bring it to standard form. For example,
    a. vageraa (वगैरा)  -> vagerah (वगैरह)
    b. yogaa (योगा) –> yoga (योग)
    c. karmaa (कर्मा) -> karma (कर्म)
    d. raamaa(रामा) -> raam(राम)

3. If a consonant is missed or skipped at the end of a word then add it to bring it to the standard form. For example,
    a. missed -ta
        i. takh (तख ) -> taKHt (तख्त)
    b. missed  -h:
        i. Guna (गुना) -> gunah (गुनाह)
        ii. fate (फते) -> fatah (फतह)

**Rules for dealing with shortened syllables**

1. If splitting of a shortened syllable followed by character additions standardizes the spellings of the words, then such a transformation should be allowed, while ensuring the sound of the resultant words do not deviate from the spoken text
    a. Bolraa (बोलरा)  -> Bol rahaa (बोल रहा)
    b. Jaraa (जारा)  -> Ja rahaa (जा रहा)
    c. Manraa (मानरा)  -> Maan rahaa (मान रहा)
    d. Mujhebi (मुझेबी) -> Mujhe bhi (मुझे भी)
    e. Muje-i (मुजेई) -> Mujhe hi (मुझे ही)
    f. Teraa-i (तेराई) -> teraa hi (तेरा ही)

2. If present tense markers have been shortened or eliminated in the phrases, then change them to arrive at the standard form.
    a. shortening of hai (है), hun (हूँ)  and hain(हैं) :
        i. Kartau (करताऊँ)   -> Karta hun (करता हूँ)
        ii. Kartaye (करताए)  -> Karta hai (करता है)
        iii. Aatain (आतए)   -> Aate hain (आते हैं)

**Rules for dealing with nasalised sounds**

1. If addition or deletion of a nasalised sound can bring a word into its standard form, then such a transformation is allowed if it does not impact the context in which the word was uttered. e.g.:
    a. addition of diacritic (बिंदु - ◌ं and चंद्रबिंदु ◌ँ )
        i. mujhe kae jana tha (मुझे कही जाना है)  -> mujhe kahin jaana tha (मुझे कहीं जाना है)
        ii. ye saare mere ghanisht mitr hai (ये सारे मेरे घनिष्ठ मित्र है) -> ye saare mere ghanisht mitr hain (ये सारे मेरे घनिष्ठ मित्र हैं)
        iii. kahaaniyaa (कहानिया) -> kahaaniyaan (कहानियाँ)
        iv. yaha (यहा) -> yahaan (यहाँ)
    b. deletion of (बिंदु) :
        i. mere kaen dost hain (मेरे कईं दोस्त हैं) -> mere kayi dost hain (मेरे कई दोस्त हैं)
    c. deletion of unnecessary '-n' sound:
        i. Karenga (करेंगा)  -> Karega  (करेगा)
        ii. Bolenga (बोलेंगा)  -> Bolega (बोलेगा)
        iii. Jaayenga (जाएँगा)  -> Jaayega (जाएगा)

**Rules for dealing with loan words**

1. If common words borrowed from english or english named entities such as proper nouns, places, brands, food, month/day names, etc are present in isolation without any inflection (-ein, -ne, -ka, etc), then add the english spelling in brackets, while leaving the original transcript as verbatim. For example,
    a. Proper Nouns
        i. amjon (ऐमजॉन) -> (ऐमजॉन) [amazon]  (note that the (हिंदी) spelling has **not** been corrected from "amjon " to "amazon" but the correct english spelling has been added in brackets []
        ii. aadidas (आदिदास)  -> आदिदास [adidas]
        iii. aktubar(अक्टूबर)  -> अक्टूबर [october]
        iv. wednasde (वेन्सडे) -> वेन्सडे [Wednesday]
        v. disember(दिसंबर)  -> दिसंबर [December]
    b. Common words
        i. on (ओन ) -> ओन [on]
        ii. ooder (ओडर) -> ओडर [order]
        iii. order(ऑर्डर)  -> ऑर्डर [order]
        iv. fusT(फस्ट)  -> फस्ट [First]
        v. baseball(बेसबॉल) ->  बेसबॉल [baseball]
        vi. Cupcake(कपकेक)  -> कपकेक [Cupcake]
        vii. Playground(प्लेग्राउंड)  ->  प्लेग्राउंड [playground]
        viii. kilogram(किलोग्राम) ->   किलोग्राम [kilogram]

2. If certain sounds in borrowed words are not present in the language's orthography (names/places from different cultures, states) then come up with the closest way as mentioned in the conventions of your language and use that consistently across all instances of the same word. The same holds true for words which can be written in multiple ways. The idea is to ensure consistency across all utterances of a spoken word.
    a. tamiḷ (தமிழ்) -> tamil (तमिल)

3. If acronyms are spoken word by word (in English)  then always write it with a space. E.g.:
    a. NRI -> N R I (एन [N] आर [R] आई [I])
    b. FBI -> F B I (ऐफ़ [F] बी [B] आई [I])
    c. MPin -> M Pin (एम [M] पिन [Pin)
    d. SBI -> S B I (एस [S] बी[B] आई [I])

4. If acronyms are spoken as a single word then write it as it is because spacing it would deviate it from the spoken-text. E.g:
    a. ASAP (एसेप) and not A S A P (ए [A] एस [S] ए [A] पी [P])
    b. BhaJPa (भाजपा) and not B J P (बी जे पी) or Bha J Pa (भा जे पा)

**Rules for dealing with numbers, units of time, measurement, etc**

1. If numbers (ordinals and cardinals), time, and other units of measurement (kg, m, cm, ft, etc) are uttered in English then represent them exactly as it is in the target transcription language. E.g.:
    a. kilogram (किलोग्राम [kilogram])
    b. kg (कि.[k] ग्रा.[g])
    c. first (फर्स्ट[first])

2. If numbers (ordinals and cardinals), time, and other units of measurement (kg, m, cm, ft, etc) are uttered in the native language, then always use the standardized native language norms and spellings. e.g.:
    a. gunniis (गुन्निस) -> unniis (उन्नीस)
    b. paach (पाच) -> paanch (पाँच)
    c. miTer (मिटर) -> miiTer (मीटर)
    d. pelii (पेली) -> pahli (पहली)

**Rules for compound/conjunction/combined words/Reduplication**

1. If compound words are used in non-standard form, then conjunction can be added to change it to the standard form.
    a. Inapplicable for Hindi

2. In case of reduplicated words, when the next word is not in the dictionary, please add a hyphen(-) in between and write it as it is spoken.e.g:
    a. chai vai (चाय-वाय)
    b. Roti voti (रोटी-वोटी)
    c. daal vaal (दाल-वाल)

3. If insertion or removal of a space is required to standardize the spelling of a compound word, such a transformation is allowed.
    a. Inapplicable for Hindi

**Rules for word contractions:**

1. Inapplicable for Hindi

**Rules for colloquial tardiness:**

1. If addition of a genitive marker results in standard form, then such a transformation is allowed.
    a. addition of -ri (री), -ra (रा), -re (रे)
        i. humaai kitab (हमाई किताब) -> humari kitaab (हमारी किताब)
        ii. humaye gaaon me (हमाए गाँव में) -> humare gaaon me (हमारे गाँव में )
    Note that here both "humaai"(हमाई) and "humaye" (हमाए) are not being changed to "hum" (हम) because it differs significantly from the spoken text.

2. If changes in dative (direction) forms help in arriving at a standard spelling, then such a transformation is allowed to a level where it does not deviate much from the spoken text:
    a. tumar ko (तुमार को) -> tumhare ko (तुम्हारे को)
    b. humar ko (हमार को) -> humare ko (हमारे को)
    c. mere ko (मेरे को) -> mere ko (मेरे को) [no change]
    d. mer ko (मेर को) -> mere ko (मेरे को)
    e. meko (मैंको) -> main ko (मैं को) [split to meaningful units]
    f. tumaare ko (तुमारे को) -> tumhare ko (तुम्हारे को)
    Note: "tumar" (तुमार) is not being changed to "tum" (तुम), "mere" (मेरे) is not being changed to "mujhe" (मुझे) because they differ significantly from the spoken text.

3. If a small change in form of a word helps in arriving at a correct spelling, then such a change is allowed.
    a. Bol riya (बोल रिया) -> bol rahaa (बोल रहा)
    b. Bol riye (बोल रिए) -> bol rahe (बोल रहे)

**Rules for gemination:**

1. The gemination (doubling consonant) is represented in orthography with diacritics, matras or by doubling consonants. Please follow the conventional way of writing geminated words in your language and maintain the consistency across all spellings of an utterance. e.g.
    a. Addition of halant (`)
        i. patta ( पत．ता) --> patt+ta (पत्+ता -> पत्ता)
        ii. kaccha (कच．चा) -> kach+cha (कच्+चा -> कच्चा)

2. If double emphasis in words is not very obvious but the speaker means it, then change it to the double emphasis form. This could happen in instances of proper nouns with no standardized spellings, brand names, lesser known food items, brands, etc. e.g.:
    a. apam (अपम) -> appam (अप्पम)
    b. Aka (अका)-> akka (अक्का)
    c. Ana (अना) -> Anna (अन्ना)
    d. bhayaa (भया) -> bhayyaa (भय्या)

**Don't change during Level-2:**

1. Borrowed words with native suffixes (mostly plural) should be written as it is, without changing or translating the suffixes. e.g.:
    a. calendar-on (क्लैंडरों) should not be written as calendar (कैलेंडर)
    b. side-ein (साइडें) should not be written as side (साइड)
    c. Kantriyoun (कंट्रीयों) should not be written as countries (कन्ट्रीज)
    d. Friendo (फ्रेण्डों) should not be written as friends (फ्रेंड्ज़)

2. If splitting of a shortened syllable followed by character additions standardises the spellings but makes it deviate significantly from the waveform (spoken text), then we should not be allowing such transformation:
    a. unne (उनने) <-> unho ne (उन्होंने) [not-allowed]
    b. inne (इनने) <-> inho ne (इन्होंने) [not-allowed]

3. Spoken/dialectic words should not be replaced by the standard versions, as they deviate a lot from the wave form.
    a. aho (आहो) <-> haan (हाँ) [not-allowed]
    b. Hav (हव) <-> haan (हाँ) [not-allowed]
    c. abe oye (अबे ओय) <-> arey o (अरे ओ) [not-allowed]

4. If new affix patterns, echo words or reduplication patterns are innovated by the speaker, or slangs used only in spoken languages are found, then retain them as it is (with space) as
corrections will deviate too much from the spoken form. E.g:
    a. Loot laat (लूट लाट) [shall be retained as it is]
    b. Khaali peeli (खाली पीली) [shall be retained as it is]
    c. Paani puuni (पानी पूनि) [shall be retained as it is]

5. If grammatical corrections make the resultant text deviate much from the spoken-text, then such changes should not be made and original words should be retained. e.g.:
    a. Corrections disallowed in dative forms
        i. mereko ghar jana hai (मेरे को घर जाना है) <-> mujhe ghar jana hai (मुझे घर जाना है) [not-allowed]

6. The conjunction should be infixed, or removed only where it's absolutely necessary as per standardized spelling rules. If the 2 words can stand alone and have a spelling then, do not add meaningful affixes as per the rule.
    a. Inapplicable for Hindi

7. Severe colloquial contractions in Tamil and other languages, such as "vandethanalla" should not be expanded to their textbook forms such as "vande vithana allava"
    a. Inapplicable for Hindi

[End]



\end{document}